\documentclass[]{sensenova}

\usepackage{hyperref}
\usepackage{cleveref}
\usepackage{verbatim}

\usepackage{wrapfig}
\usepackage{graphicx}
\usepackage{floatrow}
\usepackage{placeins}
\usepackage{subcaption}
\usepackage{listings}
\usepackage{algorithm}
\usepackage{subcaption}
\usepackage{dsfont}
\usepackage{pifont}

\usepackage{mmstyles}

\usepackage[toc,page,header]{appendix}

\usepackage{xspace}
\usepackage[utf8]{inputenc}  
\usepackage[T1]{fontenc}     
\usepackage{CJKutf8} 

\usepackage[misc]{ifsym}

\usepackage{CJKutf8} 
\usepackage[colorinlistoftodos,prependcaption,textsize=small]{todonotes}

\usepackage{fontawesome5}

\usepackage{soul} 
\definecolor{HLGreen}{rgb}{0.78,0.95,0.78}
\definecolor{HLRed}{rgb}{0.9725,0.8431,0.8549}   
\definecolor{benchblue}{RGB}{58, 122, 252}

\newcommand{\colornum}[1]{%
  \footnotesize
  \ifdim #1 pt < 0pt
    \textcolor{sensepurple}{#1}%
  \else
    \textcolor{benchblue}{#1}%
  \fi
}

\usepackage{cuted} 

\usepackage[table]{xcolor}
\definecolor{bestcolor}{RGB}{219, 208, 237}
\definecolor{secondcolor}{RGB}{241, 237, 248}
\definecolor{thirdcolor}{RGB}{211, 222, 190}
\definecolor{line-blue}{RGB}{243, 248, 252}
\definecolor{line-green}{RGB}{200,242,200}
\definecolor{line-red}{RGB}{255,215,215}

\usepackage{adjustbox}
\usepackage{makecell}
\usepackage{booktabs}

\newlength{\ModelW}
\newlength{\NonModelW}
\newcolumntype{M}{>{\centering\arraybackslash}p{\NonModelW}}
\usepackage{graphicx}
\usepackage{subcaption}

\usepackage{array}      
\usepackage{caption}
\usepackage{subcaption} 

\usepackage{enumitem}

\usepackage{multirow}
\usepackage{makecell}

\usepackage{booktabs,tabularx,array}
\newcolumntype{Y}{>{\centering\arraybackslash}X}
\newcolumntype{L}{>{\raggedright\arraybackslash}X}

\usepackage{framed}
\usepackage{algorithm}
\usepackage{algorithmic}
\usepackage{listings}
\usepackage{soul}          

\definecolor{cotredbg}{RGB}{255,220,220}   
\definecolor{cotgreenbg}{RGB}{220,255,220} 

\title{SenseNova-U1: Unifying Multimodal Understanding and Generation with NEO-unify Architecture}

\abstract{

Recent large vision–language models (VLMs) remain fundamentally constrained by a persistent dichotomy: understanding and generation are treated as distinct problems, leading to fragmented architectures, cascaded pipelines, and misaligned representation spaces. 
We argue that this divide is not merely an engineering artifact, but a structural limitation that hinders the emergence of native multimodal intelligence.
Hence, we introduce \textbf{SenseNova-U1}, a native unified multimodal paradigm built upon \textbf{NEO-unify}~\cite{sensenova2026neounify}, in which understanding and generation evolve as synergistic views of a single underlying process.

\vspace{0.4em}

We launch two native unified variants, \textbf{SenseNova-U1-8B-MoT} and \textbf{SenseNova-U1-A3B-MoT}, built on dense (8B) and mixture-of-experts (30B-A3B) understanding baselines, respectively. 
Designed from first principles, they rival top-tier understanding-only VLMs across text understanding, vision–language perception, knowledge reasoning, agentic decision-making, and spatial intelligence.
Meanwhile, they deliver strong semantic consistency and visual fidelity, excelling in conventional or knowledge-intensive any-to-image (X2I) synthesis, complex text-rich infographic generation, and interleaved vision–language generation, with or without think patterns.
Beyond performance, we show detailed model design, data preprocessing, pre-/post-training, and inference strategies to support community research.

\vspace{0.4em}

Last but not least, preliminary evidence demonstrates that our models extend beyond perception and generation, performing strongly in vision–language–action (VLA) and world model (WM) scenarios.
This points toward a broader roadmap where models do not translate between modalities, but think-and-act across them in a native manner. Multimodal AI is no longer about connecting separate systems, but about building a unified one and trusting the necessary capabilities to emerge from within.

\begin{center}
\vspace{0.4em}
\includegraphics[width=0.97\textwidth]{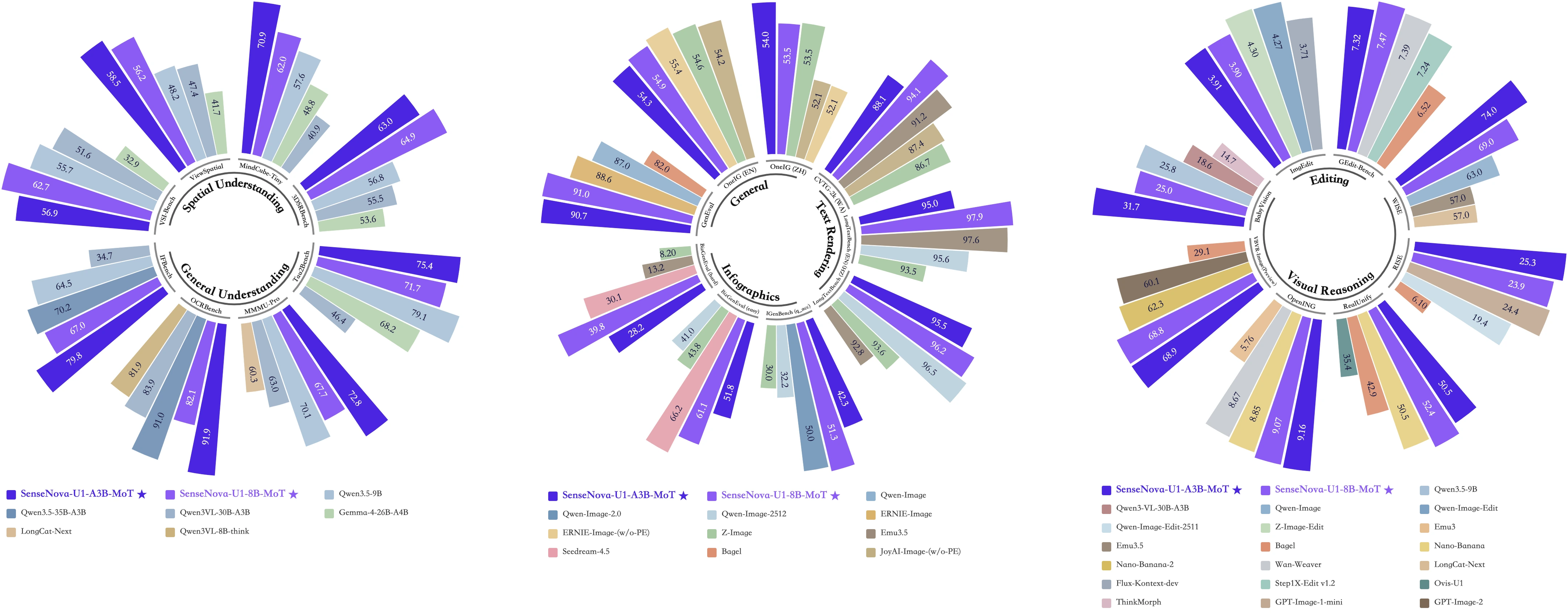}
\vspace{-1.2em}
\end{center}

\checkdata[Official Demo]{\url{https://unify.light-ai.top/}} 
\checkdata[GitHub Code]{\url{https://github.com/OpenSenseNova/SenseNova-U1}}
\checkdata[HuggingFace Model]{\url{https://huggingface.co/collections/sensenova/sensenova-u1}} 
\checkdata[NEO-unify Blog]{\url{https://huggingface.co/blog/sensenova/neo-unify} (March 5, 2026)}
} 

\begin{document}
\maketitle

\begin{figure}[H]
  \centering
  \includegraphics[width=\textwidth]{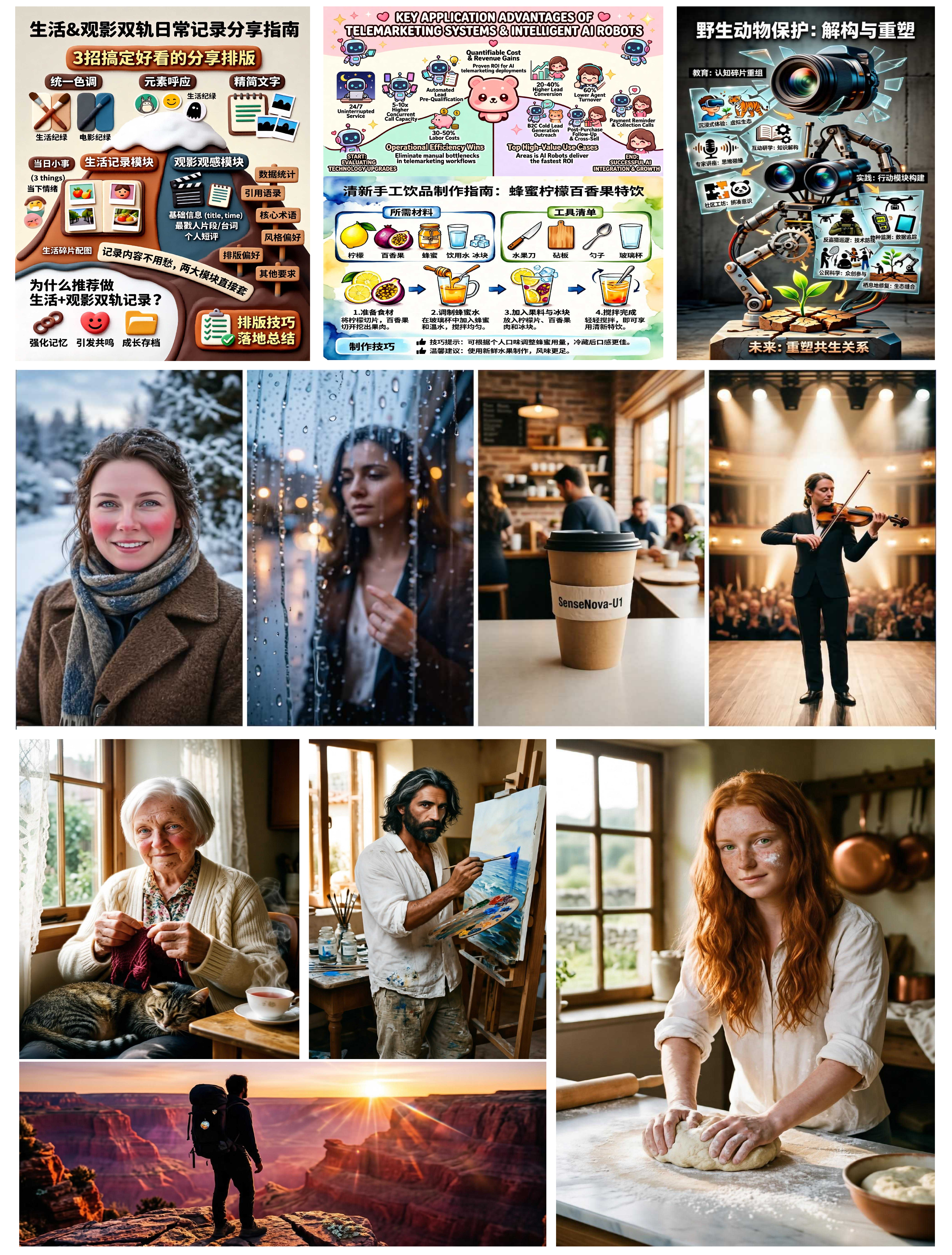}
  \caption{Showcases of SenseNova-U1-8B-MoT in infographics and human generation.}
  \label{fig:teaser_figure_t2i}
\end{figure}

\clearpage
\begin{figure}[H]
  \centering
  \includegraphics[width=\textwidth]{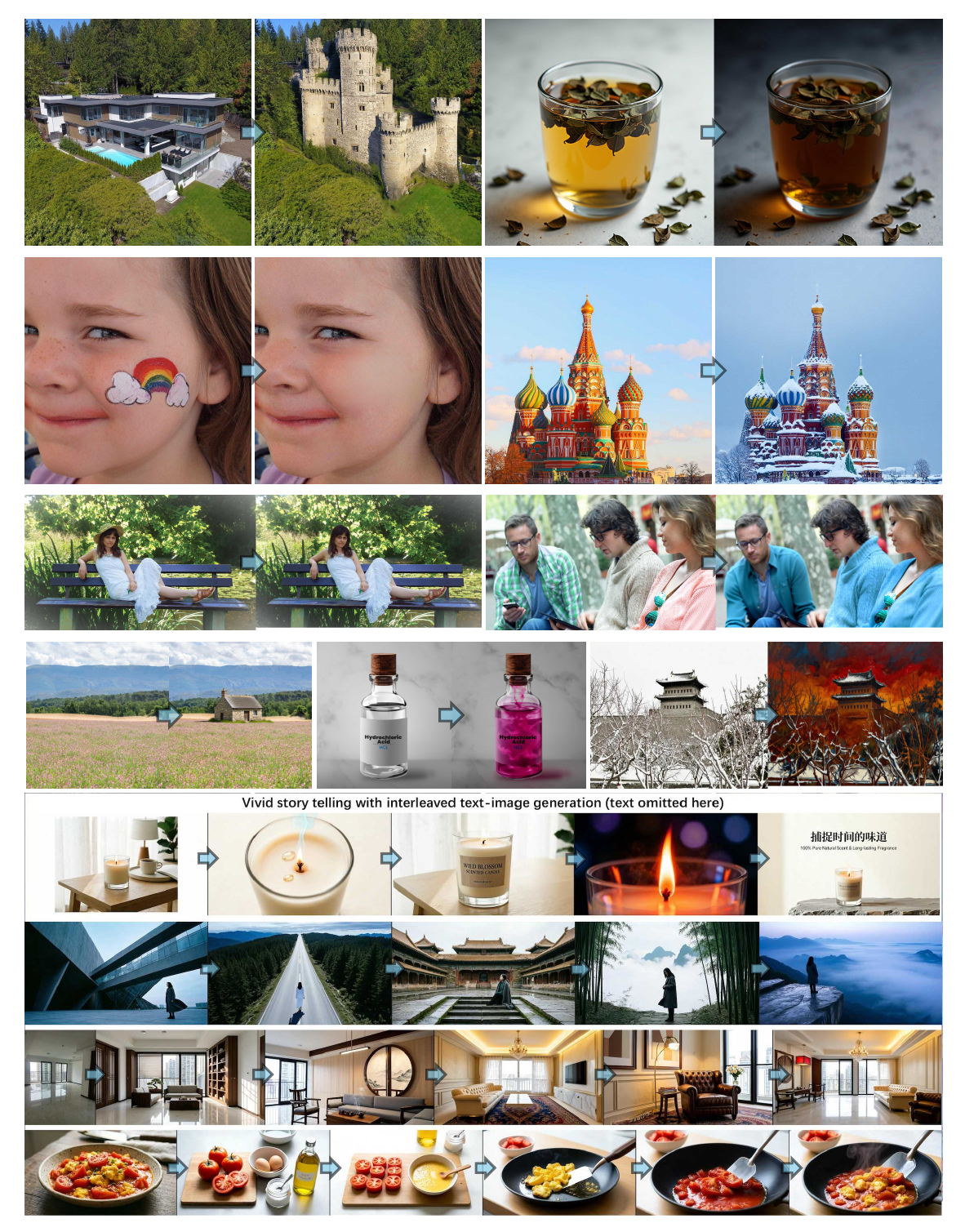}
  \caption{Showcases of SenseNova-U1-8B-MoT in image editing and interleaved generation.}
  \label{fig:teaser_figure_other}
\end{figure}
\clearpage

{
\hypersetup{linkcolor=blue}
\tableofcontents
}
\clearpage

\section{Introduction}
\label{sec:intro}

Recent advances in multimodal foundation models~\cite{Qwen3-VL,wang2025internvl3,flux2024} have markedly enhanced both perception and generation across vision and language. 
Yet these capabilities have largely evolved in isolation. 
This divide stems from the underlying system design: understanding is typically mediated by pretrained vision encoders (VEs)~\cite{sun2023eva,VLP:SigLIP,VLP:CLIP}, whereas generation relies on latent variational autoencoders (VAEs)~\cite{vae,vavae}. 
These choices impose distinct learning objectives~\cite{VLP:CLIP,vae} and training pipelines~\cite{flamingo,blip2,liu2023llava,rombach2021highresolution}, resulting in divergent feature representations that bifurcate multimodal modeling into separate regimes.
Consequently, early unified multimodal models (UMMs)~\cite{deng2025bagel,chen2025blip3o,wu2025qwenimagetechnicalreport,wu2024janus,chen2025janus,lin2025uniworld} remain loosely integrated, with perception and generation connected through different tokenizers, latent spaces, or auxiliary modules rather than being learned jointly within a truly unified system.

Against this backdrop, native vision–language models (VLMs) have emerged along two distinct directions.
One casts multimodality as an extension of language, mapping all modalities into discrete tokens within a unified autoregressive framework~\cite{Chameleon, MOMA, MoT, wang2024emu3,cui2025emu35nativemultimodalmodels, ma2025unitok, Dualtoken, team2026longcat}. 
While enabling seamless cross-modal reasoning, this discretization inevitably compresses non-linguistic signals into lossy representations, constraining both high-level semantics and visual fidelity.
The other instead pursues a unified continuous visual interface spanning understanding and generation~\cite{zhou2024transfusion, zheng2025diffusion, fan2025prism, vavae, liu2025tuna, tong2026beyond}, seeking to reconcile conceptual structure with high-fidelity reconstruction within a shared representation space — but often with trade-offs.
Yet neither resolves the fundamental tension between semantic abstraction and pixel-level granularity. This leaves open a central question: can multimodal intelligence be unified in a truly native form, breaking free from latent bottlenecks and intermediate representations?

We return to the first principles: building a model that directly engages with native inputs (\textit{i.e.} pixels and words), and steps beyond representation arguments or pre-trained priors. 
Crucially, we dispense with both pretrained vision encoders and deep decoder heads, yielding a unified architecture that supports concise and scalable training.
Hence, we introduce \textbf{SenseNova-U1}, a native unified multimodal paradigm built on the \textbf{NEO-unify}~\cite{sensenova2026neounify} model. 
As a first step toward truly end-to-end unification, it learns directly from lossless inputs and self-organizes potential representation spaces tailored to diverse application scenarios.
Specifically, it incorporates:
(i) a near-lossless visual interface that simultaneously preserves semantic structure and fine-grained pixel detail without any pretrained VEs or VAEs;
(ii) a unified end-to-end modeling over raw inputs that jointly couples autoregressive cross-entropy for language with pixel-space flow matching for vision;
(iii) a native mixture-of-transformers (MoT) architecture that synergizes understanding and generation in an intrinsically multimodal system with minimal objective interference and powerful scaling efficiency.

We launch two variants, \textbf{SenseNova-U1-8B-MoT} and \textbf{SenseNova-U1-A3B-MoT}, built upon dense (8B) and mixture-of-experts (30B-A3B) multimodal understanding backbones, respectively. 
Both models adopt a native MoT architecture, enabling efficient scaling while reducing interference across heterogeneous multimodal objectives.
Empirically, SenseNova-U1 rivals top-tier understanding-only VLMs across text understanding, vision-language perception, knowledge reasoning, agentic decision-making, and spatial intelligence, while simultaneously achieving strong any-to-image (X2I) generation performance under a 32$\times$ compression ratio across conventional, knowledge-intensive, and text-rich scenarios.
Beyond them, it supports visual-centric reasoning and coherent interleaved generation across modalities, enabling applications such as illustrated guides, visual storytelling, presentations, posters, comics, resumes, and other information-dense visual formats requiring structured layout generation and high-fidelity rendering.
Overall, SenseNova-U1 sets a brand-new paradigm for unified multimodal understanding and generation, outperforming prior open-source models across a wide range of understanding, reasoning, and generation benchmarks.

Preliminary experiments further suggest promising capabilities in vision--language--action (VLA) and world modeling (WM), indicating that our models can reason and act natively across modalities without relying on external adapters or modular bridges. More broadly, SenseNova-U1 points toward a shift in multimodal AI: from connecting separate modality-specific systems to learning perception, reasoning, and generation within a natively unified architecture.

\begin{figure*}
    \centering
    \includegraphics[
        width=0.99\linewidth,
    ]{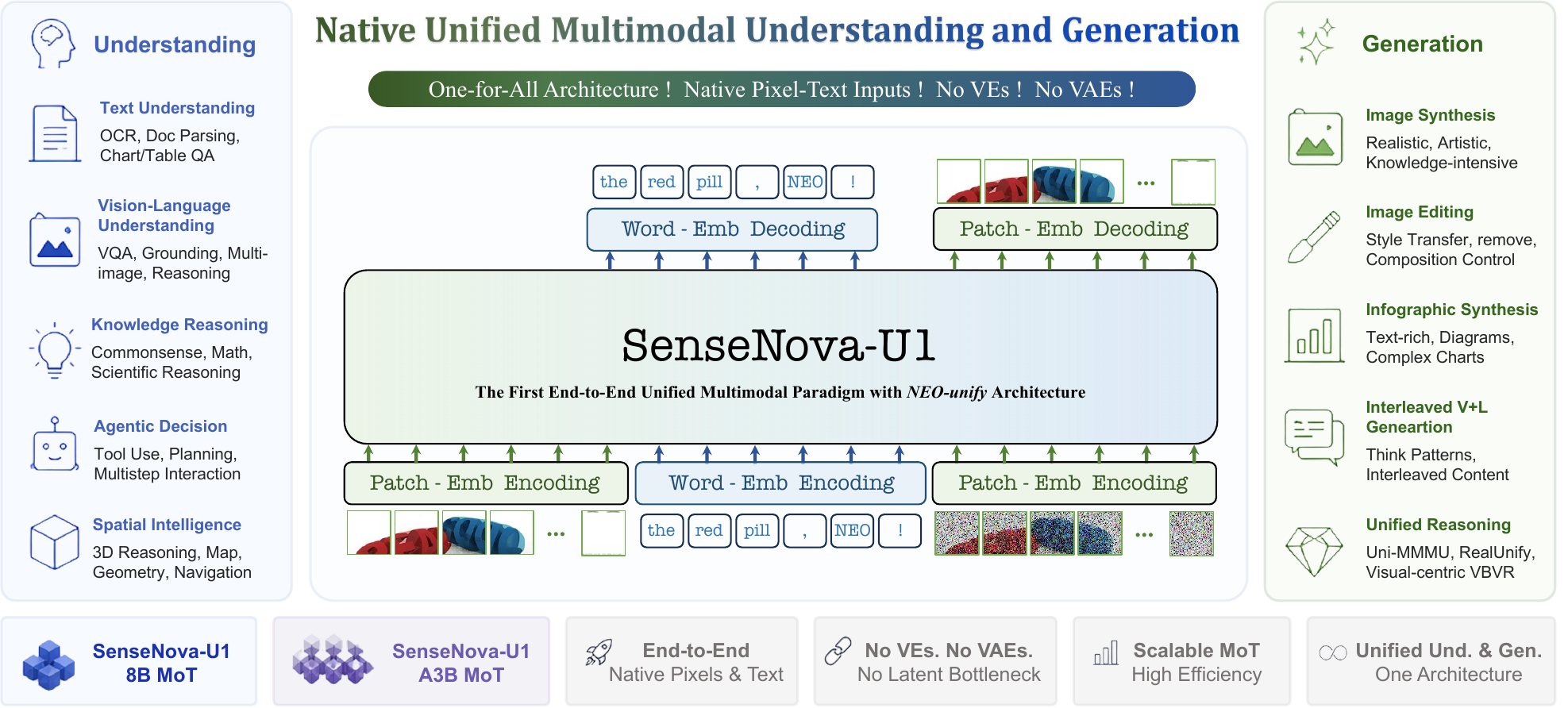}
\caption{
\textbf{Overview of SenseNova-U1.} 
With extremely lightweight encoding and decoding interfaces, SenseNova-U1 enables efficient and deeply correlated pixel-word correspondence within a single end-to-end architecture. As a native unified multimodal paradigm, it jointly supports diverse application scenarios, including perception, synthesis, and interleaved vision-language generation.
}
\label{fig:teaser_overview}
\end{figure*}

\section{Related Works}
\label{sec:related_works}

\subsection{Native Multimodal Models}

Recently, vision–language models (VLMs)~\cite{wang2025internvl3, Qwen3-VL, qwen35blog, kimik25, vteam2025glm45vglm41vthinkingversatilemultimodal, openai_gpt5_systemcard, gemini_3_pro_systemcard} have rapidly advanced multimodal understanding by coupling visual encoders (VEs) with large language models (LLMs), through either staged pretraining or joint optimization.
Despite their success, such designs inherit pretrained semantic biases and introduce additional complexity, along with inherent capacity trade-offs across components.
This has motivated a shift toward native multimodal backbones without VEs, as exemplified by Fuyu~\cite{VLM:Fuyu-8b} and EVE~\cite{VLM:EVE}.
Subsequent works push further by efficiently constructing visual perception while mitigating vision–language conflicts through distillation~\cite{VLM:EVE,VLM:BREEN,VLM:VoRA}, data mixing~\cite{VLM:SOLO,VLM:SAIL}, shared modules~\cite{VLM:HoVLE,VLM:HaploVL}, and modality decomposition~\cite{VLM:EVEv2,VLM:Mono-InternVL,VLM:Mono-InternVL-1.5}.
Notably, NEO~\cite{Diao2025NEO} advances this line by exploring a native pixel–word primitive, substantially narrowing the gap with leading modular VLMs over diverse understanding tasks.
For years, visual generation has been dominated by low-dimensional VAE or VQ-VAE latents~\cite{vae,vqvae}, with heavy compression limiting semantic expressivity under reconstruction-driven objectives.
Although recent efforts~\cite{vavae,REPA-E} enrich these latents with pretrained representations or auxiliary objectives, they remain fundamentally constrained by the compression bottleneck and fragmented training pipelines.
In parallel, emerging works~\cite{PixelFlow, DiP, yu2025pixeldit,li2025back} validate that direct pixel-space modeling can rival or even surpass latent diffusion, pointing toward a fundamentally new direction via fully end-to-end optimization from raw pixels.

\subsection{Native Multimodal Unified Models}

Early efforts to unify multimodal understanding and generation have largely converged on shared backbones, as exemplified by Show-o~\cite{xie2024show,xie2025show}, Janus~\cite{wu2024janus,ma2024janusflow,chen2025janus}, OmniGen~\cite{xiao2024omnigen,wu2025omnigen2}, and BAGEL~\cite{deng2025bagel}.
While these systems demonstrate that perception and synthesis can coexist within a single model, they remain split across fundamentally different tokenizers, diffusion heads, or decoupled pathways, reflecting a deeper mismatch between understanding and generation.
A complementary line of work shifts the focus to the visual interface itself, including shared discrete tokenizers~\cite{wu2024vila,QLIP,qu2025tokenflow,ma2025unitok,TokLIP} or continuous representation-based autoencoder~\cite{zheng2025diffusion,shi2025latent,yue2025uniflow,fan2025prism,liu2025tuna,AlignTok,tong2026beyond}. These approaches partially reconcile perception and synthesis, yet remain fundamentally constrained by intermediate representations, where semantic structure and visual fidelity must be traded against each other.

Native multimodal modeling is increasingly diverging along two distinct directions.
Discrete unified models~\cite{Chameleon,MOMA,wang2024emu3,cui2025emu35nativemultimodalmodels,MoT,li2025onecat,team2026longcat} recast multimodal learning as token-level autoregression, achieving architectural unification while sacrificing visual fidelity and expressivity under discrete tokenization.
In parallel, continuous native approaches pursue end-to-end modeling without explicit tokenizers or latent bottlenecks. 
NEO-unify~\cite{sensenova2026neounify} takes a first step toward this direction by learning directly from near-lossless inputs, achieving strong performance across diverse understanding and generation tasks. Tuna-2~\cite{tuna2} further demonstrates that pixel-space modeling can match latent-space methods, reinforcing the view that high-fidelity generation need not rely on compressed representations.
Notably, SenseNova-U1 builds on NEO-unify~\cite{sensenova2026neounify} by scaling this paradigm across data corpus, model capacity, and application scenarios, moving toward a truly unified foundation in which multimodal intelligence emerges natively.

\section{Methodology}
\label{sec:methodology}

\begin{figure}
    \centering
    \includegraphics[width=0.99\linewidth]{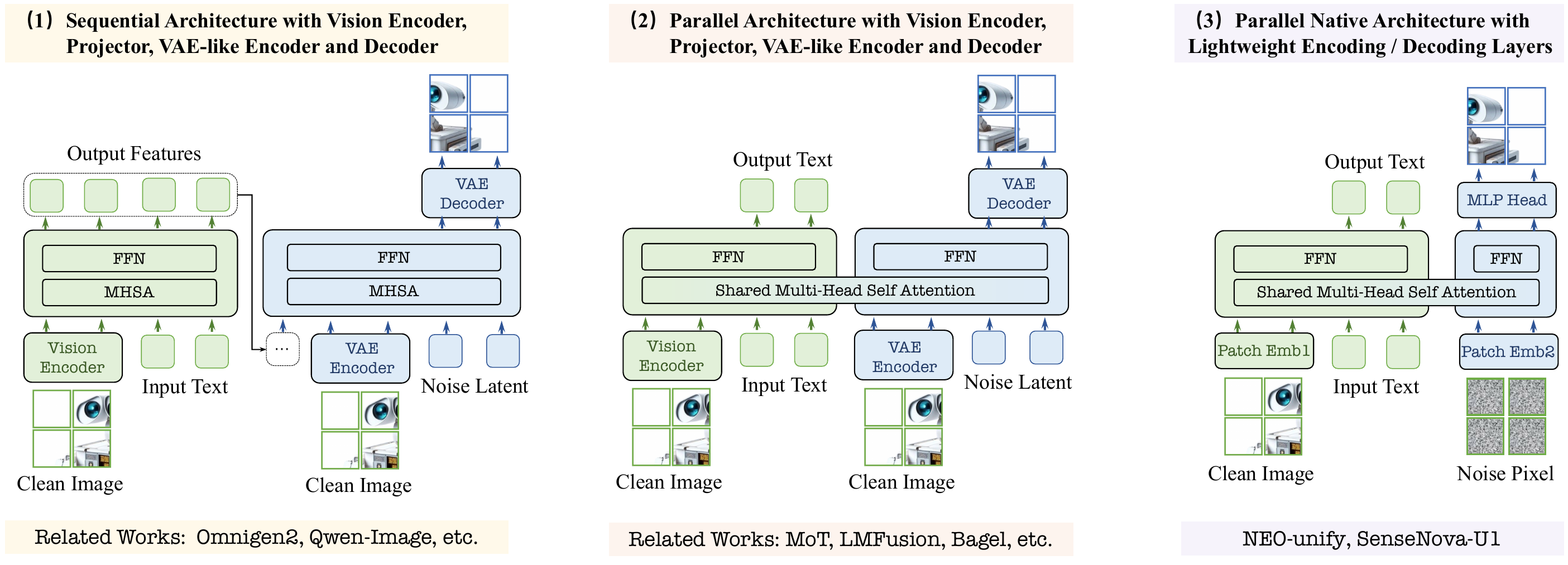}
    \caption{
    \textbf{SenseNova-U1 built on NEO-unify: one native paradigm for multimodal understanding and generation.} 
    SenseNova-U1 operates directly on native pixel and text inputs without relying on separate VEs or VAEs. The framework combines a near-lossless visual interface, implemented with two-layer convolutional encoding and MLP-like decoding layers, together with a native Mixture-of-Transformers (MoT) main architecture. Despite using a 32× compression ratio, it delivers strong performance across a broad range of understanding and generation tasks, while substantially simplifying system design and improving computational efficiency.}
    \label{fig:backbone}
\end{figure}

For years, multimodal models have relied on a vision encoder (VE) for perception and a variational autoencoder (VAE) for generation. Recent efforts attempt to unify these components through shared tokenizers, yet remain constrained by representational trade-offs.
SenseNova-U1 returns to first principles, introducing a native, unified, end-to-end framework that operates directly on pixels and words, eliminating reliance on pretrained encoder priors and the scaling limitations imposed by fixed representations.
The overall framework is illustrated in Figure~\ref{fig:backbone}.

\subsection{Near-Lossless Visual Interface}

\textbf{Patch Encoding Layer.} 
We follow NEO~\cite{Diao2025NEO} to construct lightweight patch encoding layers. Given an input image or noise, we map it into a sequence of visual tokens using two convolutional layers with GELU activation and 2D sinusoidal positional encoding. The convolutional strides are set to 16 and 2, so that each token corresponds to a 32 × 32 image patch. Two special \texttt{<img>} and \texttt{</img>} tokens are used to delimit visual content.
Besides, text words are encoded using the original tokenizer of the underlying language model without modification. After that, visual and textual tokens are projected into a shared embedding space and processed jointly within a unified backbone.

\textbf{Patch Decoding Layer.} 
The understanding stream uses a linear projection head to map tokens to the word vocabulary for text prediction. The generation stream directly predicts pixel patches via a multi-layer perceptron (MLP) head, bypassing deep diffusion heads and VAE decoders. This design enables fully end-to-end learning of the representation space, free from the inductive biases and representational constraints imposed by intermediate modules.

\noindent\textbf{Dynamic Noise Scale.}
Because the generation stream operates over varying resolutions, a naive unit-variance prior $\mathbf{z}_1 \sim \mathcal{N}(0, \mathbf{I})$ becomes mismatched to the signal scale, leading to inconsistent signal-to-noise ratios (SNRs) across resolutions at the same flow timestep.  
To address this, we introduce a resolution-adaptive noise scale $\sigma_{\!R}$. Let $N(H,W) = (H \cdot W)/32^2$ denote the number of generation tokens for an image of size $H \times W$, and let $N_0$ be a reference token count. We define:
$\sigma_{\!R}(H,W) = \sigma_0 \sqrt{N(H,W)/{N_0}},$
where $\sigma_0$ is a base noise scale.  
During training, terminal noise is sampled from a Gaussian distribution scaled by $\sigma_{\!R}$, which also initializes the flow ordinary differential equation (ODE) at inference.  
Intuitively, the square-root scaling preserves approximately constant per-token noise energy from low to high resolutions, ensuring a consistent SNR distribution for flow matching.

\noindent\textbf{Noise-Scale Conditioning.}
Since $\sigma_{\!R}$ varies with image resolutions, we explicitly feed it to the denoiser. 
We normalize the scale as $\bar\sigma = \sigma_{\!R}/\sigma_{\max} \in [0,1]$ and encode it using a dedicated sinusoidal MLP embedder $\mathrm{NSEmb}(\cdot)$. 
The resulting embedding is combined with the timestep embedding $\mathbf{\tau}_t$ to form the conditioning signal:
$\mathbf{s}_t = \mathbf{\tau}_t + \mathrm{NSEmb}\big(\bar\sigma(H, W)\big),
$
where $\mathbf{s}_t$ denotes the joint time and noise-scale conditioning applied to the input image tokens.

\subsection{Native Multimodal Unified Modeling}
\label{sec:mot}

\begin{table}[t]
\centering
\small
\setlength{\tabcolsep}{10pt}
\renewcommand{\arraystretch}{1.15}
\begin{tabular}{l|c|c}
\specialrule{1.2pt}{0pt}{0pt}
\textbf{Configuration} & \textbf{SenseNova-U1-8B-MoT} & \textbf{SenseNova-U1-A3B-MoT} \\
\midrule
Patch Size & 32 $\times$ 32 & 32 $\times$ 32 \\
Pre-Buffer & \ding{51} &\ding{55} \\
\# Num Layers & 42 & 48 \\
\# Num Heads (Q / KV) & 32 / 8 & 32 / 4 \\
Head Size (T / H / W) & 64 / 32 / 32 & 64 / 32 / 32 \\
Hidden Size & 4,096 & 2,048 \\
\midrule
\# Und / Gen Experts & 1 / 1 & 128 / 32 (A8) \\
\# Und / Gen Parameters & 8.2B / 8.2B & 30.0B / 8.2B (A3B) \\
\specialrule{1.2pt}{0pt}{0pt}
\end{tabular}
\caption{\textbf{Architectural configurations of SenseNova-U1 variants:} SenseNova-U1-8B-MoT $\&$ SenseNova-U1-A3B-MoT.}
\label{tab:model_config}
\end{table}

\textbf{Improved Native Primitive.} 
We refine the native VLM primitive from NEO~\cite{Diao2025NEO} as the base transformer block. Its native rotary position embedding (Native RoPE) unifies temporal and spatial encoding within a single representation. Text tokens evolve along the temporal axis $T$ with $H = W = 0$, while image tokens additionally carry spatial indices along height $H$ and width $W$.
The new design \textit{reallocates} pretrained LLM head dimensions across the $T$, $H$, and $W$ axes, each associated with independent frequency bases and incurring no additional parameters. 
It is applied to the Query and Key projections, along with their corresponding normalizations, all initialized from the understanding backbone.
Besides, we maintain native multimodal attention that jointly supports language and vision modeling.

\textbf{Native Mixture-of-Transformers.} 
At the core of SenseNova-U1 is a native Mixture-of-Transformers (MoT) backbone that unifies understanding and generation within a monolithic framework. The understanding stream processes clean image and text inputs, while the generation stream operates on noise-conditioned inputs.
All modalities are represented within a single sequence and processed under a shared self-attention mechanism, enabling perception and synthesis to interact natively at every layer.
Here, text tokens attend causally to preceding tokens only. Image tokens within the same block attend bidirectionally to one another while remaining causally conditioned on all preceding context. Noise tokens within each image block also attend bidirectionally, with full access to clean inputs, whereas clean tokens are prevented from attending to any noise tokens.
Crucially, we adopt full parameter decoupling between the two streams, with separate projections, normalizations, and feedforward blocks dynamically routed by token type at each layer.

\noindent\textbf{Model Variants.}
SenseNova-U1 is instantiated at two scales (detailed model configurations are provided in Table~\ref{tab:model_config}):

\begin{itemize}
    \item \textbf{SenseNova-U1-8B-MoT.} 
    The shallow Pre-Buffer layers map raw pixel and text inputs into a unified representation, while the Post-LLM layers retain the linguistic proficiency and reasoning capabilities of a pretrained LLM. Besides, both streams are instantiated as dense 8B networks in a symmetric parallel configuration.

    \item \textbf{SenseNova-U1-A3B-MoT.} 
    To scale efficiently, we extend the MoT framework with stream-wise mixture-of-experts (MoE) without Pre-Buffer layers. The understanding stream employs 128 experts with a total of 30B parameters, while the generation stream uses 32 experts totaling 8B parameters. A top-$k$ routing strategy activates 8 experts per token in each stream, resulting in approximately 3B active parameters during inference.
\end{itemize}

\subsection{Joint Training Objective}

SenseNova-U1 is optimized end-to-end with text and visual generation objectives weighted by $\lambda_1$ and $\lambda_2$:
\begin{equation}
    \mathcal{L}_\text{total} = \lambda_1 \mathcal{L}_\text{Und} + \lambda_2 \mathcal{L}_\text{Gen}
\label{eq:total_loss}
\end{equation}

\noindent\textbf{Autoregressive Text Loss.}
For understanding tasks, we employ standard next-token prediction as follows:
\begin{equation}
    \mathcal{L}_\text{Und} = -\frac{1}{N}\sum_{n=1}^{N} \log p_\theta(x_n \mid x_{<n}, \mathbf{c})
\end{equation}
where $x_n$ denotes the $n$-th text token, $x_{<n}$ the preceding tokens, and $\mathbf{c}$ the multimodal context tokens.

\noindent\textbf{Pixel-Space Flow Matching.}
For visual generation, we follow the former JiT~\cite{li2025back}
with $\mathbf{x}$-predict and $\mathbf{v}$-loss, operating directly in the pixel-level space. Given a clean image
$\mathbf{x}\in\mathbb{R}^{3\times H\times W}$ and a Gaussian sample $\boldsymbol\epsilon\sim\mathcal{N}(0,\mathbf{I})$, we form
the noisy sample along the rectified-flow interpolant, formulated as follows:
\begin{equation}
\mathbf{z}_t=t\,\mathbf{x}+(1-t)\,\sigma_{\!R}\,\boldsymbol\epsilon,
\qquad t\in[0,1],
\label{eq:interp}
\end{equation}
where $t=0$ corresponds to pure
noise ($\mathbf{z}_0=\sigma_{\!R}\boldsymbol\epsilon$), and $t=1$
corresponds to the clean image ($\mathbf{z}_1=\mathbf{x}$). 
Note that $\sigma_{\!R}$ denotes the resolution-adaptive noise scale. The unified framework directly regresses the clean signal $\hat{\mathbf{x}}_\theta$, which is then converted into a velocity term for $\mathbf{v}$-loss computation as follows:
\begin{equation}
\mathbf{v}_\theta(\mathbf{z}_t,t)
=\frac{\hat{\mathbf{x}}_\theta(\mathbf{z}_t,t,\mathbf{s}_t)-\mathbf{z}_t}{1-t},
\label{eq:vpred}
\end{equation}
where $\mathbf{s}_t$ is joint time-and-noise-scale conditioning. We adopt mean squared error (MSE) for velocity-space loss,
\begin{equation}
\mathcal{L}_{\text{Gen}}
=\mathbb{E}_{t,\mathbf{x},\boldsymbol\epsilon,(H,W)}\!\Bigl[\,
\bigl\|\mathbf{v}_\theta(\mathbf{z}_t,t)-\mathbf{v}^\star\bigr\|_2^{\,2}\,\Bigr],
\qquad
\mathbf{v}^\star=\frac{\mathbf{x}-\mathbf{z}_t}{1-t}.
\label{eq:lgen}
\end{equation}

\noindent\textbf{Classifier-Free Guidance.}
For generation tasks, including text-to-image synthesis, image editing, and interleaved image–text generation, we adopt a unified classifier-free guidance formulation that independently modulates the influence of textual and visual conditions. Let $\mathbf{c}_{\mathrm{txt}}$ denote the text condition and $\mathbf{c}_{\mathrm{img}}$ the visual context.
During training, we randomly drop the text condition with probability $10\%$, and drop both text and image conditions with an additional probability of $10\%$, enabling the model to learn conditional, image-only, and unconditional generation within a single framework.
During inference, the guided score is formulated as:
\begin{equation}
\begin{aligned}
\nabla_{\mathbf{x}} \log p(\mathbf{x}\mid c_{\mathrm{img}}, c_{\mathrm{txt}})
=&\ \gamma \Big(
\nabla_{\mathbf{x}} \log p(\mathbf{x}\mid c_{\mathrm{img}}, c_{\mathrm{txt}})
-
\nabla_{\mathbf{x}} \log p(\mathbf{x}\mid c_{\mathrm{img}})
\Big) \\
&+ \gamma_{\mathrm{img}} \Big(
\nabla_{\mathbf{x}} \log p(\mathbf{x}\mid c_{\mathrm{img}})
-
\nabla_{\mathbf{x}} \log p(\mathbf{x})
\Big)
+ \nabla_{\mathbf{x}} \log p(\mathbf{x}) .
\end{aligned}
\label{eq:generation_cfg}
\end{equation}
Here, $\gamma$ controls text guidance and $\gamma_{\mathrm{img}}$ controls image-context guidance. Empirically, $\gamma=4$ and $\gamma_{\mathrm{img}}=1$ consistently yield the best performance across X2I tasks, suggesting that explicit image-context guidance plays a comparatively minor role.
This observation implies that the model already captures visual conditioning effectively, while stronger guidance is primarily needed to enforce textual alignment. In practice, this guidance is applied to the predicted flow velocity used for generation. Note that we apply a timestep shift of $3.0$ and global CFG renormalization strategies.

\subsection{Training Procedure}

SenseNova-U1 is trained via progressive stages in Table~\ref{tab:training_stages} that incrementally build native multimodal capabilities.

\noindent\textbf{Stage 1: Understanding Warmup.}
We initialize from a pretrained NEO~\cite{Diao2025NEO} and perform two efficiency-oriented adaptations: an attention-fusion phase that simplifies original QK projections and normalization, followed by a full-model continuation phase that re-equilibrates the network under the enhanced attention modules.

\textit{(i) Attention-Fusion Phase.} 
We unify NEO’s QK projections and normalization across the temporal and spatial axes into a single shared set, halving the QK parameter footprint while preserving the native RoPE multi-axis structure and maintaining separate frequency scaling for temporal (rope theta = 5,000,000) and spatial dimensions (rope theta = 10,000).
To mitigate the short-term performance drop, we freeze the rest of the network and fine-tune only the attention layers, including Q, K, V, and output projections as well as QK normalization, until the model recovers its pre-fusion accuracy. Training data for this phase is drawn from an updated mid-training corpus described in Sec.~\ref{sec:undestanding_data_organization}.

\textit{(ii) Full-Model Continuation Phase.}
We then unfreeze the entire understanding branch and continue training on the same updated mid-training corpus, using a learning rate of $2 \times 10^{-5}$. The resulting model forms the understanding backbone of SenseNova-U1, providing rich contextual conditioning for subsequent generation phases.

\noindent\textbf{Stage 2: Generation Pre-Training.}
With the understanding branch frozen, we pretrain the generation branch on text-to-image data. It learns to synthesize pixel patches directly via pixel-space flow matching, conditioned on the semantic context from the frozen understanding branch. This stage establishes a stable generative foundation before joint optimization.
Specifically, we conduct the overall generation pre-training processes in three phases. 

In Phase I, we train on text-to-image data with resolutions ranging from $256 \times 256$ to $1024 \times 1024$ pixels, resizing images larger than $512 \times 512$ to $512 \times 512$ while preserving aspect ratios. This phase runs for 120K steps with a constant learning rate of $2 \times 10^{-4}$. 
In Phase II, we continue training on samples with resolution no smaller than $512^2$ pixels, resizing images larger than $2048 \times 2048$ to $2048 \times 2048$. This phase lasts for 60K steps, using a learning rate of $1 \times 10^{-4}$.
In Phase III, we introduce image-editing, reasoning, and interleaved image–text generation data for an additional 120K steps, progressively expanding the model’s generative capabilities over diverse application scenarios. The entire training uses a cosine learning-rate schedule that decays from $1 \times 10^{-4}$ to $2 \times 10^{-5}$.
For the A3B variant, we apply a MoE balance loss coefficient of $5 \times 10^{-3}$ to ensure balanced expert utilization in the generation branch.

\begin{table*}[t]
\centering
\small
\setlength{\tabcolsep}{4.5pt}
\renewcommand{\arraystretch}{1.15}
\resizebox{\textwidth}{!}{
\begin{tabular}{l|c|c|c|c|c|c}
\toprule
& \textbf{Stage 1 :} & \multicolumn{3}{c|}{\textbf{Stage 2 : Generation Pre-Training}} & \textbf{Stage 3 :} & \textbf{Stage 4 :} \\
& \textbf{Understanding Warmup} & \textbf{Phase I} & \textbf{Phase II} & \textbf{Phase III} & \textbf{Unified Mid-Training} & \textbf{Unified SFT} \\
\midrule
\multicolumn{6}{l}{\textbf{Hyperparameters}} \\
Peak learning rate
& $2\times10^{-5}$
& $2\times10^{-4}$
& $1\times10^{-4}$
& $1\times10^{-4}$
& $2\times10^{-5}$
& $2\times10^{-5}$ \\

Min learning rate
& --
& --
& --
& $2\times10^{-5}$
& --
& $0$ \\

LR scheduler
& Constant
& Constant
& Constant
& Cosine decay
& Constant
& Cosine decay \\

Optimizer 
& 
\multicolumn{6}{c}{AdamW ($\beta_1=0.9,\; \beta_2=0.95,\; \epsilon=10^{-8}$)} 
\\

Weight decay
& 0.0
& 0.0
& 0.0
& 0.0
& 0.0
& 0.0 \\

Gradient norm clip
& 1.0
& 1.0
& 1.0
& 1.0
& 1.0
& 1.0 \\

EMA ratio
& --
& 0.9999
& 0.9999
& 0.9999
& 0.999
& 0.999 \\

Training steps
& 120K
& 120K
& 60K
& 120K
& 84K
& 9K \\

Warmup Steps
& --
& 2000
& 2000
& 2000
& 2000
& 100 \\

Loss weight (CE : MSE)
& $1:0$
& $0:1$
& $0:1$
& $0:1$
& $0.1:1$
& $0.1:1$ \\

Und resolution
& $256^2 \rightarrow 4096^2$
& --
& --
& --
& $256^2 \rightarrow 4096^2$
& $256^2 \rightarrow 4096^2$ \\

Gen resolution
& --
& $256^2 \rightarrow 512^2$
& $512^2 \rightarrow 2048^2$
& $512^2 \rightarrow 2048^2$
& $512^2 \rightarrow 2048^2$
& $512^2 \rightarrow 2048^2$ \\

Seq Length

& 32768
& 8192
& 16384
& 16384
& 32768
& 32768 \\

Time shift 
& -- 
& \multicolumn{5}{c}{$\mu=-0.8, \sigma=0.8$ in the logit-normal
$t$-sampler} \\

Noise scale 
& -- 
& 
\multicolumn{5}{c}{$\sigma_0\sqrt{N/N_0}$, $\sigma_0=1,\; N_0=64\; (\sigma\in[1,8]\ \text{for}\ N\in[64,4096])$} 
\\

\# Training tokens
& 0.75T
& 0.25T
& 0.25T
& 0.88T
& 1.19T
& 0.13T \\

\midrule
\multicolumn{6}{l}{\textbf{Data sampling ratio}} \\
Understanding data
& 1.00
& 0.00
& 0.00
& 0.00
& 0.33
& 0.33 \\

Generation data
& 0.00
& 1.00
& 1.00
& 0.56
& 0.37
& 0.37 \\

Editing data
& 0.00
& 0.00
& 0.00
& 0.37
& 0.24
& 0.24 \\

Interleave data
& 0.00
& 0.00
& 0.00
& 0.07
& 0.06
& 0.06 \\
\bottomrule
\end{tabular}
}
\caption{\textbf{Training recipe of SenseNova-U1 from Stage 1 to Stage 4.} Stage 2 is divided into three phases for generation pre-training.}
\label{tab:training_stages}
\end{table*}

\noindent\textbf{Stage 3: Unified Mid-Training.}
Both branches are jointly trained end-to-end on a curated mixture of understanding and generation data, allowing the MoT backbone to develop coherent shared representations across various tasks.

The training mixture includes text-only \& multimodal understanding, text-to-image generation, image editing, and interleaved image–text data, with sampling ratios of $0.33:0.37:0.24:0.06$, respectively. We train the full model framework for 84K steps (nearly converges within 40K steps with $<$80\text{M} data) with a constant learning rate of $2 \times 10^{-5}$.
For joint optimization, we set the loss weights in Eq.(\ref{eq:total_loss}) to $\lambda_1 = 0.1$ and $\lambda_2 = 1.0$. Specifically for the A3B variant, we apply a MoE balance loss coefficient of $1 \times 10^{-3}$ to both the generation and understanding branches.

\noindent\textbf{Stage 4: Unified Supervised Fine-Tuning.}
The full model is fine-tuned on high-quality, instruction-following data spanning both understanding and generation tasks, including multimodal dialogue, image generation, editing, and interleaved data. This stage sharpens instruction alignment and task-specific performance across modalities.

We use the same data mixture as in Stage 3, covering understanding, text-to-image generation, image editing, and interleaved image–text data. The full model is further fine-tuned for 9k steps with a cosine learning-rate schedule decaying from $2 \times 10^{-5}$ to 0. We retain the same loss weights as in Stage 3, with $\lambda_1 = 0.1$ and $\lambda_2 = 1.0$ in Eq.(\ref{eq:total_loss}).

\noindent\textbf{Stage 5: Post Training for T2I Generation.}
We present the post-training recipe for SenseNova-U1 via an initial round of text-to-image generation training, which leverages reinforcement learning (RL) following Flow-GRPO~\cite{liu2025flow} to improve generation quality, and employs Distribution Matching Distillation~\cite{dmd2} to enhance efficiency.

\noindent
\paragraph{\textbf{Dynamic Resolution Warmup.}} Since different resolutions exhibit significant reward variance, we introduce a warmup strategy to improve training stability. The candidate
resolution set is constructed from aspect ratios
$\{1{:}1,16{:}9,9{:}16,3{:}2,2{:}3\}$ and target image areas
$\{1536^2,2048^2\}$, each assigned a base sampling probability $p_i$. We assign each resolution a difficulty score $d_i \in [0,1]$ based on its aspect ratio and pixel count, and gate the sampling probabilities as:
\begin{equation}
    \hat{p}_i \;=\; p_i \cdot \mathrm{clamp}\!\left(\frac{\min(e / E_{\text{warm}},\, 1) - d_i}{\delta} + 1,\; 0,\; 1\right),
\end{equation}
where $e$ is current epoch, $E_{\text{warm}}$ is warmup duration, and $\delta = 0.3$ is a smoothing margin. 
The gated probabilities are then renormalized for sampling, starting from easier configurations and progressively incorporating more challenging ones.

\noindent
\paragraph{\textbf{Reward Function Design.}} We employ three reward components in our RL training pipeline as follows:

\begin{itemize}
\item \textbf{Text Rendering Reward.} 
For text-rendering tasks, we use PaddleOCR~\cite{cui2025paddleocr30technicalreport} to extract text $\hat{T}$ from generated images and compare it with the ground-truth $T^*$. The reward is based on Intersection-over-Union (IoU): 
\begin{equation}
    r_{\text{ocr}} = \frac{|\mathcal{C}(\hat{T}) \cap \mathcal{C}(T^*)|}{ |\mathcal{C}(\hat{T}) \cup \mathcal{C}(T^*)|},
\end{equation}
where $\mathcal{C}(\cdot)$ denotes the Counter of texts, and $\cap$, $\cup$ represent multi-set intersection and union, respectively.

\item \textbf{Style Following Reward.}
For prompts with explicit style constraints, we use a VLM judge~\cite{gemini_3_pro_systemcard} to assess whether the generated image follows the specified style. 
The judge assigns a discrete score $s \in \{1,2,3,4\}$, linearly mapped to a style reward $r_{\mathrm{sty}} \in [0,1]$, where 0 indicates a complete mismatch and 1 denotes a perfect match.

\item \textbf{Aesthetic Reward.} For aesthetic and preference alignment, we use Human Preference Score (HPSv3)~\cite{ma2025hpsv3widespectrumhumanpreference} as an aesthetic quality reward. Given a generated image $\mathbf{x}$ and its prompt $p$, the aesthetic reward $r_{\mathrm{aes}} = H_{\mathrm{v3}}(\mathbf{x},p),$
where $r_{\mathrm{aes}}$ denotes the image–text preference score and $H_{\mathrm{v3}}$ is its corresponding scorer.
\end{itemize}

\noindent
\paragraph{\textit{(i) Text Rendering RL Training.}}
The training is conducted using text-rendering prompts in both English and Chinese, with the text rendering reward $r_{\text{ocr}}$ serving as the optimization signal. The model is trained for 600 epochs with a learning rate of $1 \times 10^{-5}$ and a KL coefficient $\beta = 0.01$. In each epoch, we sample $N = 48$ prompts, and for each prompt generate $K = 16$ images (i.e., $48 \times 16$ samples in total) using 10-step flow matching, with a guidance scale of 4.0 and a noise level of 0.7. A dynamic-resolution warmup is applied for the first $E_{\text{warm}} = 200$ epochs.

\noindent
\paragraph{\textit{(ii) Unified General RL Training.}}
This stage further improves image generation quality through interleaved multi-reward training: text rendering, style following, and general visual quality. 
To balance these heterogeneous objectives, we organize the training data into two reward groups and adopt an interleaved training strategy, where the active reward group alternates every training epoch. The two reward groups are defined as follows:

\begin{itemize}
    \item \textbf{\textit{Group 1: Text Rendering \& Style Following.}} This group uses text-rendering prompts with style constraints. The composite reward combines text accuracy and overall style, following:
    \begin{equation}
        r = r_{\text{ocr}} + \lambda_{\text{sty}} \cdot r_{\text{sty}},
    \end{equation}
    where $\lambda_{\text{sty}}$ controls the relative weight of the style following reward.

    \item \textbf{\textit{Group 2: Human Preference and Aesthetics.}} This group uses general image generation prompts and adopts $r_{\text{aes}}$ to assess aesthetics. Notably, it tends to favor darker backgrounds, with limited benefit to OCR performance.
\end{itemize}  

We interleave the two reward groups at every epoch. 
The coefficient $\lambda_{\text{sty}}$ is set to 0.5. 
The 8B variant is trained for 1,600 epochs, with all other hyperparameters consistent with the previous stage. The A3B variant is trained for 200 epochs and leaves room for further improvement.
Here, we freeze the understanding branch, the last three transformer layers, and the MLP head of the generation branch to mitigate grid artifacts. This issue likely arises because the final FFN layer and the MLP head model disjoint $32 \times 32$ pixel patches independently. \textit{A promising direction for future work is to replace the MLP head with PixelShuffle modules followed by two convolutional layers to further alleviate this issue.}

\noindent
\paragraph{\textbf{Stage 6: CFG \& Step Distillation.}}
We employ distribution matching distillation (DMD2)~\cite{dmd2} to reduce the number of function evaluations (NFE) for image synthesis from 100 to 8 for impressive efficiency.
It involves three models: a generator $G$ to be distilled, a fake flow model $F$ that estimates the score of the evolving generative distribution, and a teacher $T$ that models the score of the target data distribution. All three are initialized from the teacher model.

During distillation, only the generation branch is optimized, including the MoT parameters as well as the patch encoding and decoding layers.
The distillation process is performed in a unified setting, where text-to-image, editing, and interleaved data are jointly used for training. For editing and interleaved generation, backward simulation uses ground-truth images as references.
We follow the hyperparameter settings of Phased DMD~\cite{phased_dmd}. The generator $G$ is updated once every five updates of $F$. $G$ is optimized with AdamW optimizer at a learning rate of $2 \times 10^{-6}$, betas of $(0.0, 0.999)$, and weight decay of 0.01, while $F$ uses a learning rate of $4 \times 10^{-7}$ with the same optimizer settings. Backward simulation employs an Euler solver with a timestep shift of 3.0, and the CFG scale is set to 4.0.

\subsection{Inference Infrastructure}
\label{sec:inference_infrastructure}

Although SenseNova-U1 is exposed to users as a unified multimodal model, its
understanding and generation pathways have different inference characteristics.
The understanding path is dominated by multimodal prefill, autoregressive text
decoding, streaming, batching, and control-flow management, while the image
generation path is dominated by iterative pixel-space denoising with different
parallelism and memory-access patterns. Serving both paths in a single
monolithic runtime would unnecessarily couple their scheduling policies,
parallelization strategies, and resource budgets.

\begin{figure}[t]
    \centering
    \includegraphics[width=0.9\linewidth]{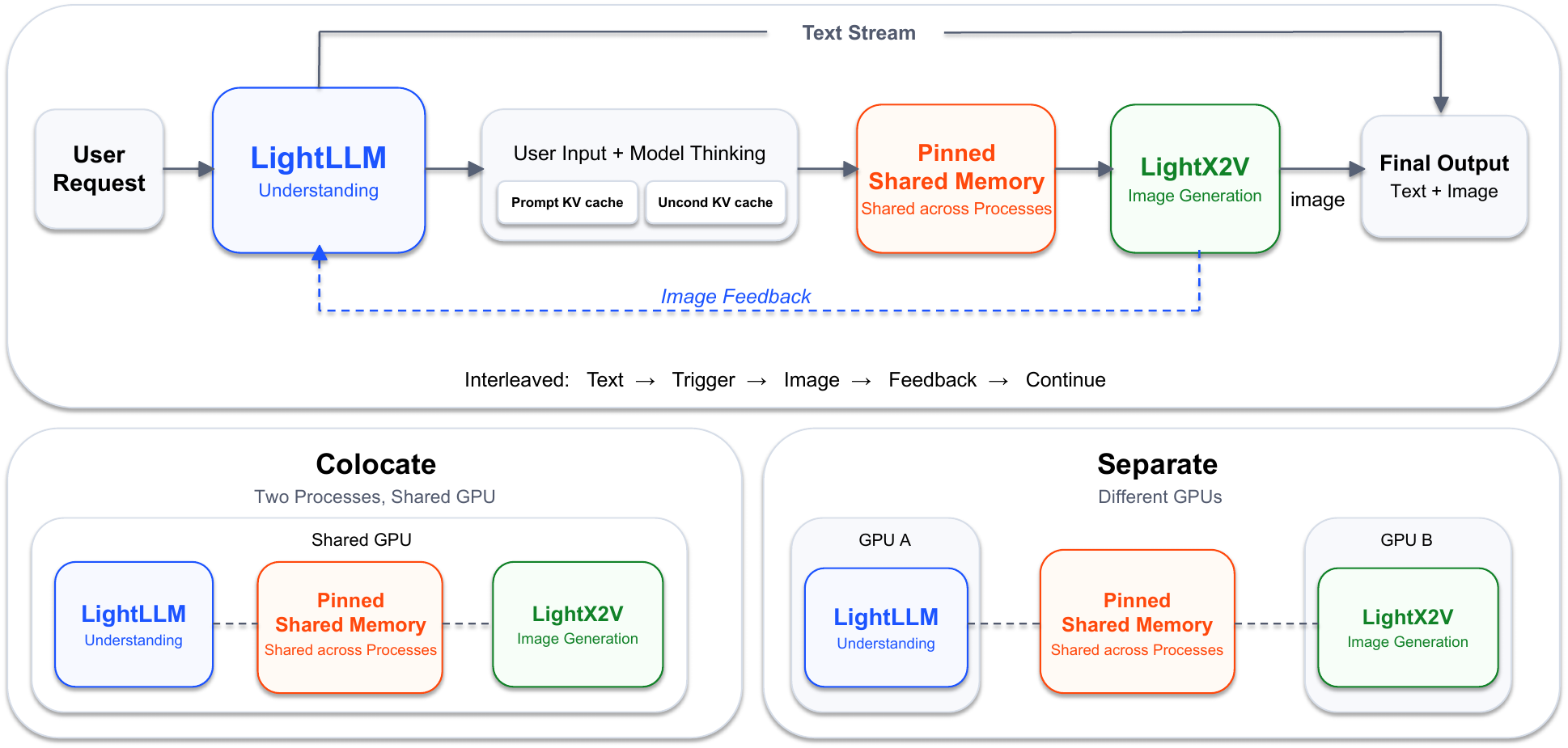}
    \caption{\textbf{Disaggregated inference architecture of SenseNova-U1.}
    \textsc{LightLLM} serves multimodal understanding, text streaming, and
    control flow, while \textsc{LightX2V} serves image generation. The two
    engines exchange generation state through pinned shared memory, enabling
    independent scheduling, parallelism, and resource allocation.}
    \label{fig:lightllm_x2v}
\end{figure}

\textbf{Disaggregated Deployment.}
We adopt a disaggregated inference
architecture using two specialized open-source engines:
\textsc{LightLLM}~\cite{gong2025pastfuture,lightllm} for multimodal
understanding, text streaming, and request orchestration, and
\textsc{LightX2V}~\cite{lightx2v} for image generation. The two engines exchange generation state through pinned shared memory and optimized transfer kernels, preserving a unified API abstraction while allowing each execution path to be independently optimized.

This design brings three practical benefits. First, it enables different parallelization strategies for different workloads: the understanding engine uses LLM-oriented Tensor Parallelism (TP), while the generation engine uses diffusion-oriented strategies, e.g., Classifier-Free Guidance Parallelism
(CFG Parallelism) or Sequence Parallelism (SP). Second, it supports independent resource allocation, including separate GPU groups, memory budgets, and batching policies. Third, it improves operational isolation, so text-heavy and image-heavy traffic can be scaled, profiled, and tuned independently.

The infrastructure supports both \emph{separate} and \emph{colocate} deployments. In separate mode, \textsc{LightLLM} and \textsc{LightX2V} run on distinct GPU groups, which is preferred in production because it provides clear bottleneck attribution and independent scaling. In colocate mode, the two
engines run as separate processes on the same GPU group, which is useful for lightweight validation, smaller hardware configurations, or deployment scenarios where the image generation workload is substantially higher than the understanding workload. For $2048{\times}2048$ image generation with SenseNova-U1-8B-MoT, both modes support a $\mathrm{TP2}{+}\mathrm{CFG2}$ configuration. In separate mode, the per-step latencies on 5090 and L40S GPUs are $0.415$ and $0.443$ seconds, respectively. Because the generation-stage key-value cache is provided by the understanding module, text-to-image generation and image editing share similar runtime characteristics.

\begin{figure}[t]
    \centering
    \includegraphics[width=0.9\linewidth]{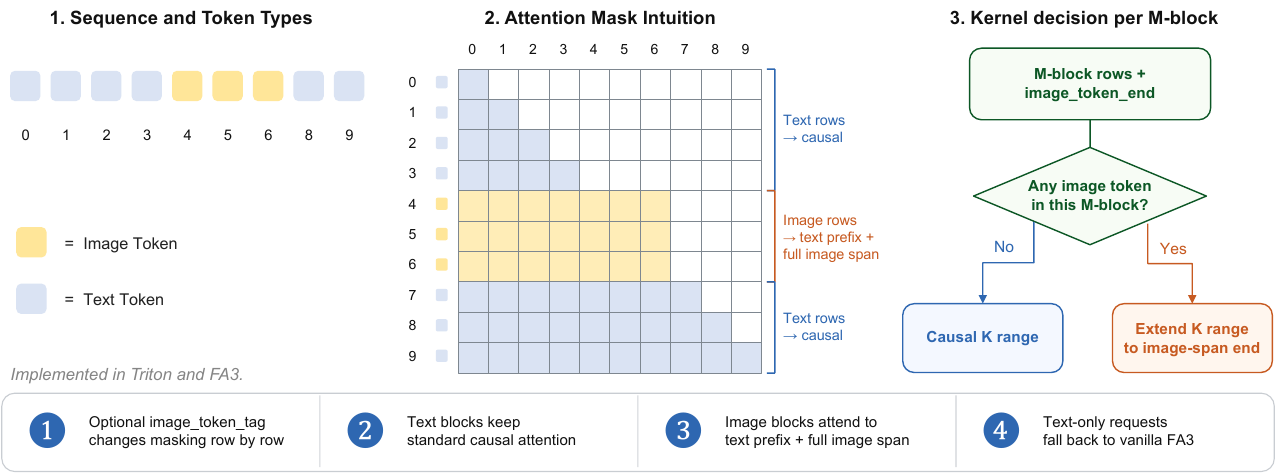}
    \caption{\textbf{Hybrid attention pattern for unified multimodal
    prefill.} Text rows follow the standard causal mask, while image rows can
    attend to the full preceding text prefix and the entire image span. The
    serving kernel preserves the causal fast path for pure-text blocks and
    only expands the key range for blocks that contain image tokens.}
    \label{fig:hybrid_attention}
\end{figure}

\textbf{Hybrid Attention Kernel.} A key systems challenge in unified multimodal prefill is the hybrid attention pattern: text rows remain causal, while image rows attend to the full text prefix and the full image span. This preserves standard autoregressive behavior for text tokens while allowing bidirectional interaction within image tokens.

To support this efficiently, we introduce an optional
\texttt{b\_image\_token\_end} in the attention kernel. The kernel makes the masking decision at the M-block level. If an M-block contains no image token, it keeps the standard causal key range. Otherwise, its key range is extended to the image-span end, so image rows can attend to the text prefix and the full image span. This design preserves the causal fast path for pure-text blocks and only introduces extra computation for blocks containing image tokens. We implement this mechanism in both a Triton kernel and a modified FlashAttention3 backend. The Triton version is easier to integrate, while the FlashAttention3 version provides higher throughput.
\section{Data Construction}

\subsection{Understanding Data Organization}
\label{sec:undestanding_data_organization}

\textbf{Pre-training Stage.}
The training corpus comprises large-scale web text, image-text pairs, and interleaved multimodal documents, organized into four categories: image-text pairs (32\%), captions (17\%), infographic understanding (14\%), and pure text (37\%). The data curation pipeline includes four stages: cross-source deduplication, content and safety filtering, image quality filtering, and CLIP-ratio-balanced re-captioning to ensure balanced alignment across the corpus.

\begin{figure}[t]
    \centering
    \includegraphics[width=0.95\linewidth]{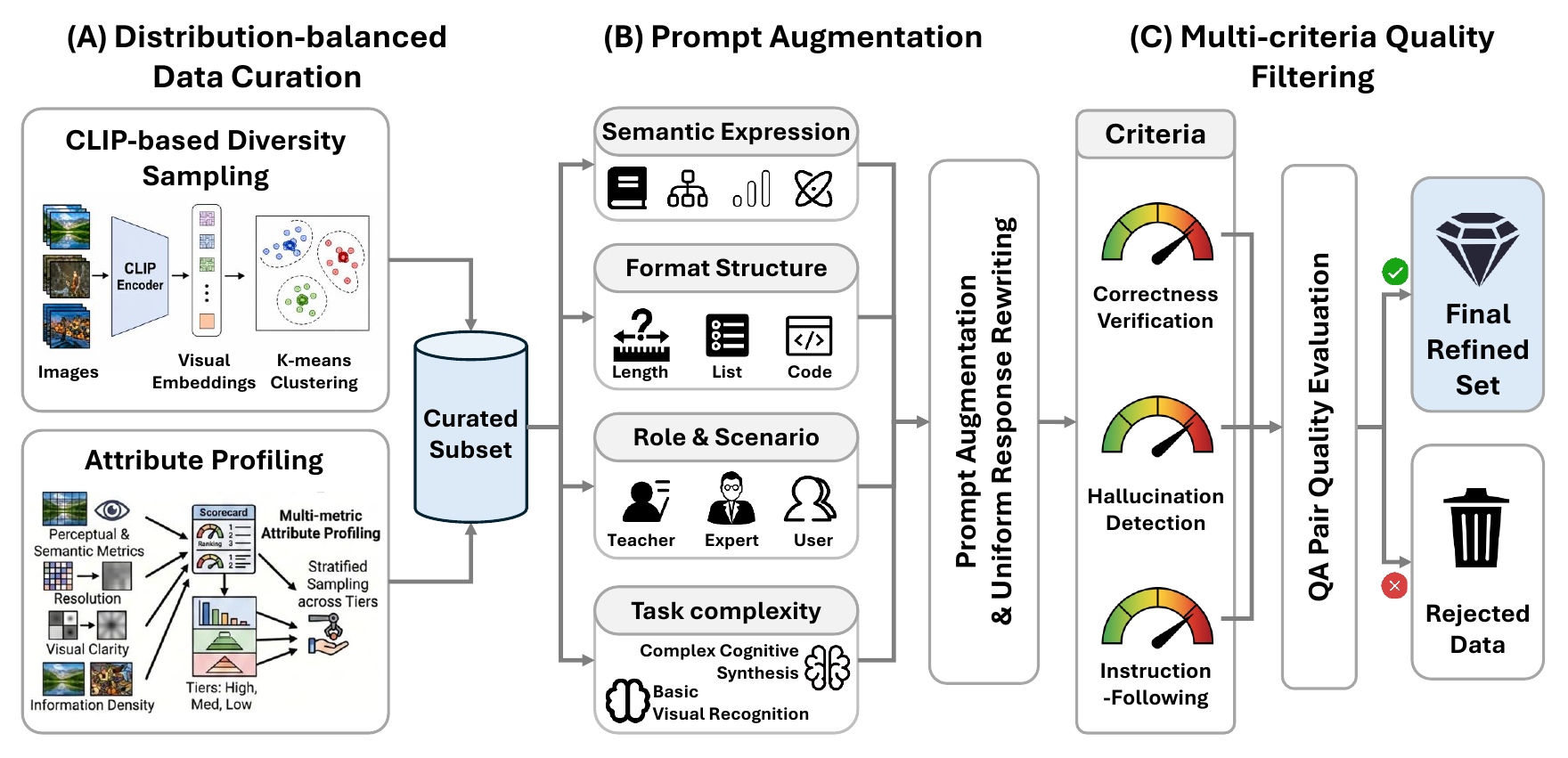}
    \caption{\textbf{Data processing pipeline for the understanding corpus.}
    Large-scale multimodal instruction data are curated across ten vertical domains through a systematic process consisting of distribution-balanced data curation, prompt augmentation, and multi-criteria quality filtering, producing a high-quality and diverse data corpus for the midtraining process.}
    \label{fig:midtrain_pipeline}
\end{figure}

\textbf{Mid-training Stage.}
This stage is primarily drawn from internal SenseNova V6.5 datasets, spanning four categories: General (39.2\%), Agent and Spatial (22.3\%), Knowledge Reasoning (19.3\%), and Pure Text (19.2\%). The General category is further divided into general visual question answering (26.6\%), multi-turn dialogue (26.4\%), captioning (20.3\%), OCR (18.6\%), and multi-image understanding (8.2\%), while Knowledge Reasoning includes knowledge-oriented (12.0\%) and reasoning-oriented (7.2\%) data. To ensure quality and diversity, we adopt a three-stage curation pipeline in Figure~\ref{fig:midtrain_pipeline}: distribution-balanced sampling, prompt augmentation, and multi-criteria filtering.

\noindent{(i) Distribution-Balanced Sampling.}
We adopt a two-stage process to extract a diverse subset from the initial pool. First, CLIP-based diversity sampling clusters visual embeddings via $K$-means and samples uniformly across clusters to improve long-tail coverage. This is followed by attribute profiling, which evaluates each sample along perceptual and semantic dimensions and applies stratified sampling to ensure balanced representation across attribute tiers.

\noindent{(ii) Prompt Augmentation.}
To improve the diversity and complexity of training instructions, we augment the initial prompts along four dimensions: semantic expression, format and structural constraints, role and scenario, and task complexity, ranging from perceptual recall to compositional reasoning. After prompt augmentation, all answers are uniformly regenerated to ensure consistent quality and stylistic coherence across the whole corpus.

\noindent{(iii) Multi-Criteria Filtering.}
To ensure dataset fidelity, we employ an automated model-based scoring pipeline to evaluate each question-answering (QA) pair across three dimensions: correctness verification against ground-truth annotations, hallucination detection to penalize visually unsupported fabrications, and instruction-following assessment to measure alignment with specified constraints such as formatting and persona.

\textbf{Supervised Fine-Tuning.}
The final SFT corpus is organized along fine-grained, capability-atomic dimensions to enable precise control over the supervision mixture. The distribution spans spatial
intelligence (${\sim}$15\%), general multimodal understanding (${\sim}$13\%),
reasoning (${\sim}$12\%), general NLP (${\sim}$11\%), OCR and document analysis
(${\sim}$11\%), agentic function calling (${\sim}$10\%), long-context
conversation (${\sim}$8\%), code (${\sim}$6\%), multi-turn dialogue
(${\sim}$4\%), complex compositional understanding (${\sim}$4\%), and
supplementary data covering additional capabilities for the remaining
proportion. 

Rather than recollecting data from scratch, we refine the midtraining candidate pool with a dual emphasis on quality and difficulty. For quality-oriented selection, we reuse the multi-criteria filtering framework from midtraining, scoring each candidate across visual fidelity, instruction clarity, response correctness, reasoning quality, and safety, while increasing the sampling proportion of high-scoring examples relative to midtraining. For difficulty-oriented reconstruction, we rebalance supervision along three axes: composing longer and structurally richer instances by concatenating short samples into long-context, multi-image, and multi-turn settings; applying rejection sampling for reasoning-intensive domains to retain examples in the intermediate difficulty regime where learning is most effective; and rewriting under-specified queries to inject explicit constraints on output format, stylistic attributes, and target granularity.

\begin{figure}[t]
    \centering
    \includegraphics[width=1\linewidth]{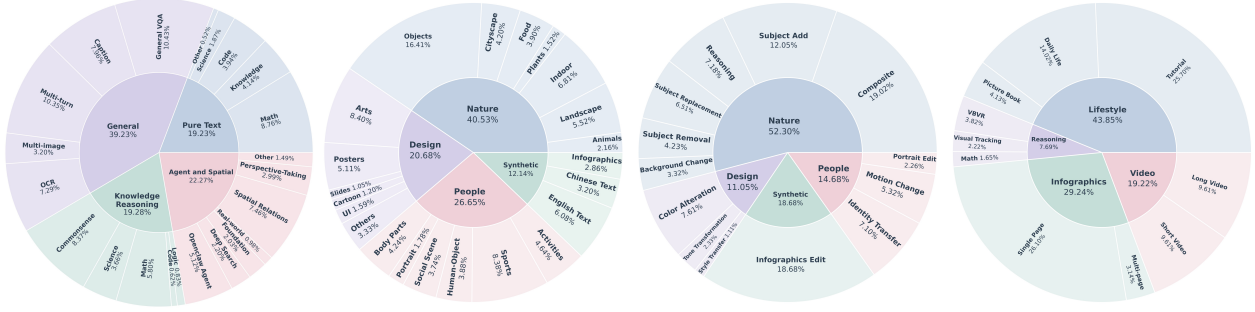}
    \caption{\textbf{Data distribution of SenseNova-U1's training corpus.}
    From left to right, the four sunburst charts depict the hierarchical composition of the \textit{Understanding}, \textit{Text-to-Image}, \textit{Editing}, and \textit{Interleaved} datasets. The inner ring shows top-level categories and their proportions, while the outer ring breaks them down into fine-grained subclasses. Together, they illustrate the diversity of natural images, synthetic content, and text-rich samples across all four capabilities.}
    \label{fig:data_distribution}
\end{figure}

\subsection{Generation Data Organization}
\label{sec:generation_data_organization}

\begin{figure}[t]
    \centering
    \includegraphics[width=0.95\textwidth]{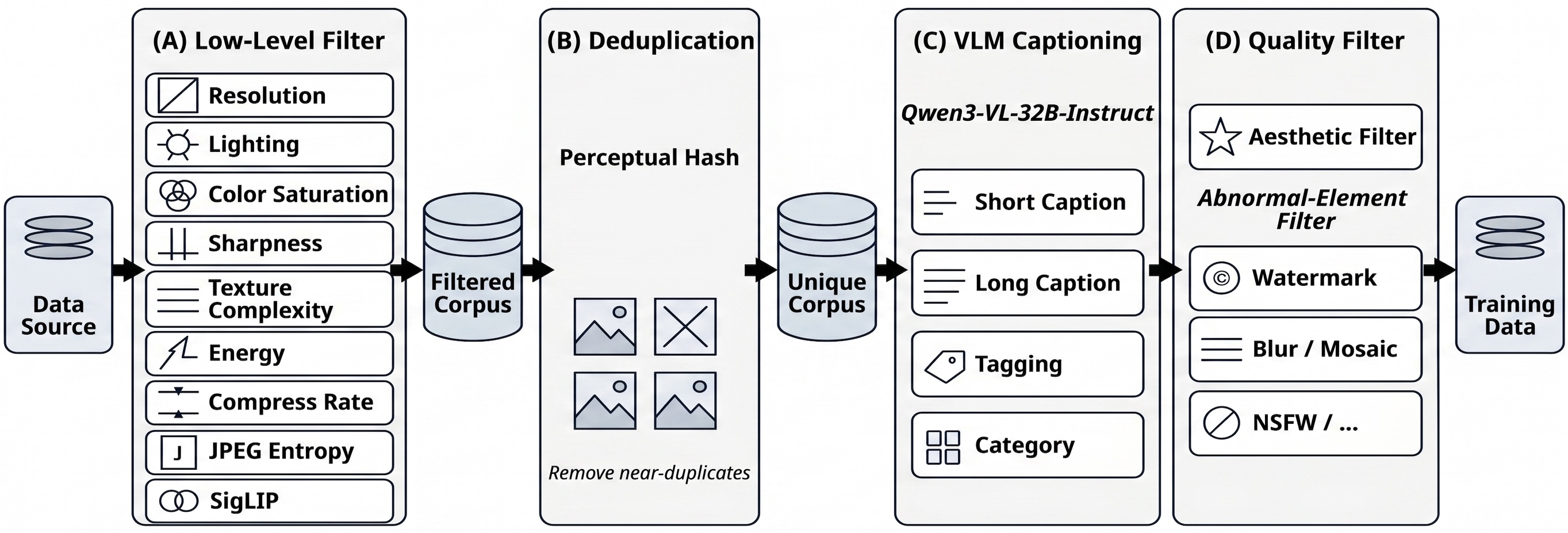}
    \caption{\textbf{Data processing pipeline for the generation corpus.}
    The same four-stage flow, \ie, low-level filtering, deduplication, VLM captioning, and quality filtering, is applied to T2I and image editing to ensure high-quality and diverse generation data.}
    \label{fig:data_pipeline}
\end{figure}

To balance broad coverage with high fidelity, we curate a large-scale generation corpus spanning text-to-image and image-editing data, organized into four domains: Nature, Design, People, and Synthetic. 
As shown in Figure~\ref{fig:data_distribution}, the distribution is carefully balanced across domains while preserving a pronounced long tail of natural, synthetic, and text-rich content.
All samples are processed through the unified pipeline in Figure~\ref{fig:data_pipeline}, combining low-level filtering, deduplication, VLM-based captioning, and quality-aware filtering to ensure a consistent standard of quality.

\textbf{Text-to-Image Data.}
The entire corpus is composed of Nature ($\sim$40.5\%), People ($\sim$26.7\%), and Design ($\sim$20.7\%), and further enriched with complex infographics and bilingual text rendering data, along with a long tail of fine-grained subclasses such as posters, charts, and cityscapes. This diversity provides broad visual coverage, structured layouts, and text-intensive scenarios, which foster strong visual priors, aesthetic quality, and robust text-rendering ability.

\textbf{Image Editing Data.}
The editing corpus is primarily sourced from web-scale data. For in-domain data, it reflects similar diversity at content and operation levels. For content level, natural scenes ($\sim$52.3\%) and human subjects ($\sim$14.7\%) dominate real-world coverage, with the remainder consisting of infographic and synthetic edits. For operation level, the data spans subject addition and removal, background and color changes, identity transfer, motion manipulation, portrait editing, compositing, and reasoning-driven transformations.
Beyond standard filtering, each editing pair is further validated by decomposing its instruction into dynamic objectives that specify what should change and what must remain unchanged. These objectives are jointly verified with a static physical-consistency constraint against source image.

\textbf{Interleaved Data.}
To strengthen interleaved reasoning and generation, we construct a compact vision–text corpus in which sequences alternate between text and images to form coherent multimodal narratives~\cite{cui2025emu35nativemultimodalmodels,xing2026wan}. The corpus spans four complementary categories: Video, Lifestyle, Infographics, and Reasoning domains, each targeting distinct capabilities, and is built under a unified pipeline of preprocessing, task-specific synthesis, and post-processing that jointly verifies text quality, image quality, image–text consistency, and trajectory-level correctness.
As shown in Figure~\ref{fig:data_distribution}, lifestyle data dominates at roughly 44\%, including tutorials (26\%), daily-life scenarios (14\%), and picture books (4\%). Infographics account for about 29\%, providing dense supervision for text-rich page synthesis. Video contributes around 19\%, capturing temporal continuity and world dynamics. Reasoning comprises approximately 8\% and represents the most reasoning-intensive subset, as each sample includes an explicit chain-of-thought trace.

\section{Experiments}

\subsection{Main Results}

\subsubsection{Image Understanding}
We evaluate SenseNova-U1 on various multimodal understanding and reasoning benchmarks, covering perception-centric understanding, multimodal reasoning, OCR recognition, visual-centric reasoning, and spatial intelligence. Note that we evaluate multimodal and pure-text understanding using \texttt{EvalScope}, following the LLM-as-a-judge paradigm with the \texttt{llm\_recall} strategy and \texttt{gpt-4o-mini-2024-07-18} as the judge model. All experiments use a standardized inference setup with $\texttt{temperature} = 0.6$, $\texttt{top\_p}=0.95$, $\texttt{top\_k}=20$, and $\texttt{repetition penalty}=1.00$. To support long-context multimodal reasoning, we set the maximum sequence length to $40{,}960$ tokens and the request timeout to $600$ seconds, and enable internal reasoning via \texttt{enable\_thinking:} \texttt{true}. 

\begin{table}[t]
\centering
\scriptsize
\caption{\textbf{Quantitative evaluation results on multimodal understanding benchmarks}. 
For spatial intelligence, we adopt EASI~\cite{easi2025} as the standard evaluation, using 32 input frames on VSI-Bench for all models. We observe that Qwen variants require 128-frame inputs to reach their best performance; we report these results separately and mark them with an asterisk (*).}
\label{tab:multimodal_understanding_results}
\vspace{1em}
\setlength{\tabcolsep}{1pt}
\begin{tabularx}{\textwidth}{@{}l*{3}{>{\centering\arraybackslash}X}|*{5}{>{\centering\arraybackslash}X}@{}}
\toprule
\textbf{Benchmark}
& \makecell{\textbf{SenseNova-U1}\\\textbf{8B-Think}}
& \makecell{\textbf{Qwen3VL}\\\textbf{8B-Think}}
& \makecell{\textbf{Qwen3.5}\\\textbf{9B}}
& \makecell{\textbf{SenseNova-U1}\\\textbf{30BA3B-Think}}
& \makecell{\textbf{Qwen3VL}\\\textbf{30BA3B-Think}}
& \makecell{\textbf{Qwen3.5}\\\textbf{35BA3B}}
& \makecell{\textbf{Gemma4}\\\textbf{26BA4B}}
& \makecell{\textbf{LongCat-Next}\\\textbf{68BA3B}} \\
\midrule

\multicolumn{9}{c}{\textit{STEM \& Reasoning}} \\
\midrule
MMMU~\cite{yue2024mmmu} & 74.78 & 74.10 & 78.40 & 80.55 & 76.00 & 81.40 & 76.56 & 70.60 \\
MMMU-Pro~\cite{yue2025mmmu} & 67.69 & 60.40 & 70.10 & 72.83 & 63.00 & 75.10 & 73.80 & 60.30 \\
MathVista$_{\text{mini}}$~\cite{lu2023mathvista} & 84.20 & 81.40 & 85.70 & 85.30 & 81.90 & 86.20 & 72.70 & 83.10 \\
MathVision~\cite{wang2024measuring} & 75.82 & 62.70 & 78.90 & 79.63 & 65.70 & 83.90 & 68.95 & 64.70 \\

\midrule
\multicolumn{9}{c}{\textit{General VQA}} \\
\midrule
MMBench-EN~\cite{liu2024mmbench} & 90.25 & 87.50 & 90.10 & 91.59 & 88.90 & 91.50 & 91.68 & -- \\
MMStar~\cite{chen2024we} & 78.27 & 75.30 & 79.70 & 80.92 & 75.50 & 91.90 & 76.93 & 69.30 \\

\midrule
\multicolumn{9}{c}{\textit{OCR}} \\
\midrule
InfoVQA~\cite{mathew2022infographicvqa} & 82.46 & 86.00 & 90.76 & 83.04 & 86.00 & 94.22 & -- & 83.30 \\
OCRBench-v2~\cite{fu2024ocrbench} & 61.30 & 61.55 & 66.54 & 68.64 & 61.50 & 73.71 & -- & 58.90 \\
AI2D~\cite{hiippala2021ai2d} & 91.74 & 84.90 & 90.20 & 92.23 & 86.90 & 92.60 & 86.04 & -- \\
OCRBench~\cite{liu2024ocrbench} & 82.10 & 81.90 & 89.20 & 91.90 & 83.90 & 91.00 & 86.30 & 86.50 \\

\midrule
\multicolumn{9}{c}{\textit{Hallucination}} \\
\midrule
HallusionBench~\cite{guan2024hallusionbench} & 67.75 & 65.40 & 69.30 & 68.95 & 66.00 & 67.90 & -- & -- \\

\midrule
\multicolumn{9}{c}{\textit{Visual Reasoning}} \\
\midrule
BabyVision~\cite{chen2026babyvision} & 25.00 & 17.78 & 25.80 & 31.70 & 18.60 & 29.60 & 11.34 & -- \\
TiR~\cite{li2025tir} & 28.15 & 22.30 & 31.90 & 29.30 & 22.50 & 42.30 & 24.19 & -- \\

\midrule
\multicolumn{9}{c}{\textit{Spatial Intelligence}} \\
\midrule
VSI-Bench~\cite{vsi} & 62.66 & 56.61* & 55.67* & 56.90 & 51.56* & 58.10* & 32.91 & -- \\
ViewSpatial~\cite{li2025viewspatial} & 56.19 & 47.25 & 48.19 & 58.52 & 47.37 & 50.78 & 41.68 & -- \\
MindCube-Tiny~\cite{mindcube} & 62.01 & 43.17 & 57.59 & 70.86 & 40.86 & 63.46 & 48.84 & -- \\
3DSR-Bench~\cite{ma20243dsrbench} & 64.88 & 54.48 & 56.77 & 62.96 & 55.55 & 66.60 & 53.61 & -- \\
\bottomrule
\end{tabularx}
\end{table}

\textbf{General Understanding.}
We report multimodal tasks including multimodal reasoning (MMMU~\cite{yue2024mmmu}, MMMU-Pro~\cite{yue2025mmmu}, MathVista~\cite{lu2023mathvista}, MathVision~\cite{wang2024measuring}), general VLM understanding (MMBench-EN~\cite{liu2024mmbench}, MMStar~\cite{chen2024we}), OCR (InfoVQA~\cite{mathew2022infographicvqa}, OCRBench~\cite{liu2024ocrbench}, OCRBench-v2~\cite{fu2024ocrbench}, AI2D~\cite{hiippala2021ai2d}), hallucination detection (HallusionBench~\cite{guan2024hallusionbench}), and advanced visual reasoning (BabyVision~\cite{chen2026babyvision}, TiR~\cite{li2025tir}). For fair comparison, we prioritize official results when available and otherwise re-evaluate baselines using \texttt{vLLM} under their recommended settings within the \texttt{EvalScope} framework.

In Table~\ref{tab:multimodal_understanding_results}, our native SenseNova-U1 achieves strong performance, even without particular reinforcement learning for understanding domains. It consistently outperforms strong baselines such as Qwen3VL-8B~\cite{Qwen3-VL} built on the same LLM~\cite{yang2025qwen3} on multimodal reasoning benchmarks, while showing clear advantages in mathematical reasoning, highlighting the effectiveness of the encoder-free architecture.
In text-rich understanding, our models obtain notable gains over both similarly sized and larger competitors~\cite{Qwen3-VL,gemma42026,team2026longcat}. On general vision–language benchmarks and hallucination evaluation, SenseNova-U1 matches or surpasses leading models while preserving robust and grounded predictions.
Overall, compared to Qwen3.5~\cite{qwen35blog}, SenseNova-U1 can achieve competitive performance across a wide range of tasks while establishing a new frontier in efficient training and modeling through encoder-free architectures.

\textbf{Spatial Intelligence.}
We employ \texttt{EASI}~\cite{easi2025} toolkit to evaluate performance across key spatial intelligence benchmarks, covering capabilities such as metric measurement, spatial relations, perspective-taking, and comprehensive reasoning~\cite{sensenova-si}. 
SenseNova-U1 demonstrates remarkable performance on VSI-Bench~\cite{vsi}, ViewSpatial~\cite{li2025viewspatial}, MindCube-Tiny~\cite{mindcube}, and 3DSR-Bench~\cite{ma20243dsrbench}, highlighting strong spatial intelligence across both high-level reasoning and low-level geometric representation. These results suggest that native end-to-end multimodal modeling not only benefits semantic and compositional spatial understanding, but also preserves fine-grained structural and geometric perception essential for robust spatial reasoning.
Note that the \texttt{EASI} protocol uses 32 input frames on VSI-Bench; we adopt this standard for all models except the Qwen variants, which require 128-frame input for the best performance.

\begin{table}[t]
\centering
\scriptsize
\caption{\textbf{Quantitative evaluation results on language understanding benchmarks}.}
\label{tab:pure_text_results}
\vspace{1em}
\setlength{\tabcolsep}{1pt}
\begin{tabularx}{\textwidth}{@{}l*{3}{>{\centering\arraybackslash}X}|*{4}{>{\centering\arraybackslash}X}@{}}
\toprule
\textbf{Benchmark}
& \makecell{\textbf{SenseNova-U1}\\\textbf{8B-Think}}
& \makecell{\textbf{Qwen3VL}\\\textbf{8B-Think}}
& \makecell{\textbf{Qwen3.5}\\\textbf{9B}}
& \makecell{\textbf{SenseNova-U1}\\\textbf{30BA3B-Think}}
& \makecell{\textbf{Qwen3VL}\\\textbf{30BA3B-Think}}
& \makecell{\textbf{Qwen3.5}\\\textbf{35BA3B}}
& \makecell{\textbf{Gemma4}\\\textbf{26BA4B}} \\
\midrule
\multicolumn{8}{c}{\textit{Knowledge}} \\
\midrule
MMLU-Pro~\cite{wang2024mmlu} & 81.44 & 77.30 & 82.50 & 84.04 & 80.50 & 85.30 & 84.89 \\
MMLU-Redux~\cite{gema2025we} & 87.61 & 88.80 & 91.10 & 89.44 & 90.90 & 93.30 & 91.91 \\
C-Eval~\cite{huang2023c} & 84.40 & 83.88 & 88.20 & 85.89 & 87.29 & 90.20 & 82.39 \\
SuperGPQA~\cite{du2025supergpqa} & 49.67 & 51.20 & 58.20 & 59.71 & 56.40 & 63.40 & 61.88 \\
\midrule
\multicolumn{8}{c}{\textit{Instruction Following}} \\
\midrule
IFEval~\cite{zhou2023instruction} & 91.13 & 83.20 & 91.50 & 92.39 & 81.70 & 91.90 & -- \\
IFBench~\cite{zhang2025if} & 67.01 & 29.93 & 64.50 & 79.79 & 34.69 & 70.20 & 25.51 \\
\midrule
\multicolumn{8}{c}{\textit{Agent}} \\
\midrule
Tau2~\cite{tau2bench} & 71.70 & 31.65 & 79.10 & 75.39 & 46.40 & 81.20 & 68.20 \\
Claw eval~\cite{claweval} & 58.90 & 21.70 & 65.40 & 58.50 & 22.10 & 36.50 & 60.60 \\
\bottomrule
\end{tabularx}
\end{table}

\subsubsection{Text Understanding}
\textbf{General Understanding.} 
We evaluate pure-text capabilities across two categories. For knowledge-intensive reasoning, we report results on MMLU-Pro~\cite{wang2024mmlu}, MMLU-Redux~\cite{gema2025we}, C-Eval~\cite{huang2023c}, and SuperGPQA~\cite{du2025supergpqa}, covering broad academic knowledge and advanced reasoning across disciplines and languages. For instruction following, we evaluate on IFEval~\cite{zhou2023instruction} and IFBench~\cite{zhang2025if}, which measure adherence to complex and constraint-heavy instructions.

As shown in Table~\ref{tab:pure_text_results}, SenseNova-U1 demonstrates particularly strong gains in instruction following, consistently outperforming the Qwen3.5 series on both IFEval and IFBench. These results indicate that native encoder-free architectures can effectively handle complex, constraint-heavy instructions.
It surpasses the Qwen3VL series and narrows the gap with the Qwen3.5 series on MMLU-Pro and SuperGPQA, reflecting strong academic knowledge and professional-level reasoning capabilities. Besides, strong text knowledge capability is further evidenced by consistent trends on C-Eval and MMLU-Redux.
These results demonstrate that our native models using fewer training resources can rival the top-tier encoder-based architecture, further proving its effectiveness with efficient data-scaling capability.

\textbf{Agentic Function.}
We evaluate the agentic capabilities of SenseNova-U1 on two complementary benchmarks: 
$\tau^2$-Bench~\cite{tau2bench} and Claw-Eval~\cite{claweval}. Specifically, $\tau^2$-Bench focuses on end-to-end task completion across domains such as Retail, Airline, and Telecom, requiring sustained interactions, adaptive tool use, and environment-aware decision making. Claw-Eval instead emphasizes trustworthy agent behavior through trajectory-aware evaluation over task completion, safety, and robustness, using execution traces, audit logs, and repeated-trial verification.

As shown in Table~\ref{tab:pure_text_results}, SenseNova-U1 demonstrates powerful multi-turn reasoning and agent capabilities across both $\tau^2$-Bench and Claw-Eval. Notably, our A3B variant consistently outperforms existing multimodal baselines~\cite{Qwen3-VL} and approaches the performance of substantially larger reasoning-oriented models~\cite{qwen35blog,gemma42026} despite using fewer active parameters. In particular, SenseNova-U1 exhibits strong trajectory-level reliability, coherent long-horizon interaction, and robust tool-use behavior in complex multi-step environments. We attribute the remaining gap to dense reasoning-focused baselines partly to deliberate training trade-offs that prioritize high-fidelity multimodal generation and interleaved vision-language capabilities. Nevertheless, these results highlight the effectiveness of our native multimodal framework in sustaining trustworthy and capable agent behavior across diverse real-world scenarios.

\begin{table}[h!]
\centering
\renewcommand{\arraystretch}{1}
\setlength{\tabcolsep}{3.3pt}
\caption{\textbf{Quantitative evaluation results on GenEval}. The parameters of the generation component are denoted as \textit{\# Params}; \textit{A} in this column denotes activated parameters, e.g., 8BA3B means 8B total generation parameters with 3B activated during inference.}
\label{tab:geneval_results}
\begin{adjustbox}{width=\textwidth}
\begin{tabular}{lc|cccccc|c}
\toprule
\textbf{Model} & \textbf{\# Params} & \textbf{Single Object} & \textbf{Two Object} & \textbf{Counting} & \textbf{Colors} & \textbf{Position} & \textbf{Attribute Binding} & \textbf{Overall}$\uparrow$ \\
\midrule
\multicolumn{9}{c}{\textit{Closed-source Models}} \\
\midrule
GPT-Image-1~\cite{GPT-Image-1} & - & 0.99 & 0.92 & 0.85 & 0.92 & 0.75 & 0.61 & 0.84 \\
Seedream 4.0~\cite{seedream2025seedream} & - &	0.99	&0.92	&0.72	&0.91	&0.76	&0.74	&0.84 \\
Seedream 3.0~\cite{gao2025seedream}& - &0.99& 0.96& 0.91& 0.93& 0.47& 0.80& 0.84\\
\midrule
\multicolumn{9}{c}{\textit{Open-source Models}} \\
\midrule
\textbf{SenseNova-U1} & \textbf{8BA3B} & 1.00 & 0.96 & 0.89 & 0.91 & 0.92 & 0.77 & \textbf{0.91} \\
\textbf{SenseNova-U1} & \textbf{8B} & 1.00 & 0.96 & 0.92 & 0.92 & 0.91 & 0.76 & \textbf{0.91} \\
Tuna~\cite{liu2025tuna} & 7B & 1.00 & 0.97 & 0.81 & 0.91 & 0.88 & 0.83 & 0.90 \\
OneCAT~\cite{li2025onecat} & 9BA3B & 1.00 & 0.96 & 0.84 & 0.94 & 0.84 & 0.80 & 0.90 \\
NEO-unify~\cite{sensenova2026neounify} & 8B & 1.00 & 0.96 & 0.90 & 0.91 & 0.91 & 0.77 & 0.90 \\
Mogao~\cite{liao2025mogao} & 7B & 1.00 & 0.97 & 0.83 & 0.93 & 0.84 & 0.80 & 0.89 \\
Lumina-DiMOO~\cite{xin2025lumina} & 8B & 1.00 & 0.94 & 0.85 & 0.89 & 0.85 & 0.76 & 0.88 \\
Qwen-Image~\cite{wu2025qwenimagetechnicalreport} & 20B & 0.99 & 0.92 & 0.89 & 0.88 & 0.76 & 0.77 & 0.87 \\
NEO-unify~\cite{sensenova2026neounify} & 2B & 0.99 & 0.92 & 0.89 & 0.86 & 0.77 & 0.76 & 0.87 \\
Tuna-2~\cite{tuna2} & 7B & 0.99 & 0.96 & 0.80 & 0.91 & 0.84 & 0.76 & 0.87 \\
InternVL-U~\cite{tian2026internvludemocratizingunifiedmultimodal} & 1.7B & 0.99 & 0.94 & 0.74 & 0.91 & 0.77 & 0.74 & 0.85 \\
LongCat-Next~\cite{team2026longcat} & 68BA3B & - & - & - & - & - & - & 0.84 \\
Z-Image~\cite{cai2025z} & 6B & 1.00 & 0.94 & 0.78 & 0.93 & 0.62 & 0.77 & 0.84 \\
BLIP3-o~\cite{chen2025blip3o} & 1.4B & - & - & - & - & - & - & 0.84 \\
X-Omni~\cite{geng2025x} & 12B & 0.98 & 0.95 & 0.75 & 0.91 & 0.71 & 0.68 & 0.83 \\
BAGEL~\cite{deng2025bagel} & 7B & 0.99 & 0.94 & 0.81 & 0.88 & 0.64 & 0.63 & 0.82 \\
Janus-Pro~\cite{chen2025janus} & 7B & 0.99 & 0.89 & 0.59 & 0.90 & 0.79 & 0.66 & 0.80 \\
OmniGen2~\cite{wu2025omnigen2} & 4B & 1.00 & 0.95 & 0.64 & 0.88 & 0.55 & 0.76 & 0.80 \\
UniWorld-V1~\cite{lin2025uniworld} & 12B & 0.99 & 0.93 & 0.79 & 0.89 & 0.49 & 0.70 & 0.80 \\
Show-o2~\cite{xie2025show} & 7B & 1.00 & 0.87 & 0.58 & 0.92 & 0.52 & 0.62 & 0.76 \\
SD3-Medium~\cite{esser2024scaling} & 2B & 0.99 & 0.94 & 0.72 & 0.89 & 0.33 & 0.60 & 0.74 \\
Emu3.5~\cite{cui2025emu35nativemultimodalmodels} & 32B & - & - & - & - & - & - & 0.73 \\
FLUX.1-dev~\cite{flux2024} & 12B & 0.98 & 0.81 & 0.74 & 0.79 & 0.22 & 0.45 & 0.66 \\
\bottomrule
\end{tabular}
\end{adjustbox}
\end{table}

\subsubsection{Image Generation}

We evaluate SenseNova-U1 on text-to-image generation from complementary perspectives: general generation, text-centric generation, complex infographic generation, and reasoning-centric generation. Together, these evaluations cover object-level composition, prompt following, long-text rendering, knowledge-informed generation, structured professional visual content creation, and more tightly coupled understanding-generation behaviors. 
Given our $32\times32$ downsampling ratios, we generate 2K images and downsample them to 1K for evaluation under comparable computational budgets.

\textbf{General Generation.}
For general text-to-image generation, we adopt GenEval~\cite{ghosh2023geneval}, DPG-Bench~\cite{hu2024ella}, OneIG-Bench~\cite{chang2025oneig}, and TIIF-Bench~\cite{wei2025tiif}. These benchmarks examine object-level compositional generation, dense prompt following, and fine-grained overall capability from complementary perspectives. Across them, SenseNova-U1 remains highly competitive, showing that the native unified modeling paradigm does not sacrifice fundamental generation quality.

\paragraph{GenEval.}
GenEval~\cite{ghosh2023geneval} mainly evaluates compositional generation ability across object co-occurrence, counting, color, position, and attribute binding. As shown in Table~\ref{tab:geneval_results}, SenseNova-U1-A3B-MoT and SenseNova-U1-8B-MoT both achieve an overall score of 0.91, outperforming representative open-source models such as Qwen-Image at 0.87, Lumina-DiMOO at 0.88, and BAGEL at 0.82. More specifically, our model maintains consistently strong performance on Single Object, Two Object, Counting, Colors, and Position, which indicates stable object-level control and compositional consistency. Although the Attribute Binding score remains slightly below the 0.80 achieved by OneCAT and Mogao, SenseNova-U1 still attains the best overall result through a more balanced performance profile. These results show that SenseNova-U1 maintains strong and balanced compositional generation ability across object-level factors.

\begin{table}[t]
\centering
\caption{\textbf{Quantitative evaluation results on DPG-Bench}.
The parameters of the generation component are denoted as \textit{\# Params}; \textit{A} in this column denotes activated parameters, e.g., 8BA3B means 8B total generation parameters with 3B activated during inference.}
\label{tab:dpg_results}
\renewcommand{\arraystretch}{1}
\setlength{\tabcolsep}{3.1pt}
\begin{adjustbox}{width=0.75\textwidth}
\begin{tabular}{lc|ccccc|c}
\toprule
\textbf{Model} & \textbf{\# Params} & \textbf{Global} & \textbf{Entity} & \textbf{Attribute} & \textbf{Relation} & \textbf{Other} & \textbf{Overall}$\uparrow$ \\
\midrule
\multicolumn{8}{c}{\textit{Closed-source Models}} \\
\midrule
Seedream 4.5~\cite{seedream45} & - & 89.24 & 94.30 & 92.14 & 92.23 & 93.83 & 88.63 \\
Nano-Banana-Pro~\cite{deepmind_gemini3proimage_2025} & - & 91.00 & 92.85 & 91.56 & 92.39 & 89.93 & 87.16 \\
GPT-Image-1~\cite{GPT-Image-1} & - & 88.89 & 88.94 & 89.84 & 92.63 & 90.96 & 85.15 \\
\midrule
\multicolumn{8}{c}{\textit{Open-source Models}} \\
\midrule
Qwen-Image~\cite{wu2025qwenimagetechnicalreport} & 20B & 91.32 & 91.56 & 92.02 & 94.31 & 92.73 & 88.32 \\
\textbf{SenseNova-U1} & \textbf{8BA3B} & 94.19 & 92.05 & 91.05 & 93.22 & 93.44 & \textbf{88.14} \\
Z-Image~\cite{cai2025z} & 6B & 93.39 & 91.22 & 93.16 & 92.22 & 91.52 & 88.14 \\
JoyAI-Image~\cite{joyai_image} & 16B & - & - & - & - & - & 88.05 \\
\textbf{SenseNova-U1} & \textbf{8B} & 88.74 & 90.90 & 92.43 & 92.43 & 92.50  & \textbf{87.78} \\
X-Omni~\cite{geng2025x} & 12B & - & - & - & - & - & 87.65 \\
Tuna~\cite{liu2025tuna} & 7B & 90.42 & 91.68 & 90.94 & 91.87 & 90.73 & 86.76 \\
NEO-unify~\cite{sensenova2026neounify} & 8B & 91.00 & 91.53 & 92.06 & 94.14 & 90.43 & 86.71 \\
Tuna-2~\cite{tuna2} & 7B & 89.50 & 91.40 & 92.07 & 91.91 & 88.81 & 86.54 \\
NEO-unify~\cite{sensenova2026neounify} & 2B & 89.49 & 92.87 & 91.26 & 92.29 & 92.13 & 86.54 \\
Show-o2~\cite{xie2025show} & 7B & - & - & - & - & - & 86.14 \\
Lumina-DiMOO~\cite{xin2025lumina} & 8B & 81.46 & 92.08 & 88.98 & 94.31 & 82.00 & 86.04 \\
InternVL-U~\cite{tian2026internvludemocratizingunifiedmultimodal} & 1.7B & 90.39 & 90.78 & 90.68 & 90.29 & 88.77 & 85.18 \\
BAGEL~\cite{deng2025bagel} & 7B & 88.94 & 90.37 & 91.29 & 90.82 & 88.67 & 85.07 \\
LongCat-Next~\cite{team2026longcat} & 68BA3B & - & - & - & - & - & 84.66 \\
OneCAT~\cite{li2025onecat} & 9BA3B & - & - & - & - & - & 84.53 \\
Mogao~\cite{liao2025mogao} & 7B & - & - & - & - & - & 84.33 \\
Janus-Pro~\cite{chen2025janus} & 7B & 86.90 & 88.90 & 89.40 & 89.32 & 89.48 & 84.19 \\
SD3-Medium~\cite{esser2024scaling} & 2B & 87.90 & 91.01 & 88.83 & 80.70 & 88.68 & 84.08 \\
FLUX.1-dev~\cite{flux2024} & 12B & 74.35 & 90.00 & 88.96 & 90.87 & 88.33 & 83.84 \\
Ovis-U1~\cite{wang2025ovis} & 1.2B & 82.37 & 90.08 & 88.68 & 93.35 & 85.20 & 83.72 \\
OmniGen2~\cite{wu2025omnigen2} & 4B & 88.81 & 88.83 & 90.18 & 89.37 & 90.27 & 83.57 \\
BLIP3-o~\cite{chen2025blip3o} & 1.4B & - & - & - & - & - & 81.60 \\
UniWorld-V1~\cite{lin2025uniworld} & 12B & 83.64 & 88.39 & 88.44 & 89.27 & 87.22 & 81.38 \\
\bottomrule
\end{tabular}
\end{adjustbox}
\end{table}

\paragraph{DPG-Bench.}
DPG-Bench~\cite{hu2024ella} evaluates fine-grained instruction following using dense prompts across five dimensions: Global, Entity, Attribute, Relation, and Other. As shown in Table~\ref{tab:dpg_results}, SenseNova-U1 ranks among the top-performing models on this benchmark.
Both SenseNova-U1-A3B-MoT and SenseNova-U1-8B-MoT remain highly competitive with leading specialized generation systems such as closed-source Seedream 4.5, open-source Qwen-Image, and Z-Image, despite being trained within a unified multimodal framework rather than optimized solely for image synthesis. Notably, SenseNova-U1-A3B-MoT achieves the highest Global score, underscoring the strong semantic planning, holistic scene composition, and long-range instruction consistency enabled by native end-to-end multimodal modeling.
\begin{table}[t]
\centering
\caption{\textbf{Quantitative evaluation results on OneIG-EN}.
The parameters of the generation component are denoted as \textit{\# Params}; \textit{A} in this column denotes activated parameters, e.g., 8BA3B means 8B total generation parameters with 3B activated during inference.}
\label{tab:oneig_en_results}
\renewcommand{\arraystretch}{1}
\setlength{\tabcolsep}{3pt}
\begin{adjustbox}{width=0.88\textwidth}
\begin{tabular}{lc|ccccc|c}
\toprule
\textbf{Model} & \textbf{\# Params} & \textbf{Alignment} & \textbf{Text} & \textbf{Reasoning} & \textbf{Style} & \textbf{Diversity} & \textbf{Overall}$\uparrow$ \\
\midrule
\multicolumn{8}{c}{\textit{Closed-source Models}} \\
\midrule
Gemini-2.5-Flash-Image~\cite{google2025gemini25flashmodelcard} & - & 0.878 & 0.894 & 0.346 & 0.450 & 0.182 & 0.550 \\
GPT-Image-1~\cite{GPT-Image-1} & - & 0.851 & 0.857 & 0.345 & 0.462 & 0.151 & 0.533 \\
Seedream 3.0~\cite{gao2025seedream} & - & 0.818 & 0.865 & 0.275 & 0.413 & 0.277 & 0.530 \\
Imagen4~\cite{google2025imagen4} & - & 0.857 & 0.805 & 0.338 & 0.377 & 0.199 & 0.515 \\
Recraft V3~\cite{recraftv3} & - & 0.810 & 0.795 & 0.323 & 0.378 & 0.205 & 0.502 \\
Kolors 2.0~\cite{kuaishou2025kolors} & - & 0.820 & 0.427 & 0.262 & 0.360 & 0.300 & 0.434 \\
Imagen3~\cite{Imagen3} & - & 0.843 & 0.343 & 0.313 & 0.359 & 0.188 & 0.409 \\
\midrule
\multicolumn{8}{c}{\textit{Open-source Models}} \\
\midrule
Emu3.5~\cite{cui2025emu35nativemultimodalmodels} & 32B & 0.902 & 0.994 & 0.345 & 0.427 & 0.151 & 0.564 \\
\textbf{SenseNova-U1} & \textbf{8B} & 0.882 & 0.969 & 0.330 & 0.396 & 0.166 & \textbf{0.549} \\
\textbf{SenseNova-U1} & \textbf{8BA3B} & 0.887 & 0.861 & 0.317 & 0.458 & 0.194 & \textbf{0.543} \\
Qwen-Image~\cite{wu2025qwenimagetechnicalreport} & 20B & 0.882 & 0.891 & 0.306 & 0.418 & 0.197 & 0.539 \\
HiDream-I1-Full~\cite{cai2025hidream} & 17B & 0.829 & 0.707 & 0.317 & 0.347 & 0.186 & 0.477 \\
SD3.5 Large~\cite{esser2024scaling} & 8B & 0.809 & 0.629 & 0.294 & 0.353 & 0.225 & 0.462 \\
FLUX.1 [Dev]~\cite{flux2024} & 12B & 0.786 & 0.523 & 0.253 & 0.368 & 0.238 & 0.434 \\
BAGEL~\cite{deng2025bagel} & 7B & 0.769 & 0.244 & 0.173 & 0.367 & 0.251 & 0.361 \\
BLIP3-o~\cite{chen2025blip3o} & 1.4B & 0.711 & 0.013 & 0.223 & 0.361 & 0.229 & 0.307 \\
Janus-Pro~\cite{chen2025janus} & 7B & 0.553 & 0.001 & 0.139 & 0.276 & 0.365 & 0.267 \\
\bottomrule
\end{tabular}
\end{adjustbox}
\end{table}

\begin{table}[h!]
\centering
\caption{\textbf{Quantitative evaluation results on OneIG-ZH}.
The parameters of the generation component are denoted as \textit{\# Params}; \textit{A} in this column denotes activated parameters, e.g., 8BA3B means 8B total generation parameters with 3B activated during inference.}
\label{tab:oneig_zh_results}
\renewcommand{\arraystretch}{1}
\setlength{\tabcolsep}{3pt}
\begin{adjustbox}{width=0.88\textwidth}
\begin{tabular}{lc|ccccc|c}
\toprule
\textbf{Model} & \textbf{\# Params} & \textbf{Alignment} & \textbf{Text} & \textbf{Reasoning} & \textbf{Style} & \textbf{Diversity} & \textbf{Overall}$\uparrow$ \\
\midrule
\multicolumn{8}{c}{\textit{Closed-source Models}} \\
\midrule
Seedream 3.0~\cite{gao2025seedream} & - & 0.793 & 0.928 & 0.281 & 0.397 & 0.243 & 0.528 \\
GPT-Image-1~\cite{GPT-Image-1} & - & 0.812 & 0.650 & 0.300 & 0.449 & 0.159 & 0.474 \\
Kolors 2.0~\cite{kuaishou2025kolors} & - & 0.738 & 0.502 & 0.226 & 0.331 & 0.333 & 0.426 \\
Gemini-2.5-Flash-Image~\cite{google2025gemini25flashmodelcard} & - & 0.825 & 0.276 & 0.298 & 0.427 & 0.198 & 0.337 \\
\midrule
\multicolumn{8}{c}{\textit{Open-source Models}} \\
\midrule
Qwen-Image~\cite{wu2025qwenimagetechnicalreport} & 20B & 0.825 & 0.963 & 0.267 & 0.405 & 0.279 & 0.548 \\
\textbf{SenseNova-U1} & \textbf{8BA3B} & 0.847 & 0.906 & 0.301 & 0.446 & 0.202 & \textbf{0.540} \\
\textbf{SenseNova-U1} & \textbf{8B} & 0.826 & 0.977 & 0.303 & 0.392 & 0.176 & \textbf{0.535} \\
Emu3.5~\cite{cui2025emu35nativemultimodalmodels} & 32B & 0.853 & 0.941 & 0.300 & 0.386 & 0.166 & 0.529 \\
BAGEL~\cite{deng2025bagel} & 7B & 0.672 & 0.365 & 0.186 & 0.357 & 0.268 & 0.370 \\
HiDream-I1-Full~\cite{cai2025hidream} & 17B & 0.620 & 0.205 & 0.256 & 0.304 & 0.300 & 0.337 \\
BLIP3-o~\cite{chen2025blip3o} & 1.4B & 0.608 & 0.092 & 0.213 & 0.369 & 0.233 & 0.303 \\
Janus-Pro~\cite{chen2025janus} & 7B & 0.324 & 0.148 & 0.104 & 0.264 & 0.358 & 0.240 \\
\bottomrule
\end{tabular}
\end{adjustbox}
\end{table}

\paragraph{OneIG-Bench.}
OneIG-Bench~\cite{chang2025oneig} provides a fine-grained evaluation of image generation quality across Alignment, Text, Reasoning, Style, and Diversity in both English and Chinese. As shown in Table~\ref{tab:oneig_en_results} and Table~\ref{tab:oneig_zh_results}, SenseNova-U1 remains highly competitive across both language settings, demonstrating strong multilingual generation capability within a unified framework. In particular, our models exhibit clear strengths in Alignment and Text understanding. SenseNova-U1-A3B-MoT achieves a leading Alignment score on the English benchmark, while SenseNova-U1-8B-MoT attains the best Text performance on both the English and Chinese tracks among all compared methods. These results underscore the effectiveness of native end-to-end multimodal modeling in preserving fine-grained text-image alignment, robust multilingual text rendering, and precise instruction-following under complex generation scenarios.

\paragraph{TIIF-Bench.}
TIIF-Bench~\cite{wei2025tiif} systematically evaluates image generation spanning attributes, relations, reasoning, style, and text. As shown in Table~\ref{tab:tiif_results_short} and Table~\ref{tab:tiif_results_long}, SenseNova-U1 achieves consistently strong performance under both short and long instruction settings. In particular, SenseNova-U1-8B-MoT attains the best overall results among all compared methods, while SenseNova-U1-A3B-MoT also remains highly competitive. These results suggest that SenseNova-U1 extends beyond accurate text rendering to more challenging text-centric generation scenarios that require jointly satisfying fine-grained textual constraints, compositional reasoning, and global instruction consistency.

\begin{table}[h!]
\centering
\renewcommand{\arraystretch}{1}
\setlength{\tabcolsep}{3pt}
\caption{\textbf{Quantitative evaluation results on TIIF testmini (short)}. Abbrev.: Avg = Average, Attr = Attribute, Rel = Relation, Rsn = Reasoning, ARel = Attribute+Relation, ARsn = Attribute+Reasoning, RRsn = Relation+Reasoning, RealW = Real World.}
\label{tab:tiif_results_short}
\begin{adjustbox}{width=\textwidth}
\begin{tabular}{lc|c|cccc|cccccc|c}
\toprule
\multirow{2}{*}{\textbf{Model}} & \multirow{2}{*}{\textbf{\# Params}} & \multirow{2}{*}{\textbf{Overall}$\uparrow$}
& \multicolumn{4}{c|}{\textbf{Basic}}
& \multicolumn{6}{c|}{\textbf{Advanced}}
& \textbf{Design} \\
\cmidrule(lr){4-7} \cmidrule(lr){8-13} \cmidrule(lr){14-14}
& & & Avg & Attr & Rel & Rsn & Avg & ARel & ARsn & RRsn & Style & Text & RealW \\
\midrule
\multicolumn{14}{c}{\textit{Closed-source Models}} \\
\midrule
GPT-Image-1~\cite{GPT-Image-1} & - & 89.15 & 90.75 & 91.33 & 84.57 & 96.32 & 88.55 & 87.07 & 87.22 & 85.59 & 90.00 & 89.83 & 89.73 \\
Seedream 3.0~\cite{gao2025seedream} & - & 86.02 & 87.07 & 90.50 & 89.85 & 80.86 & 79.16 & 79.76 & 77.23 & 75.64 & 100.00 & 97.17 & 83.21 \\
DALL-E 3~\cite{betker2023dalle3} & - & 74.96 & 78.72 & 79.50 & 80.82 & 75.82 & 73.39 & 73.45 & 72.01 & 63.59 & 89.66 & 66.83 & 72.93 \\
MidJourney v7~\cite{midjourneyV7} & - & 68.74 & 77.41 & 77.58 & 82.07 & 72.57 & 64.66 & 67.20 & 81.22 & 60.72 & 83.33 & 24.83 & 68.83 \\
FLUX.1 [Pro]~\cite{flux2024} & - & 67.32 & 79.08 & 78.83 & 82.82 & 75.57 & 61.10 & 62.32 & 69.84 & 65.96 & 63.00 & 35.83 & 71.80 \\
\midrule
\multicolumn{14}{c}{\textit{Open-source Models}} \\
\midrule
\textbf{SenseNova-U1} & \textbf{8B} & \textbf{89.74} & 90.38 & 91.50 & 88.46 & 91.19 & 85.21 & 85.37 & 83.78 & 84.10 & 100.00 & 89.59 & 93.66 \\
Emu3.5~\cite{cui2025emu35nativemultimodalmodels} & 32B & 89.48 & 87.05 & 90.50 & 89.80 & 80.85 & 84.65 & 82.91 & 83.76 & 83.45 & 100.00 & 100.00 & 94.03 \\
\textbf{SenseNova-U1} & \textbf{8BA3B} & \textbf{89.25} & 88.28 & 90.50 & 86.83 & 87.50 & 87.60 & 87.48 & 86.85 & 87.03 & 96.67 & 90.05 & 90.30 \\
Qwen-Image~\cite{wu2025qwenimagetechnicalreport} & 20B & 86.14 & 86.18 & 90.50 & 88.22 & 79.81 & 79.30 & 79.21 & 78.85 & 75.57 & 100.00 & 92.76 & 90.30 \\
FLUX.1 [dev]~\cite{flux2024} & 12B & 71.09 & 83.12 & 87.05 & 87.25 & 75.01 & 65.79 & 67.07 & 73.84 & 69.09 & 66.67 & 43.83 & 70.72 \\
SD 3~\cite{esser2024scaling} & 8B & 67.46 & 78.32 & 83.33 & 82.07 & 71.07 & 61.46 & 61.07 & 68.84 & 50.96 & 66.67 & 59.83 & 63.23 \\
Janus-Pro~\cite{chen2025janus} & 7B & 66.50 & 79.33 & 79.33 & 78.32 & 80.32 & 59.71 & 66.07 & 70.46 & 67.22 & 60.00 & 28.83 & 65.84 \\
Infinity~\cite{han2025infinity} & 8B & 62.07 & 73.08 & 74.33 & 72.82 & 72.07 & 56.64 & 60.44 & 74.22 & 60.22 & 80.00 & 10.83 & 54.28 \\
Show-o~\cite{xie2024show} & 1.3B & 59.72 & 73.08 & 74.83 & 78.82 & 65.57 & 53.67 & 60.95 & 68.59 & 66.46 & 63.33 & 3.83 & 55.02 \\
\bottomrule
\end{tabular}
\end{adjustbox}
\end{table}

\begin{table}[h!]
\centering
\renewcommand{\arraystretch}{1}
\setlength{\tabcolsep}{3pt}
\caption{\textbf{Quantitative evaluation results on TIIF testmini (long)}. Abbrev.: Avg = Average, Attr = Attribute, Rel = Relation, Rsn = Reasoning, ARel = Attribute+Relation, ARsn = Attribute+Reasoning, RRsn = Relation+Reasoning, RealW = Real World.}
\label{tab:tiif_results_long}
\begin{adjustbox}{width=\textwidth}
\begin{tabular}{lc|c|cccc|cccccc|c}
\toprule
\multirow{2}{*}{\textbf{Model}} & \multirow{2}{*}{\textbf{\# Params}} & \multirow{2}{*}{\textbf{Overall}$\uparrow$}
& \multicolumn{4}{c|}{\textbf{Basic}}
& \multicolumn{6}{c|}{\textbf{Advanced}}
& \textbf{Design} \\
\cmidrule(lr){4-7} \cmidrule(lr){8-13} \cmidrule(lr){14-14}
& & & Avg & Attr & Rel & Rsn & Avg & ARel & ARsn & RRsn & Style & Text & RealW \\
\midrule
\multicolumn{14}{c}{\textit{Closed-source Models}} \\
\midrule
GPT-Image-1~\cite{GPT-Image-1} & - & 88.29 & 89.66 & 87.08 & 84.57 & 97.32 & 88.35 & 89.44 & 83.96 & 83.21 & 93.33 & 86.83 & 93.46 \\
Seedream 3.0~\cite{gao2025seedream} & - & 84.31 & 84.93 & 90.00 & 85.94 & 78.86 & 80.60 & 81.82 & 78.85 & 78.64 & 93.33 & 87.78 & 83.58 \\
DALL-E 3~\cite{betker2023dalle3} & - & 70.81 & 78.50 & 79.83 & 78.82 & 76.82 & 67.27 & 67.20 & 71.34 & 60.72 & 86.67 & 54.83 & 60.99 \\
FLUX.1 [Pro]~\cite{flux2024} & - & 69.89 & 78.91 & 81.33 & 83.82 & 71.57 & 65.37 & 65.57 & 71.47 & 67.72 & 63.00 & 55.83 & 68.80 \\
MidJourney v7~\cite{midjourneyV7} & - & 65.69 & 76.00 & 81.83 & 76.82 & 69.32 & 60.53 & 62.70 & 71.59 & 64.59 & 80.00 & 20.83 & 63.61 \\
\midrule
\multicolumn{14}{c}{\textit{Open-source Models}} \\
\midrule
\textbf{SenseNova-U1} & \textbf{8B} & \textbf{89.17} & 91.02 & 94.50 & 90.49 & 88.06 & 85.34 & 87.23 & 84.14 & 83.14 & 100.00 & 82.81 & 92.16 \\
Emu3.5~\cite{cui2025emu35nativemultimodalmodels} & 32B & 88.18 & 88.41 & 92.50 & 90.78 & 81.94 & 84.04 & 83.08 & 85.73 & 81.09 & 90.00 & 95.93 & 92.54 \\
\textbf{SenseNova-U1} & \textbf{8BA3B} & \textbf{87.36} & 87.95 & 92.50 & 89.45 & 81.90 & 84.39 & 83.88 & 86.52 & 81.44 & 96.67 & 82.81 & 91.04 \\
Qwen-Image~\cite{wu2025qwenimagetechnicalreport} & 20B & 86.83 & 87.22 & 91.50 & 90.78 & 79.38 & 80.88 & 78.94 & 81.69 & 78.59 & 100.00 & 89.14 & 91.42 \\
FLUX.1 [dev]~\cite{flux2024} & 12B & 71.78 & 78.65 & 83.17 & 80.39 & 72.39 & 68.54 & 73.69 & 73.34 & 71.59 & 66.67 & 52.83 & 71.47 \\
SD 3~\cite{esser2024scaling} & 8B & 66.09 & 77.75 & 79.83 & 78.82 & 74.07 & 59.56 & 64.07 & 70.34 & 57.84 & 76.67 & 20.83 & 67.34 \\
Janus-Pro~\cite{chen2025janus} & 7B & 65.02 & 78.25 & 82.33 & 73.32 & 79.07 & 58.82 & 56.20 & 70.84 & 59.97 & 70.00 & 33.83 & 60.25 \\
Infinity~\cite{han2025infinity} & 8B & 62.32 & 75.41 & 76.83 & 77.57 & 71.82 & 54.98 & 55.57 & 64.71 & 59.71 & 73.33 & 23.83 & 56.89 \\
Show-o~\cite{xie2024show} & 1.3B & 58.86 & 75.83 & 79.83 & 78.32 & 69.32 & 50.38 & 56.82 & 68.96 & 56.22 & 66.67 & 2.83 & 50.92 \\
\bottomrule
\end{tabular}
\end{adjustbox}
\end{table}

\begin{table}[t]
\centering
\caption{\textbf{Quantitative evaluation results on CVTG-2K}.
The parameters of the generation component are denoted as \textit{\# Params}.}
\label{tab:cvtg_results}
\renewcommand{\arraystretch}{1}
\setlength{\tabcolsep}{3pt}
\begin{adjustbox}{width=0.9\textwidth}
\begin{tabular}{lc|cc|cccc|c}
\toprule
\multirow{2}{*}{\textbf{Model}} & \multirow{2}{*}{\textbf{\# Params}} & \multirow{2}{*}{\textbf{NED}} & \multirow{2}{*}{\textbf{CLIPScore}} & \multicolumn{4}{c|}{\textbf{Word Accuracy}} & \multirow{2}{*}{\textbf{Average}$\uparrow$} \\
\cmidrule(lr){5-8}
& & & & 2 regions& 3 regions & 4 regions & 5 regions & \\
\midrule
\multicolumn{9}{c}{\textit{Closed-source Models}} \\
\midrule
Seedream 4.5~\cite{seedream45} & - & 0.948 & 0.807 & 0.878 & 0.895 & 0.908 & 0.901 & 0.899 \\
GPT-Image-1~\cite{GPT-Image-1} & - & 0.948 & 0.798 & 0.878 & 0.866 & 0.873 & 0.822 & 0.857 \\
Nano-Banana-Pro~\cite{deepmind_gemini3proimage_2025} & - & 0.875 & 0.737 & 0.737 & 0.775 & 0.786 & 0.793 & 0.779 \\
\midrule
\multicolumn{9}{c}{\textit{Open-source Models}} \\
\midrule
\textbf{SenseNova-U1} & \textbf{8B} & 0.972 & 0.825 & 0.945 & 0.954 & 0.944 & 0.936 & \textbf{0.940} \\
Emu3.5~\cite{cui2025emu35nativemultimodalmodels} & 32B & - & - & - & - & - & - & 0.912 \\
\textbf{SenseNova-U1} & \textbf{8BA3B} & 0.944 & 0.824 & 0.884 & 0.883 & 0.883 & 0.875 & \textbf{0.881} \\
JoyAI-Image~\cite{joyai_image} & 16B & 0.937 & 0.799 & - & - & - & - & 0.874 \\
Z-Image~\cite{cai2025z} & 6B & 0.937 & 0.797 & 0.901 & 0.872 & 0.865 & 0.851 & 0.867 \\
Qwen-Image~\cite{wu2025qwenimagetechnicalreport} & 20B & 0.912 & 0.802 & 0.837 & 0.836 & 0.831 & 0.816 & 0.829 \\
LongCat-Next~\cite{team2026longcat} & 68BA3B & - & - & - & - & - & - & 0.764 \\
InternVL-U~\cite{tian2026internvludemocratizingunifiedmultimodal} & 1.7B & 0.804 & 0.816 & 0.729 & 0.660 & 0.618 & 0.549 & 0.623 \\
Lumina-DiMOO~\cite{xin2025lumina} & 8B & 0.805 & 0.831 & 0.723 & 0.646 & 0.571 & 0.505 & 0.590 \\
FLUX.1-dev~\cite{flux2024} & 12B & 0.688 & 0.740 & 0.609 & 0.553 & 0.466 & 0.432 & 0.497 \\
BAGEL~\cite{deng2025bagel} & 7B & 0.657 & 0.779 & 0.498 & 0.391 & 0.332 & 0.291 & 0.356 \\
Ovis-U1~\cite{wang2025ovis} & 1.2B & 0.477 & 0.725 & 0.133 & 0.109 & 0.091 & 0.065 & 0.093 \\
\bottomrule
\end{tabular}
\end{adjustbox}
\end{table}

\begin{table}[!h]
\centering
\caption{\textbf{Quantitative evaluation results on LongText-Bench}.
\textit{A} in \textit{\# Params} denotes activated parameters during inference.}
\label{tab:longtext_results}
\renewcommand{\arraystretch}{1}
\setlength{\tabcolsep}{3pt}
\begin{adjustbox}{width=0.70\textwidth}
\begin{tabular}{lc|cc}
\toprule
\textbf{Model} & \textbf{\# Params} & \textbf{LongText-Bench-EN}$\uparrow$ & \textbf{LongText-Bench-ZH}$\uparrow$ \\
\midrule
\multicolumn{4}{c}{\textit{Closed-source Models}} \\
\midrule
Seedream 4.5~\cite{seedream45} & - & 0.989 & 0.987 \\
Nano-Banana-Pro~\cite{deepmind_gemini3proimage_2025} & - & 0.981 & 0.949 \\
GPT-Image-1~\cite{GPT-Image-1} & - & 0.956 & 0.619 \\
\midrule
\multicolumn{4}{c}{\textit{Open-source Models}} \\
\midrule
\textbf{SenseNova-U1} & \textbf{8B} & \textbf{0.979} & \textbf{0.962} \\
Emu3.5~\cite{cui2025emu35nativemultimodalmodels} & 32B & 0.976 & 0.928 \\
JoyAI-Image~\cite{joyai_image} & 16B & 0.963 & 0.963 \\
\textbf{SenseNova-U1} & \textbf{8BA3B} & \textbf{0.950} & \textbf{0.955} \\
Qwen-Image~\cite{wu2025qwenimagetechnicalreport} & 20B & 0.943 & 0.946 \\
Z-Image~\cite{cai2025z} & 6B & 0.935 & 0.936 \\
LongCat-Next~\cite{team2026longcat} & 68BA3B & 0.932 & 0.891 \\
X-Omni~\cite{geng2025x} & 12B & 0.900 & 0.814 \\
InternVL-U~\cite{tian2026internvludemocratizingunifiedmultimodal} & 1.7B & 0.738 & 0.860 \\
NEO-unify~\cite{sensenova2026neounify} & 2B & 0.748 & 0.495 \\
FLUX.1-dev~\cite{flux2024} & 12B & 0.607 & 0.005 \\
OmniGen2~\cite{wu2025omnigen2} & 4B & 0.561 & 0.059 \\
Lumina-DiMOO~\cite{xin2025lumina} & 8B & 0.437 & 0.047 \\
BAGEL~\cite{deng2025bagel} & 7B & 0.373 & 0.310 \\
\bottomrule
\end{tabular}
\end{adjustbox}
\end{table}

\textbf{Text-centric Generation.}
We evaluate text-centric generation, with a focus on long-text rendering, multi-region text generation, and complex text-conditioned instruction following. We conduct experiments on CVTG-2K~\cite{du2025textcrafter} and LongText-Bench~\cite{geng2025x}. Across these benchmarks, SenseNova-U1 shows consistently strong performance, indicating that it has reached a text-centric generation capability comparable to the strongest current text-to-image models.

\begin{table*}[t]
\centering
\caption{\textbf{Quantitative evaluation results on IGenBench}. 
The parameters of the generation component are denoted as \textit{\# Params}; \textit{A} in this column denotes activated parameters, e.g., 8BA3B means 8B total generation parameters with 3B activated during inference.
}
\label{tab:igenbench_results}
\small
\setlength{\tabcolsep}{2pt}
\renewcommand{\arraystretch}{1}
\begin{tabular}{l c|cccccccccc|cc}
\toprule
\multirow{2}{*}{\textbf{Model}} &
\multirow{2}{*}{\textbf{\# Params}} &
\multicolumn{10}{c|}{\textbf{Question Type}} &
\multicolumn{2}{c}{\textbf{Overall}} \\
\cmidrule(lr){3-12} \cmidrule(lr){13-14}
 & & \scriptsize\faCalculator~Comp.
 & \scriptsize\faCode~Enc.
 & \scriptsize\faChartLine~Order
 & \scriptsize\faMarker~Marks
 & \scriptsize\faEdit~Anno.
 & \scriptsize\faRuler~Axes
 & \scriptsize\faListUl~Leg.
 & \scriptsize\faChartPie~Chart
 & \scriptsize\faHeading~Title
	 & \scriptsize\faGem~Deco.
	 & Q-ACC$\uparrow$ & I-ACC \\
\midrule
\multicolumn{14}{c}{\textit{Closed-source Models}} \\
\midrule
Nano-Banana-Pro~\cite{deepmind_gemini3proimage_2025}
& -
& 0.84 & 0.86 & 0.90 & 0.87 & 0.93 & 0.93 & 0.96 & 0.92 & 0.98 & 0.94
& 0.90 & 0.49 \\
Seedream-4.5~\cite{seedream45}
& -
& 0.34 & 0.37 & 0.47 & 0.48 & 0.70 & 0.70 & 0.81 & 0.68 & 0.95 & 0.84
& 0.61 & 0.06 \\
GPT-Image-1.5~\cite{GPT-Image-1.5}
& -
& 0.38 & 0.48 & 0.44 & 0.57 & 0.50 & 0.54 & 0.57 & 0.68 & 0.60 & 0.80
& 0.55 & 0.12 \\
Nano-Banana~\cite{google2025gemini25flashmodelcard}
& -
& 0.18 & 0.31 & 0.27 & 0.44 & 0.54 & 0.57 & 0.52 & 0.60 & 0.65 & 0.81
& 0.48 & 0.02 \\
P-Image~\cite{pimage}
& -
& 0.08 & 0.15 & 0.19 & 0.27 & 0.36 & 0.28 & 0.54 & 0.43 & 0.58 & 0.68
& 0.34 & 0.00 \\
Image-01~\cite{minimax_image01}
& -
& 0.01 & 0.05 & 0.04 & 0.10 & 0.10 & 0.14 & 0.03 & 0.22 & 0.14 & 0.47
& 0.13 & 0.00 \\
\midrule
\multicolumn{14}{c}{\textit{Open-source Models}} \\
\midrule
\textbf{SenseNova-U1}
& \textbf{8B}
& 0.27 & 0.23 & 0.49 & 0.45 & 0.54 & 0.61 & 0.70 & 0.65 & 0.74 & 0.82
& \textbf{0.51} & \textbf{0.04} \\
\textbf{SenseNova-U1}
& \textbf{8BA3B}
& 0.17 & 0.22 & 0.33 & 0.41 & 0.36 & 0.49 & 0.56 & 0.60 & 0.55 & 0.78
& \textbf{0.42} & \textbf{0.02} \\
Qwen-Image~\cite{wu2025qwenimagetechnicalreport}
& 20B
& 0.10 & 0.13 & 0.19 & 0.29 & 0.43 & 0.37 & 0.51 & 0.48 & 0.56 & 0.78
& 0.36 & 0.01 \\
Z-Image-Turbo~\cite{cai2025z}
& 6B
& 0.10 & 0.16 & 0.16 & 0.25 & 0.38 & 0.31 & 0.58 & 0.42 & 0.61 & 0.73
& 0.35 & 0.00 \\
HiDream-I1~\cite{cai2025hidream}
& 17B
& 0.01 & 0.03 & 0.03 & 0.10 & 0.07 & 0.14 & 0.10 & 0.26 & 0.19 & 0.20
& 0.11 & 0.00 \\
FLUX.1-dev~\cite{flux2024}
& 12B
& 0.00 & 0.03 & 0.01 & 0.08 & 0.06 & 0.06 & 0.01 & 0.24 & 0.09 & 0.39
& 0.10 & 0.00 \\
\bottomrule
\end{tabular}
\end{table*}

\begin{table}[h!]
\centering
\small
\caption{\textbf{Quantitative evaluation results on BizGenEval}. Each cell reports hard / easy testset scores.}
\label{tab:bizgeneval_results}
\resizebox{0.9\linewidth}{!}{%
\begin{tabular}{lc|cccc|c}
\toprule
\textbf{Model} & \textbf{\# Params} & \textbf{Layout} & \textbf{Attribute} & \textbf{Text} & \textbf{Knowledge} & \textbf{Average}$\uparrow$ \\
\midrule
\multicolumn{7}{c}{\textit{Closed-source Models}} \\
\midrule
GPT-Image-2~\cite{gpt_image_2} & - & 88.5 / 95.3 & 81.9 / 91.0 & 83.9 / 92.6 & 74.2 / 90.9 & 82.1 / 92.5 \\
Nano-Banana-Pro~\cite{deepmind_gemini3proimage_2025} & - & 72.2 / 91.2 & 65.6 / 92.2 & 86.4 / 95.0 & 82.6 / 96.2 & 76.7 / 93.7 \\
Nano-Banana-2.0~\cite{nanobanana2} & - & 68.4 / 91.0 & 57.4 / 91.6 & 83.4 / 94.6 & 64.6 / 93.0 & 68.5 / 92.5 \\
Seedream-5.0~\cite{Seedream5} & - & 67.6 / 89.0 & 42.4 / 77.2 & 43.4 / 75.6 & 41.8 / 75.2 & 48.8 / 79.2 \\
GPT-Image-1.5~\cite{GPT-Image-1.5} & - & 51.6 / 84.8 & 25.8 / 75.2 & 40.4 / 82.8 & 26.0 / 83.6 & 35.9 / 81.6 \\
Seedream-4.5~\cite{seedream45} & - & 35.4 / 71.6 & 22.4 / 62.8 & 41.4 / 72.4 & 21.4 / 58.2 & 30.1 / 66.2 \\
Wan2.6-T2I~\cite{wan2025} & - & 46.4 / 80.6 & 16.6 / 60.6 & 12.6 / 52.6 & 12.2 / 41.0 & 21.9 / 58.7 \\
Seedream-4.0~\cite{seedream2025seedream} & - & 27.6 / 73.4 & 11.4 / 59.2 & 11.4 / 52.8 & 6.8 / 54.8 & 14.3 / 60.1 \\
GPT-Image-1~\cite{GPT-Image-1} & - & 21.4 / 60.2 & 6.8 / 48.6 & 8.6 / 41.0 & 7.8 / 60.0 & 11.2 / 52.4 \\
\midrule
\multicolumn{7}{c}{\textit{Open-source Models}} \\
\midrule
\textbf{SenseNova-U1} & \textbf{8B} & 61.6 / 81.6 & 47.5 / 72.8 & 46.3 / 74.6 & 3.5 / 17.9 & \textbf{39.7 / 61.7} \\
\textbf{SenseNova-U1} & \textbf{8BA3B} & 50.9 / 76.7 & 35.5 / 60.9 & 24.5 / 58.7 & 2.0 / 11.5 & \textbf{28.2 / 51.9} \\
Emu3.5~\cite{cui2025emu35nativemultimodalmodels} & 32B & 30.4 / 63.4 & 14.2 / 52.6 & 7.0 / 33.6 & 1.2 / 11.0 & 13.2 / 40.2 \\
HunyuanImage-3.0~\cite{cao2025hunyuanimage} & 80BA13B & 27.8 / 65.0 & 13.8 / 53.6 & 10.2 / 39.6 & 0.0 / 2.0 & 13.0 / 40.1 \\
Z-Image~\cite{cai2025z} & 6B & 26.8 / 69.2 & 2.6 / 47.6 & 2.8 / 45.0 & 0.6 / 13.2 & 8.2 / 43.8 \\
Qwen-Image-2512~\cite{wu2025qwenimagetechnicalreport} & 20B & 22.2 / 70.6 & 1.2 / 47.8 & 1.8 / 39.2 & 0.0 / 6.4 & 6.3 / 41.0 \\
FLUX.2-dev~\cite{flux-2-2025} & 32B & 17.2 / 67.8 & 1.2 / 49.2 & 1.0 / 43.0 & 0.0 / 8.2 & 4.9 / 42.0 \\
Qwen-Image~\cite{wu2025qwenimagetechnicalreport} & 20B & 10.4 / 51.2 & 0.2 / 22.2 & 0.6 / 17.6 & 0.0 / 4.4 & 2.8 / 23.8 \\
GLM-Image~\cite{glm_image} & 7B & 5.4 / 43.2 & 0.0 / 13.4 & 0.2 / 4.4 & 0.0 / 0.4 & 1.4 / 15.3 \\
LongCat-Image~\cite{team2025longcat} & 6B & 2.4 / 35.8 & 0.2 / 11.6 & 0.0 / 4.4 & 0.0 / 0.0 & 0.7 / 13.0 \\
X-Omni~\cite{geng2025x} & 12B & 2.0 / 22.8 & 0.0 / 5.6 & 0.0 / 8.0 & 0.0 / 1.4 & 0.5 / 9.4 \\
BAGEL~\cite{deng2025bagel} & 7B & 0.6 / 12.8 & 0.0 / 1.6 & 0.0 / 0.0 & 0.0 / 0.2 & 0.2 / 3.7 \\
\bottomrule
\end{tabular}
}
\end{table}

\paragraph{CVTG-2K.}
CVTG-2K~\cite{du2025textcrafter} evaluates complex text-centric generation with multiple text regions. As shown in Table~\ref{tab:cvtg_results}, SenseNova-U1 performs well on this benchmark. Impressively, SenseNova-U1-8B-MoT achieves the best average word accuracy of 0.940, with consistently strong results across settings from 2 to 5 text regions. These results demonstrate that SenseNova-U1 can accurately render textual content under dense multi-region settings.

\paragraph{LongText-Bench.}
LongText-Bench~\cite{geng2025x} primarily evaluates the accuracy and stability of long-text generation in both English and Chinese. 
As shown in Table~\ref{tab:longtext_results}, SenseNova-U1-8B-MoT achieves 0.979 on LongText-Bench-EN and 0.962 on LongText-Bench-ZH, while SenseNova-U1-A3B-MoT also reaches 0.950 and 0.955, despite it not being fully converged. These results indicate that our model can accurately render long-form text in both languages while maintaining high readability and semantic accuracy as text length and structural complexity increase.

\begin{table}[h!]
\centering
\caption{\textbf{Quantitative evaluation results on WISE}.
The parameters of the generation component are denoted as \textit{\# Params}; \textit{A} in this column denotes activated parameters, e.g., 8BA3B means 8B total generation parameters with 3B activated during inference.
}
\label{tab:wise_results}
\renewcommand{\arraystretch}{1}
\setlength{\tabcolsep}{3pt}
\begin{adjustbox}{width=0.84\textwidth}
\begin{tabular}{lc|cccccc|c}
\toprule
\textbf{Model} & \textbf{\# Params} & \textbf{Cultural} & \textbf{Time} & \textbf{Space} & \textbf{Biology} & \textbf{Physics} & \textbf{Chemistry} & \textbf{Overall}$\uparrow$ \\
\midrule
\multicolumn{9}{c}{\textit{Closed-source Models}} \\
\midrule
Nano-Banana-Pro~\cite{deepmind_gemini3proimage_2025} & -  &0.89& 0.80& 0.89& 0.88 &0.86 &0.85& 0.87\\
GPT-Image-1~\cite{GPT-Image-1} & - & 0.81 & 0.71 & 0.89 & 0.83 & 0.79 & 0.74 & 0.80 \\
Seedream 4.0~\cite{seedream2025seedream}&- &0.78& 0.73& 0.85 &0.79& 0.84& 0.67& 0.78 \\
\midrule
\multicolumn{9}{c}{\textit{Open-source Models}} \\
\midrule
\textbf{SenseNova-U1-SFT (w/ CoT)} & \textbf{8BA3B} & 0.81 & 0.77 & 0.84 & 0.81 & 0.84 & 0.82 & \textbf{0.81} \\
\textbf{SenseNova-U1-SFT (w/ CoT)} & \textbf{8B} & 0.78 & 0.73 & 0.82 & 0.80 & 0.85 & 0.77 & \textbf{0.78} \\
\textbf{SenseNova-U1-SFT} & \textbf{8BA3B} & 0.73 & 0.69 & 0.80 & 0.73 & 0.82 & 0.69 & \textbf{0.74} \\
NEO-unify (w/ CoT)~\cite{sensenova2026neounify} & 8B & 0.73 & 0.67 & 0.79 & 0.70 & 0.75 & 0.66 & 0.72 \\
BAGEL (w/ CoT)~\cite{deng2025bagel} & 7B & 0.76 & 0.69 & 0.75 & 0.65  &0.75  &0.58  &0.70 \\
\textbf{SenseNova-U1-SFT} & \textbf{8B} & 0.65 & 0.66 & 0.82 & 0.68 & 0.81 & 0.66 & \textbf{0.69} \\
Qwen-Image~\cite{wu2025qwenimagetechnicalreport} & 20B & 0.63 & 0.62 & 0.76 & 0.60 & 0.72 & 0.39 & 0.63 \\
BLIP3-o~\cite{chen2025blip3o} & 1.4B & - & - & - & - & - & - & 0.62 \\
NEO-unify (w/ CoT)~\cite{sensenova2026neounify} & 2B & 0.59 & 0.54 & 0.68 & 0.57 & 0.69 & 0.50 & 0.59 \\
InternVL-U (w/ CoT)~\cite{tian2026internvludemocratizingunifiedmultimodal} & 1.7B & 0.55 & 0.57 & 0.74 & 0.51 & 0.72 & 0.46 & 0.58 \\
Emu3.5~\cite{cui2025emu35nativemultimodalmodels} & 32B & - & - & - & - & - & - & 0.58 \\
LongCat-Next~\cite{team2026longcat} & 68BA3B & - & - & - & - & - & - & 0.57 \\
UniWorld-V1~\cite{lin2025uniworld} & 12B & 0.53 & 0.55 & 0.73 & 0.45 & 0.59 & 0.41 & 0.55 \\
FLUX.1-dev~\cite{flux2024} & 12B & 0.48 & 0.58 & 0.62 & 0.42 & 0.51 & 0.35 & 0.50 \\
BAGEL~\cite{deng2025bagel} & 7B & 0.44 & 0.52 & 0.65 & 0.42 & 0.62 & 0.41 & 0.49 \\
NEO-unify~\cite{sensenova2026neounify} & 8B & - & - & - & - & - & - & 0.47 \\
InternVL-U~\cite{tian2026internvludemocratizingunifiedmultimodal} & 1.7B & 0.37 & 0.51 & 0.68 & 0.39 & 0.62 & 0.39 & 0.46 \\
SD3-Medium~\cite{esser2024scaling} & 2B & 0.43 & 0.50 & 0.52 & 0.41 & 0.53 & 0.33 & 0.45 \\
Ovis-U1~\cite{wang2025ovis} & 1.2B & 0.36 & 0.46 & 0.64 & 0.35 & 0.52 & 0.28 & 0.42 \\
NEO-unify~\cite{sensenova2026neounify} & 2B & - & - & - & - & - & - & 0.41 \\
Lumina-DiMOO~\cite{xin2025lumina} & 8B & 0.35 & 0.43 & 0.59 & 0.31 & 0.49 & 0.34 & 0.40 \\
Janus-Pro~\cite{chen2025janus} & 7B & 0.30 & 0.37 & 0.49 & 0.36 & 0.42 & 0.26 & 0.35 \\
\bottomrule
\end{tabular}
\end{adjustbox}
\end{table}

\textbf{Complex Infographic Generation.}
Here we evaluate complex infographic and commercial visual content generation on IGenBench~\cite{tang2026igenbench} and BizGenEval~\cite{li2026bizgeneval}. Compared with general text-to-image generation, these infographic tasks are substantially more challenging, since the model must not only generate correct text and visual elements, but also satisfy structured layouts, chart expression, and multiple semantic constraints at the same time.

\paragraph{IGenBench.}
IGenBench~\cite{tang2026igenbench} evaluates the reliability of text-to-infographic generation, requiring models to jointly satisfy textual, chart, data, and structural constraints. As shown in Table~\ref{tab:igenbench_results}, SenseNova-U1 achieves the strongest performance among open-source models, substantially outperforming Qwen-Image and Z-Image-Turbo while remaining competitive with several closed-source systems. These results demonstrate the strong infographic generation reliability of SenseNova-U1, although truly robust infographic synthesis remains challenging for current models.

\paragraph{BizGenEval.}
BizGenEval~\cite{li2026bizgeneval} evaluates visual generation in real-world commercial scenarios across dimensions, including Layout, Attribute, Text, and Knowledge. As shown in Table~\ref{tab:bizgeneval_results}, SenseNova-U1 achieves the best hard-split average among all open-source models while remaining competitive on the easy split. In particular, our models exhibit strong layout planning, attribute control, and text rendering capabilities, highlighting the potential of native unified multimodal modeling for complex professional visual content generation under multi-constraint settings.

\textbf{Reasoning-centric Generation.}
We further evaluate reasoning-driven WISE~\cite{niu2025wise}. It examines whether a model can effectively utilize internal world knowledge and combine it with reasoning during image generation, thereby handling tasks involving cultural commonsense, temporal understanding, spatial understanding, and scientific knowledge. 

\paragraph{WISE.}
As shown in Table~\ref{tab:wise_results}, SenseNova-U1 displays a clear advantage on WISE. Even without chain-of-thought (CoT), SenseNova-U1-A3B-MoT-SFT substantially outperforms representative open-source models such as Qwen-Image, BAGEL, and InternVL-U, while CoT further boosts performance to a level competitive with closed-source systems. A similar trend is observed for the 8B variant, indicating that reasoning-enhanced generation scales consistently across model sizes. Notably, our models excel in Cultural, Biology, Physics, and Chemistry, particularly in science-oriented tasks requiring explicit knowledge retrieval and multi-step reasoning. This suggests that SenseNova-U1 effectively translates structured semantic reasoning into accurate and knowledge-consistent visual generation.

\subsubsection{Image Editing}

We evaluate SenseNova-U1 on image editing across three distinct settings: single-image editing, multi-image editing, and reasoning-driven editing. Collectively, these evaluations probe the model’s ability to follow complex instructions, perform realistic human-guided modifications, and synthesize knowledge-aware visual content.

\begin{table}[t]
\centering
\caption{\textbf{Quantitative evaluation results on ImgEdit}.
The parameters of the generation component are denoted as \textit{\# Params}; \textit{A} in this column denotes activated parameters, e.g., 8BA3B means 8B total generation parameters with 3B activated during inference.
}
\label{tab:imgedit_results}
\renewcommand{\arraystretch}{1}
\setlength{\tabcolsep}{2pt}
\begin{adjustbox}{width=\textwidth}
\begin{tabular}{lc|ccccccccc|c}
\toprule
\textbf{Model} & \textbf{\# Params} & \textbf{Add} & \textbf{Adjust} & \textbf{Extract} & \textbf{Replace} & \textbf{Remove} & \textbf{Background} & \textbf{Style} & \textbf{Hybrid} & \textbf{Action} & \textbf{Overall}$\uparrow$ \\
\midrule
\multicolumn{12}{c}{\textit{Closed-source Models}} \\
\midrule
UniWorld-V2~\cite{li2025uniworld} & - & 4.29 & 4.44 & 4.32 & 4.69 & 4.72 & 4.41 & 4.91 & 3.83 & 4.83 & 4.49 \\
Nano-Banana-Pro~\cite{deepmind_gemini3proimage_2025} & - & 4.44 & 4.62 & 3.42 & 4.60 & 4.63 & 4.32 & 4.97 & 3.64 & 4.69 & 4.37 \\
Seedream 4.5~\cite{seedream45} & - & 4.57 & 4.65 & 2.97 & 4.66 & 4.46 & 4.37 & 4.92 & 3.71 & 4.56 & 4.32 \\
Seedream 4.0~\cite{seedream2025seedream} & - & 4.33 & 4.38 & 3.89 & 4.65 & 4.57 & 4.35 & 4.22 & 3.71 & 4.61 & 4.30 \\
Nano-Banana~\cite{google2025gemini25flashmodelcard} & - & 4.62 & 4.41 & 3.68 & 4.34 & 4.39 & 4.40 & 4.18 & 3.72 & 4.83 & 4.29 \\
GPT-Image-1~\cite{GPT-Image-1} & - & 4.61 & 4.33 & 2.90 & 4.35 & 3.66 & 4.57 & 4.93 & 3.96 & 4.89 & 4.20 \\
FLUX.1 Kontext [Pro]~\cite{labs2025flux1kontextflowmatching} & - & 4.25 & 4.15 & 2.35 & 4.56 & 3.57 & 4.26 & 4.57 & 3.68 & 4.63 & 4.00 \\
\midrule
\multicolumn{12}{c}{\textit{Open-source Models}} \\
\midrule
Qwen-Image-Edit-2511~\cite{wu2025qwenimagetechnicalreport} & 20B & 4.54 & 4.57 & 4.13 & 4.70 & 4.46 & 4.36 & 4.89 & 4.16 & 4.81 & 4.51 \\
LongCat-Image-Edit~\cite{team2025longcat} & 6B & 4.44 & 4.53 & 3.83 & 4.80 & 4.60 & 4.33 & 4.92 & 3.75 & 4.82 & 4.45 \\
Emu3.5~\cite{cui2025emu35nativemultimodalmodels} & 32B & 4.61 & 4.32 & 3.96 & 4.84 & 4.58 & 4.35 & 4.79 & 3.69 & 4.57 & 4.41 \\
FLUX.2 [Dev]~\cite{flux-2-2025} & 32B & 4.50 & 4.18 & 3.83 & 4.65 & 4.65 & 4.31 & 4.88 & 3.46 & 4.70 & 4.35 \\
Qwen-Image-Edit-2509~\cite{wu2025qwenimagetechnicalreport} & 20B & 4.32 & 4.36 & 4.04 & 4.64 & 4.52 & 4.37 & 4.84 & 3.39 & 4.71 & 4.35 \\
Z-Image-Edit~\cite{cai2025z} & 6B & 4.40 & 4.14 & 4.30 & 4.57 & 4.13 & 4.14 & 4.85 & 3.63 & 4.50 & 4.30 \\
Qwen-Image-Edit~\cite{wu2025qwenimagetechnicalreport} & 20B & 4.38 & 4.16 & 3.43 & 4.66 & 4.14 & 4.38 & 4.81 & 3.82 & 4.69 & 4.27 \\
Ovis-U1~\cite{wang2025ovis} & 1.2B & 3.99 & 3.73 & 2.66 & 4.38 & 4.15 & 4.05 & 4.86 & 3.43 & 4.68 & 3.97 \\
\textbf{SenseNova-U1} & \textbf{8BA3B} & 4.03 & 4.10 & 2.73 & 4.27 & 3.91 & 4.06 & 4.92 & 2.87 & 4.29 & \textbf{3.91} \\
\textbf{SenseNova-U1} & \textbf{8B} & 3.83 & 4.15 & 3.12 & 4.32 & 3.26 & 4.18 & 4.85 & 3.03 & 4.41 & \textbf{3.90} \\
InternVL-U (w/ CoT)~\cite{tian2026internvludemocratizingunifiedmultimodal} & 1.7B & 4.24 & 3.80 & 2.58 & 4.36 & 3.51 & 3.92 & 4.69 & 3.00 & 4.31 & 3.82 \\
FLUX.1 Kontext [Dev]~\cite{labs2025flux1kontextflowmatching} & 12B & 4.12 & 3.80 & 2.04 & 4.22 & 3.09 & 3.97 & 4.51 & 3.35 & 4.25 & 3.71 \\
InternVL-U~\cite{tian2026internvludemocratizingunifiedmultimodal} & 1.7B & 4.13 & 3.40 & 2.27 & 4.13 & 3.39 & 3.84 & 4.77 & 3.03 & 4.05 & 3.67 \\
OmniGen2~\cite{wu2025omnigen2} & 4B & 3.57 & 3.06 & 1.77 & 3.74 & 3.20 & 3.57 & 4.81 & 2.52 & 4.68 & 3.44 \\
UniWorld-V1~\cite{lin2025uniworld} & 12B & 3.82 & 3.64 & 2.27 & 3.47 & 3.24 & 2.99 & 4.21 & 2.96 & 2.74 & 3.26 \\
BAGEL~\cite{deng2025bagel} & 7B & 3.56 & 3.31 & 1.70 & 3.30 & 2.62 & 3.24 & 4.49 & 2.38 & 4.17 & 3.20 \\
Step1X-Edit~\cite{liu2025step1x} & 12B & 3.88 & 3.14 & 1.76 & 3.40 & 2.41 & 3.16 & 4.63 & 2.64 & 2.52 & 3.06 \\
ICEdit~\cite{zhang2025context} & 12B & 3.58 & 3.39 & 1.73 & 3.15 & 2.93 & 3.08 & 3.84 & 2.04 & 3.68 & 3.05 \\
OmniGen~\cite{xiao2024omnigen} & 3.8B & 3.47 & 3.04 & 1.71 & 2.94 & 2.43 & 3.21 & 4.19 & 2.24 & 3.38 & 2.96 \\
\bottomrule
\end{tabular}
\end{adjustbox}
\end{table}

\textbf{General Editing.}
We further evaluate SenseNova-U1 on general image editing. Compared with text-to-image generation, image editing requires the model not only to follow textual instructions, but also to preserve the original image content, structure, and visual style while performing precise local or global modifications. This places substantially higher demands on instruction following, content preservation, and fine-grained controllability.

\paragraph{ImgEdit.}
ImgEdit~\citep{ye2025imgedit} provides a fine-grained evaluation of image editing across diverse dimensions, including Add, Adjust, Replace, Remove, Background, Style, Hybrid, and Action. As shown in Table~\ref{tab:imgedit_results}, both SenseNova-U1-A3B-MoT and SenseNova-U1-8B-MoT achieve decent overall performance, outperforming existing open-source unified editing systems while remaining competitive with several specialized editing models.
A performance gap nevertheless remains relative to the strongest dedicated editing approaches, particularly on complex hybrid edits and scenarios requiring precise content preservation under substantial transformations. 
\textit{We attribute this gap primarily to limitations in the current editing data, which remains dominated by open-source resources and lacks sufficiently diverse editing pipelines and large-scale preference-aligned optimization.}
Despite these limitations, the results indicate that SenseNova-U1 already provides a strong general-purpose foundation for image editing within a native unified framework. We expect future improvements to arise naturally from richer editing data, stronger editing-oriented supervision, and reinforcement learning strategies more closely aligned with human editing preferences and long-horizon editing objectives.

\paragraph{GEdit-Bench.}
GEdit-Bench~\citep{liu2025step1x} evaluates general instruction-based image editing with a stronger emphasis on overall editing quality and prompt consistency. As shown in Table~\ref{tab:gedit_results}, both SenseNova-U1-A3B-MoT and SenseNova-U1-8B-MoT achieve competitive performance against representative editing models such as Emu-3.5, Z-Image-Edit, and Qwen-Image-Edit. While specialized editing systems still retain an advantage in highly optimized editing workflows, SenseNova-U1 demonstrates strong generalization across diverse editing instructions, maintaining coherent semantics and stable visual consistency under a unified native modeling framework.

\begin{table}[t]
\centering
\caption{\textbf{Quantitative evaluation results on GEdit-Bench}.}
\label{tab:gedit_results}
\renewcommand{\arraystretch}{1}
\setlength{\tabcolsep}{3pt}
\begin{adjustbox}{width=0.53\textwidth}
\begin{tabular}{lc|ccc}
\toprule
\multirow{2}{*}{\textbf{Model}} & \multirow{2}{*}{\textbf{\# Params}} & \multicolumn{3}{c}{\textbf{GEdit-Bench-EN}} \\
\cmidrule(lr){3-5}
& & G\_SC  & G\_PQ & G\_O $\uparrow$ \\
\midrule
\multicolumn{5}{c}{\textit{Closed-source Models}} \\
\midrule
UniWorld-V2~\cite{li2025uniworld} & - & 8.39 & 8.02 & 7.83 \\
Seedream 4.5~\cite{seedream45} & - & 8.27 & 8.17 & 7.82 \\
Nano-Banana-Pro~\cite{deepmind_gemini3proimage_2025} & - & 8.10 & 8.34 & 7.74 \\
Seedream 4.0~\cite{seedream2025seedream} & - & 8.14 & 8.12 & 7.70 \\
GPT-Image-1~\cite{GPT-Image-1} & - & 7.85 & 7.62 & 7.53 \\
Nano-Banana~\cite{google2025gemini25flashmodelcard} & - & 7.40 & 8.45 & 7.29 \\
FLUX.1 Kontext [Pro]~\cite{labs2025flux1kontextflowmatching} & - & 7.02 & 7.60 & 6.56 \\
\midrule
\multicolumn{5}{c}{\textit{Open-source Models}} \\
\midrule
Qwen-Image-Edit-2511~\cite{wu2025qwenimagetechnicalreport} & 20B & 8.30 & 8.20 & 7.88 \\
Longcat-Image-Edit~\cite{team2025longcat} & 6B & 8.13 & 8.18 & 7.75 \\
Emu3.5~\cite{cui2025emu35nativemultimodalmodels} & 32B & 8.11 & 7.70 & 7.59 \\
Z-Image-Edit~\cite{cai2025z} & 6B & 8.11 & 7.72 & 7.57 \\
Qwen-Image-Edit~\cite{wu2025qwenimagetechnicalreport} & 20B & 8.00 & 7.86 & 7.56 \\
Qwen-Image-Edit-2509~\cite{wu2025qwenimagetechnicalreport} & 20B & 8.15 & 7.86 & 7.54 \\
\textbf{SenseNova-U1} & \textbf{8B} & 8.27 & 7.49 & \textbf{7.47} \\
FLUX.2 [Dev]~\cite{flux-2-2025} & 32B & 7.84 & 8.06 & 7.41 \\
\textbf{SenseNova-U1} & \textbf{8BA3B} & 8.07 & 7.36 & \textbf{7.32} \\
Step1X-Edit~\cite{liu2025step1x} & 12B & 7.66 & 7.35 & 6.97 \\
BAGEL~\cite{deng2025bagel} & 7B & 7.36 & 6.83 & 6.52 \\
OmniGen2~\cite{wu2025omnigen2} & 4B & 7.16 & 6.77 & 6.41 \\
FLUX.1 Kontext [Dev]~\cite{labs2025flux1kontextflowmatching} & 12B & 6.52 & 7.38 & 6.00 \\
OmniGen~\cite{xiao2024omnigen} & 3.8B & 5.96 & 5.89 & 5.06 \\
UniWorld-V1~\cite{lin2025uniworld} & 12B & 4.93 & 7.43 & 4.85 \\
\bottomrule
\end{tabular}
\end{adjustbox}
\end{table}

\textbf{Reasoning-centric Editing.}
Beyond general editing ability, we further evaluate our SenseNova-U1 on reasoning-driven image editing. Compared with standard editing tasks, these scenarios are substantially more challenging because the model must first infer implicit temporal, causal, spatial, or logical relationships from the prompt instruction, and then translate the reasoning outcome into precise and visually consistent modifications. 

\paragraph{RISEBench.}
RISEBench~\cite{zhao2025envisioning} mainly evaluates reasoning-informed image editing, covering four types of reasoning-centric editing tasks: Temporal, Causal, Spatial, and Logical, together with auxiliary metrics such as IR, AC, and VP. As shown in Table~\ref{tab:rise_results}, SenseNova-U1-A3B-MoT-SFT achieves an overall score of 25.3 without CoT, substantially outperforming most open-source unified editing models. With CoT enabled, SenseNova-U1-A3B-MoT-SFT further improves to 30.0, reaching the best level among the open-source methods in our comparison. A similar trend also appears for SenseNova-U1-8B-MoT-SFT, which improves from 23.9 to 26.9, indicating that this pattern remains stable across model scales. A closer look shows that the gains from CoT are especially notable on dimensions that rely more heavily on explicit reasoning, such as Causal and Logical. For example, the Logical score of SenseNova-U1-A3B-MoT-SFT improves from 7.1 to 20.0. This suggests that the model can use an explicit reasoning process to better decompose complex editing goals. Overall, these results indicate that the advantage of SenseNova-U1 lies not only in executing edits, but also in performing the necessary understanding and inference before editing, which leads to stronger performance on reasoning-centric image editing than existing unified multimodal models.

\begin{table}[t]
\centering
\caption{\textbf{Quantitative evaluation results on RISEBench}.
The parameters of the generation component are denoted as \textit{\# Params}; \textit{A} in this column denotes activated parameters, e.g., 8BA3B means 8B total generation parameters with 3B activated during inference.}
\label{tab:rise_results}
\begin{adjustbox}{width=0.95\textwidth}
\begin{tabular}{lc|cccc|c|ccc}
\toprule
\cellcolor{white}{\textbf{Model}} & \textbf{\# Params} & \textbf{Temporal} & \textbf{Causal} & \textbf{Spatial} & \textbf{Logical} & \textbf{Overall}$\uparrow$ & \textbf{IR} & \textbf{AC} & \textbf{VP} \\
\midrule

\multicolumn{10}{c}{\textit{Closed-source Models}} \\
\midrule
GPT-Image-1.5~\cite{GPT-Image-1.5} & - & 54.1 & 60.0 & 62.0 & 21.2 & 50.0 & 69.7 & 92.5 & 94.9 \\
Nano-Banana-Pro~\cite{deepmind_gemini3proimage_2025} & - & 41.2 & 61.1 & 48.0 & 37.6 & 47.2 & 77.0 & 85.5 & 94.4 \\
Nano-Banana~\cite{google2025gemini25flashmodelcard} & - & 25.9 & 47.8 & 37.0 & 18.8 & 32.8 & 61.2 & 86.0 & 91.3 \\
GPT-Image-1~\cite{GPT-Image-1} & - & 34.1 & 32.2 & 37.0 & 10.6 & 28.9 & 62.8 & 80.2 & 94.9 \\
GPT-Image-1-mini~\cite{GPT-Image-1} & - & 24.7 & 28.9 & 33.0 & 9.4 & 24.4 & 54.1 & 71.5 & 93.7 \\
Gemini-2.0-Flash-exp~\cite{gemini-2.0-flash} & - & 8.2 & 15.5 & 23.0 & 4.7 & 13.3 & 48.9 & 68.2 & 82.7 \\
Seedream 4.0~\cite{seedream2025seedream} & - & 12.9 & 12.2 & 11.0 & 7.1 & 10.8 & 58.9 & 67.4 & 91.2 \\
Gemini-2.0-Flash-pre~\cite{gemini-2.0-flash} & - & 10.6 & 13.3 & 11.0 & 2.3 & 9.4 & 49.9 & 68.4 & 84.9 \\

\midrule
\multicolumn{10}{c}{\textit{Open-source Models}} \\
\midrule
\textbf{SenseNova-U1-SFT (w/ CoT)} & \textbf{8BA3B} & 24.7 & 46.7 & 28.0 & 20.0 & \textbf{30.0} & 63.2 & 84.1 & 87.4 \\
\textbf{SenseNova-U1-SFT (w/ CoT)} & \textbf{8B} & 31.8 & 33.3 & 27.0 & 15.3 & \textbf{26.9} & 60.8 & 86.6 & 88.2 \\
\textbf{SenseNova-U1-SFT} & \textbf{8BA3B} & 25.9 & 41.1 & 26.0 & 7.1 & \textbf{25.3} & 57.4 & 82.6 & 85.4 \\
\textbf{SenseNova-U1-SFT} & \textbf{8B} & 22.4 & 33.3 & 27.0 & 11.8 & \textbf{23.9} & 58.2 & 84.1 & 82.4 \\
Qwen-Image-Edit-2511~\cite{wu2025qwenimagetechnicalreport} & 20B & 21.2 & 18.9 & 31.0 & 4.7 & 19.4 & 49.9 & 71.0 & 91.5 \\
BAGEL (w/ CoT)~\cite{deng2025bagel} & 7B & 5.9 & 17.8 & 21.0 & 1.2 & 11.9 & 45.9 & 73.8 & 80.1 \\
InternVL-U (w/ CoT)~\cite{tian2026internvludemocratizingunifiedmultimodal} & 1.7B & 4.7 & 7.8 & 1.8 & 5.9 & 9.4 & 43.9 & 64.4 & 79.7 \\
Qwen-Image-Edit-2509~\cite{wu2025qwenimagetechnicalreport} & 20B & 4.7 & 10.0 & 17.0 & 2.4 & 8.9 & 37.2 & 66.4 & 86.9 \\
BAGEL~\cite{deng2025bagel} & 7B & 2.4 & 5.6 & 14.0 & 1.2 & 6.1 & 36.5 & 53.5 & 73.0 \\
FLUX.1-Kontext-Dev~\cite{labs2025flux1kontextflowmatching} & 12B & 2.3 & 5.5 & 13.0 & 1.2 & 5.8 & 26.0 & 71.6 & 85.2 \\
InternVL-U~\cite{tian2026internvludemocratizingunifiedmultimodal} & 1.7B & 3.5 & 2.2 & 5.0 & 3.5 & 3.6 & 35.6 & 52.7 & 75.9 \\
Ovis-U1~\cite{wang2025ovis} & 1.2B & 1.2 & 3.3 & 4.0 & 2.4 & 2.8 & 33.9 & 52.7 & 72.9 \\
Lumina-DiMOO~\cite{xin2025lumina} & 8B & 2.4 & 1.1 & 4.0 & 1.2 & 2.2 & 34.0 & 50.7 & 72.3 \\
Step1X-Edit~\cite{liu2025step1x} & 12B & 0.0 & 2.2 & 2.0 & 3.5 & 1.9 & 25.1 & 41.5 & 73.5 \\
OmniGen~\cite{xiao2024omnigen} & 3.8B & 1.2 & 1.0 & 0.0 & 1.2 & 0.8 & 22.0 & 32.6 & 55.3 \\
Emu2~\cite{emu2} & 37B & 1.2 & 1.1 & 0.0 & 0.0 & 0.5 & 22.6 & 38.2 & 78.3 \\
\bottomrule
\end{tabular}
\end{adjustbox}
\end{table}

\begin{table*}[t]
\centering
\caption{\textbf{Quantitative evaluation results on OpenING}.
The parameters of the generation component are denoted as \textit{\# Params}; \textit{A} in this column denotes activated parameters, e.g., 8BA3B means 8B total generation parameters with 3B activated during inference.}
\label{tab:opening_results}
\renewcommand{\arraystretch}{1}
\setlength{\tabcolsep}{3pt}
\begin{adjustbox}{width=\textwidth}
\begin{tabular}{lc|ccccccc|c}
\toprule
\textbf{Model} & \textbf{\# Params} & \textbf{Complete} & \textbf{Quality} & \textbf{Richness} & \textbf{Correct} & \textbf{Human Align.} & \textbf{IT Coherency} & \textbf{Multi-step} & \textbf{Overall}$\uparrow$ \\
\midrule
\multicolumn{10}{c}{\textit{Closed-source Models}} \\
\midrule
Nano-Banana~\cite{google2025gemini25flashmodelcard} & - & 9.34 & 8.58 & 8.00 & 9.17 & 8.88 & 9.27 & 8.70 & 8.85 \\
Wan-Weaver~\cite{xing2026wan} & - & 9.41 & 8.32 & 8.03 & 8.90 & 8.69 & 8.78 & 8.56 & 8.67 \\
GPT-4o~\cite{hurst2024gpt}+DALL-E3~\cite{betker2023dalle3} & - & 8.66 & 8.01 & 7.42 & 7.98 & 8.77 & 8.15 & 8.38 & 8.20 \\
Gemini~\cite{team2023gemini}+Flux~\cite{flux2024} & 12B & 7.58 & 7.26 & 6.48 & 7.03 & 7.98 & 6.98 & 7.33 & 7.23 \\
\midrule
\multicolumn{10}{c}{\textit{Open-source Models}} \\
\midrule
\textbf{SenseNova-U1-SFT (w/ CoT)} & \textbf{8BA3B} & 9.27 &	9.11 &	8.45 &	9.16 &	9.40 &	9.55 &	9.21 &	\textbf{9.16}\\
\textbf{SenseNova-U1-SFT (w/ CoT)} & \textbf{8B} & 9.14& 	9.03& 	8.43& 	9.08& 	9.35& 	9.40& 	9.09& \textbf{9.07} \\
SEED-X~\cite{ge2024seed} & 17B & 5.65 & 6.07 & 4.92 & 5.77 & 7.03 & 5.72 & 5.72 & 5.84 \\
Emu3~\cite{wang2024emu3} & 8B & 5.90 & 5.96 & 5.52 & 5.43 & 6.47 & 5.66 & 5.37 & 5.76 \\
Anole~\cite{chern2024anole} & 7B & 6.27 & 6.02 & 5.28 & 5.06 & 6.91 & 4.90 & 5.81 & 5.75 \\
SEED-LLaMA~\cite{ge2023making} & 14B & 5.59 & 5.50 & 4.61 & 4.59 & 6.5 & 4.43 & 5.13 & 5.19 \\
VILA-U~\cite{wu2024vila} & 7B & 5.60 & 5.14 & 4.68 & 4.78 & 5.69 & 4.74 & 4.79 & 5.06 \\
Show-o~\cite{xie2024show} & 1.3B & 4.37 & 4.79 & 3.83 & 3.76 & 5.78 & 4.04 & 4.33 & 4.41 \\
MiniGPT-5~\cite{zheng2023minigpt} & 0.86B & 3.91 & 4.5 & 3.61 & 3.63 & 5.51 & 3.56 & 4.10 & 4.12 \\
NExT-GPT~\cite{wunext} & 1.3B & 3.89 & 4.25 & 3.35 & 3.61 & 5.35 & 3.32 & 3.85 & 3.95 \\

\bottomrule
\end{tabular}
\end{adjustbox}
\end{table*}

\subsubsection{Interleaved Generation}
We further evaluate the model on interleaved generation and unified reasoning, examining whether understanding and generation can reinforce each other within a single framework. These evaluations cover open-ended interleaved generation, generation-assisted multimodal reasoning, and the bidirectional synergy between understanding and generation.

\textbf{Interleaved Generation.} This is no longer to generate a single image in one shot, but to alternately produce text and images in an open-ended output process while maintaining semantic coherence, cross-modal consistency, and overall completeness across multiple generation steps. Here we adopt OpenING~\cite{zhou2025opening} and VBVR-Image (Preview)~\cite{wang2026very}.

\paragraph{OpenING.}
OpenING~\cite{zhou2025opening} evaluates the overall quality of open-ended interleaved image-text generation across dimensions including completeness, quality, richness, correctness, human alignment, image-text coherence, and multi-step consistency. As shown in Table~\ref{tab:opening_results}, SenseNova-U1 demonstrates consistently strong performance under this challenging setting. In particular, SenseNova-U1-A3B-MoT-SFT with CoT achieves the best overall score of 9.16, while SenseNova-U1-8B-MoT-SFT with CoT reaches 9.07, outperforming representative systems such as Nano Banana, Wan-Weaver, and GPT-4o+DALL-E3.
These results suggest that SenseNova-U1 not only produces high-quality unimodal outputs, but also maintains strong semantic coherence, long-range consistency, and instruction fidelity across interleaved multimodal generation trajectories. More importantly, the strong performance under multi-step image-text interaction indicates that our unified framework can effectively coordinate generation and reasoning within a shared native modeling space, rather than treating image synthesis and language generation as isolated processes.

\paragraph{VBVR-Image (Preview).} 
VBVR-Image~\cite{wang2026very} is a recently introduced benchmark derived from VBVR that evaluates reasoning behaviors emerging through visual generation. It extends this setting from video generation to interleaved image generation, where models must generate images to solve visual reasoning tasks such as maze navigation, pattern discovery, and spatial inference. As the benchmark is currently available only in a preview version, we report results on this subset in Table~\ref{tab:vbvr_image_preview}.
The results show that SenseNova-U1 exhibits strong reasoning capability within the generation process itself, outperforming both competitive in-domain trained baselines and several powerful proprietary systems. These findings suggest that the unified native framework of SenseNova-U1 supports not only high-quality generation, but also reasoning behaviors that can emerge and be executed directly through multimodal generation trajectories.

\begin{table*}[t]
\centering
\renewcommand{\arraystretch}{1}
\setlength{\tabcolsep}{3pt}
\caption{\textbf{Quantitative evaluation results on VBVR-Image (Preview)}. \textit{VBVR-} prefix is the model tuned on VBVR training set.}
\label{tab:vbvr_image_preview}
\begin{adjustbox}{width=\textwidth}
\begin{tabular}{lc|c|cccccc|cccccc}
\toprule
\multirow{2}{*}{\textbf{Model}} & \multirow{2}{*}{\textbf{\# Params}} & \multirow{2}{*}{\textbf{Overall}$\uparrow$}
& \multicolumn{6}{c|}{\textbf{In-Domain by Category}} 
& \multicolumn{6}{c}{\textbf{Out-of-Domain by Category}} \\
\cmidrule(lr){4-9}
\cmidrule(lr){10-15}
&&& Avg.
& Abst. & Know. & Perc. & Spat. & Trans. & Avg. & Abst. & Know. & Perc. & Spat. & Trans. \\
\midrule
\multicolumn{15}{c}{\textit{Closed-source Models}} \\
\midrule

Nano-Banana-2~\cite{nanobanana2}
& -
& 62.3 & 61.1 & 64.4 & 49.6 & 78.6 & 51.1 & 53.5
& 63.6 & 83.2 & 63.1 & 61.4 & 61.0 & 54.2 \\ 

GPT-Image-2~\cite{gpt_image_2}
& -
& 60.1 & 57.9 & 62.0 & 46.9 & 70.1 & 45.8 & 58.1
& 62.3 & 82.9 & 63.3 & 61.9 & 48.2 & 55.4 \\

\midrule
\multicolumn{15}{c}{\textit{Open-source Models}} \\
\midrule

\textbf{SenseNova-U1-SFT}
& \textbf{8BA3B}
& \textbf{68.9} & 73.9 & 76.1 & 72.2 & 80.8 & 66.5 & 68.9
& 64.0 & 89.2 & 61.9 & 66.2 & 75.5 & 41.8 \\

\textbf{SenseNova-U1-SFT}
& \textbf{8B}
& \textbf{68.8} & 70.8 & 73.9 & 60.5 & 81.6 & 65.8 & 66.3
& 66.8 & 83.7 & 65.4 & 69.9 & 68.0 & 51.9 \\

VBVR-ThinkMorph~\cite{gu2025thinkmorph}
& 7B
& 63.0 & 64.7 & 67.3 & 61.3 & 66.1 & 64.1 & 62.0
& 61.4 & 84.9 & 55.5 & 60.4 & 69.5 & 48.0 \\

ThinkMorph~\cite{gu2025thinkmorph}
& 7B
& 38.7 & 37.0 & 45.8 & 26.8 & 36.3 & 36.7 & 33.3
& 40.5 & 57.3 & 46.0 & 42.5 & 55.6 & 19.3 \\

VBVR-BAGEL~\cite{deng2025bagel}
& 7B
& 36.5 & 37.0 & 44.1 & 28.6 & 36.7 & 32.2 & 38.8
& 36.0 & 51.9 & 40.3 & 35.6 & 31.9 & 26.6 \\

BAGEL~\cite{deng2025bagel}
& 7B
& 29.1 & 32.1 & 35.5 & 26.6 & 33.1 & 27.3 & 36.1
& 26.0 & 45.3 & 20.1 & 20.5 & 22.0 & 25.9 \\

\bottomrule
\end{tabular}
\end{adjustbox}
\end{table*}

\textbf{Unified Reasoning.}
Beyond one-way generation ability, we further investigate whether the model can achieve genuine bidirectional synergy between understanding and generation. Unlike conventional evaluations that assess these capabilities in isolation, the Generation-aids-Understanding (GaU) components of Uni-MMMU~\cite{zou2025unimmmu} and RealUnify~\cite{shi2025realunify} explicitly examine whether generation can enhance understanding (GEU), and conversely, whether understanding can improve generation (UEG) within a unified multimodal framework.

\paragraph{Uni-MMMU.}
Uni-MMMU~\cite{zou2025unimmmu} (GaU) evaluates whether generation can actively assist multimodal understanding and reasoning. As shown in Table~\ref{tab:unimmmu_gau_results}, SenseNova-U1 achieves strong performance under this setting. SenseNova-U1-8B-MoT-SFT attains a GaU average of 35.0, substantially outperforming unified baselines such as BAGEL, OmniGen2, and Ovis-U1, while SenseNova-U1-A3B-MoT-SFT also achieves a competitive score of 32.6.
This suggests that the generation branch of SenseNova-U1 can provide meaningful support for multimodal reasoning, highlighting the synergistic interaction between generation and understanding within our unified framework.

\paragraph{RealUnify.}
RealUnify~\cite{shi2025realunify} further evaluates bidirectional synergy through both Understanding Enhances Generation (UEG) and Generation Enhances Understanding (GEU). As shown in Table~\ref{tab:realunify_results}, SenseNova-U1 demonstrates clear advantages under both settings. We can observe that SenseNova-U1-8B-MoT-SFT achieves the best overall average of 52.4, including 55.7 on Avg-UEG and 47.5 on Avg-GEU, while SenseNova-U1-A3B-MoT-SFT also attains a strong overall average of 50.5.
These results suggest that SenseNova-U1 achieves genuine synergy between understanding and generation, rather than merely colocating the two capabilities within a shared backbone.

\begin{table}[t]
\centering
\caption{\textbf{Quantitative evaluation results on UniMMMU} (Generation aids Understanding, GaU). For all models, we report text accuracy (T) for all tasks. For multi-step tasks (Maze, Sliding Puzzle), we report sample-level accuracy.}
\label{tab:unimmmu_gau_results}
\renewcommand{\arraystretch}{1}
\setlength{\tabcolsep}{3pt}
\begin{adjustbox}{width=0.68\textwidth}
\begin{tabular}{lc|cccc|c}
\toprule
\textbf{Model} & \textbf{\# Params} & \textbf{Jigsaw-T} & \textbf{Maze-T} & \textbf{Sliding-T} & \textbf{Geometry-T} & \textbf{Avg}$\uparrow$ \\
\midrule
\multicolumn{7}{c}{\textit{Closed-source Models}} \\
\midrule
Nano-Banana~\cite{google2025gemini25flashmodelcard} & - & 57.0 & 4.7 & 0.0 & 47.8 & 27.4 \\
\midrule
\multicolumn{7}{c}{\textit{Open-source Models}} \\
\midrule
\textbf{SenseNova-U1-SFT} & \textbf{8B} & 87.3 & 28.6 & 0.0 & 24.2 & \textbf{35.0} \\
\textbf{SenseNova-U1-SFT} & \textbf{8BA3B} & 88.0 & 34.0 & 1.2 & 7.1 & \textbf{32.6} \\
BAGEL~\cite{deng2025bagel} & 7B & 48.0 & 0.0 & 1.2 & 32.8 & 20.5 \\
Ovis-U1~\cite{wang2025ovis} & 1.2B & 53.0 & 0.0 & 0.0 & 3.5 & 14.1 \\
OmniGen2~\cite{wu2025omnigen2} & 4B & 48.0 & 0.0 & 0.0 & 5.7 & 13.4 \\
Qwen-Image-Edit~\cite{wu2025qwenimagetechnicalreport} & 20B & 43.3 & 0.7 & 0.0 & 8.5 & 13.1 \\
\bottomrule
\end{tabular}
\end{adjustbox}
\end{table}

\begin{table}[h!]
\centering
\caption{\textbf{Quantitative evaluation results on RealUnify}. We report the step-wise inference results; for SenseNova-U1, the GEU results are obtained via a single interleaved process, while the UEG results are obtained via text-to-image generation inference.}
\label{tab:realunify_results}
\renewcommand{\arraystretch}{1}
\setlength{\tabcolsep}{3pt}
\begin{adjustbox}{width=\textwidth}
\begin{tabular}{lc|ccccccc|ccccc|c}
\toprule
\multirow{2}{*}{\textbf{Model}} & \multirow{2}{*}{\textbf{\# Params}} 
& \multicolumn{7}{c|}{\textbf{Understanding Enhances Generation}}
& \multicolumn{5}{c|}{\textbf{Generation Enhances Understanding}}
& \multirow{2}{*}{\textbf{Avg}$\uparrow$} \\
\cmidrule(lr){3-9} \cmidrule(lr){10-14}
& & WK & CR & MR-I & LR & SR & C2I & Avg-UEG & MR-II & MT & AF & CN & Avg-GEU & \\
\midrule
\textbf{SenseNova-U1-SFT} & \textbf{8B} & 88 & 68 & 33 & 45 & 54 & 46 & 55.7 & 36 & 63 & 51 & 40 & 47.5 & \textbf{52.4} \\
\textbf{SenseNova-U1-SFT} & \textbf{8BA3B} & 81 & 56 & 41 & 40 & 55 & 44 & 52.8 & 30 & 65 & 58 & 35 & 47.0 & \textbf{50.5} \\
BAGEL~\cite{deng2025bagel} & 7B & 74 & 80 & 26 & 37 & 29 & 40 & 47.7 & 38 & 25 & 52 & 28 & 35.8 & 42.9 \\
Ovis-U1~\cite{wang2025ovis} & 1.2B & 59 & 71 & 30 & 34 & 17 & 25 & 39.3 & 38 & 25 & 31 & 24 & 29.5 & 35.4 \\
OneCAT~\cite{li2025onecat} & 9BA3B & 64 & 65 & 20 & 27 & 31 & 27 & 39.0 & 29 & 26 & 26 & 36 & 29.2 & 35.1 \\
UniPic2~\cite{wei2025skywork} & 2B & 62 & 72 & 30 & 38 & 26 & 15 & 40.5 & 28 & 24 & 27 & 16 & 23.8 & 33.8 \\
UniWorld-V1~\cite{lin2025uniworld} & 12B & 56 & 59 & 26 & 37 & 24 & 9 & 35.2 & 33 & 25 & 36 & 20 & 28.5 & 32.5 \\
OmniGen2~\cite{wu2025omnigen2} & 4B & 55 & 60 & 26 & 28 & 20 & 6 & 32.5 & 42 & 24 & 38 & 19 & 30.8 & 31.8 \\
ILLUME+~\cite{huang2025illume+} & 3B & 52 & 62 & 22 & 25 & 26 & 7 & 32.3 & 27 & 20 & 38 & 25 & 27.5 & 30.4 \\
\bottomrule
\end{tabular}
\end{adjustbox}
\end{table}

\subsection{Ablation Studies}
\label{sec:ablation-studies}

\begin{table}[t]
\centering
\caption{Reconstruction performance with a frozen understanding branch on MS-COCO 2017.}
\renewcommand{\arraystretch}{1}
\setlength{\tabcolsep}{3pt}
\resizebox{0.6\textwidth}{!}{
\begin{tabular}{lcccc}
\toprule
\textbf{Method} 
& \textbf{Downsampling Ratio}
& \textbf{Resolution} 
& \textbf{PSNR$\uparrow$} & \textbf{SSIM$\uparrow$} \\
\midrule
SDXL VAE~\cite{podell2023sdxl} & 8 & 256
& 25.76 & 0.76 \\
SD3 VAE~\cite{lopez2025sd3} & 8 & 256
& 29.47 & 0.86 \\
FLUX.1-dev VAE~\cite{labs2025flux1kontextflowmatching} & 8 & 256
& 30.43 & 0.93 \\
RAE (DINOv2-B)~\cite{zheng2025diffusion}  & 14 & 256 
& 18.36 & 0.47 \\ 
UniFlow (DINOv2-L)~\cite{yue2025uniflow}  & 14 & 256 & 30.66 & 0.94 \\ 
UAE (DINOv2-L)~\cite{fan2025prism}  & 14 & 256 & 32.74 & 0.94 \\ 
\midrule
FLUX.1-dev VAE~\cite{labs2025flux1kontextflowmatching} & 8 & 512
& 31.56 & 0.93 \\
Neo-unify (2B) & 32 & 512 & 31.56 & 0.85 \\
\rowcolor{lightgray}
\bottomrule
\end{tabular}
}
\label{tab:token_comp}
\end{table}

We conduct a series of ablation studies about SenseNova-U1, focusing on three key questions: whether the encoder-free design preserves both semantic and pixel-level representations, whether it synergizes effectively with the MoT backbone while minimizing intrinsic modality conflict, and whether it exhibits strong data-scaling efficiency.

\begin{figure}[!h]
    \centering
    \includegraphics[width=0.8\linewidth]{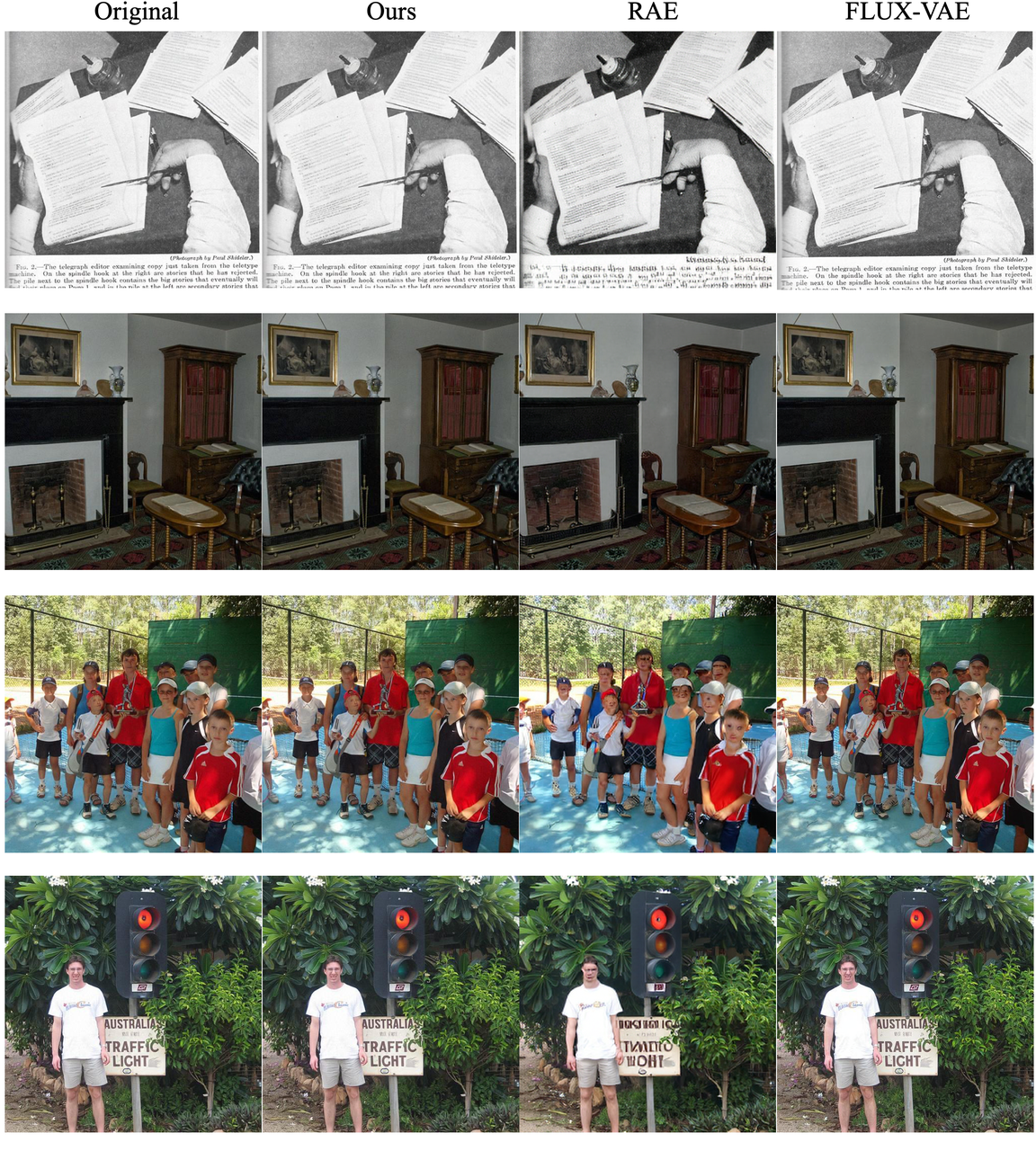}
    \caption{Reconstructing out-domain images with 2B NEO-unify under a frozen understanding branch.}
    \label{fig:ablation-recon}
\end{figure}

\begin{figure}[!h]
    \centering
    \includegraphics[width=0.8\linewidth]{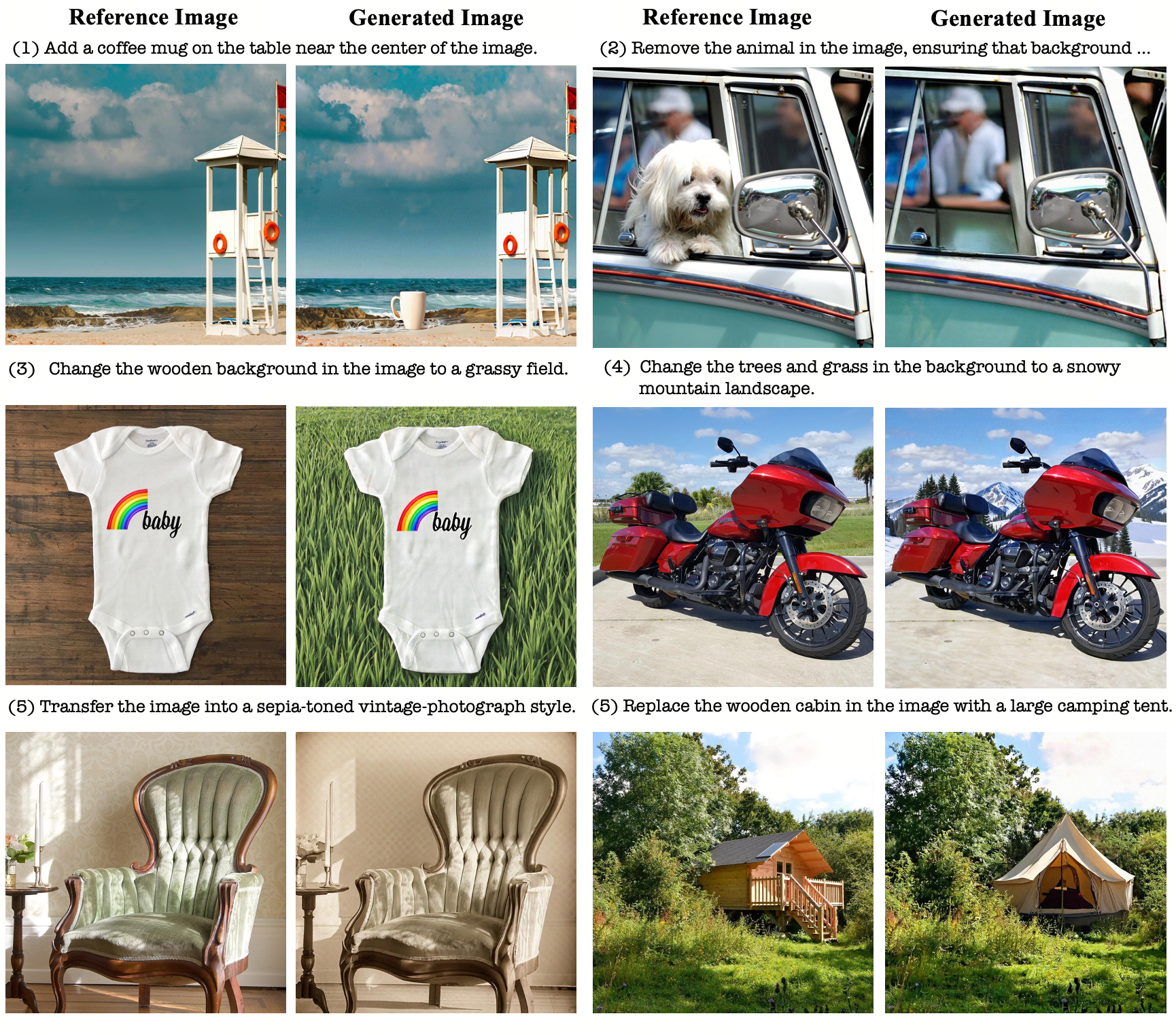}
    \caption{Validating ImgEdit prompts with 2B NEO-unify under a frozen understanding branch.}
    \label{fig:ablation-edit}
\end{figure}

\begin{figure}[!h]
    \centering
    \includegraphics[width=0.8\linewidth]{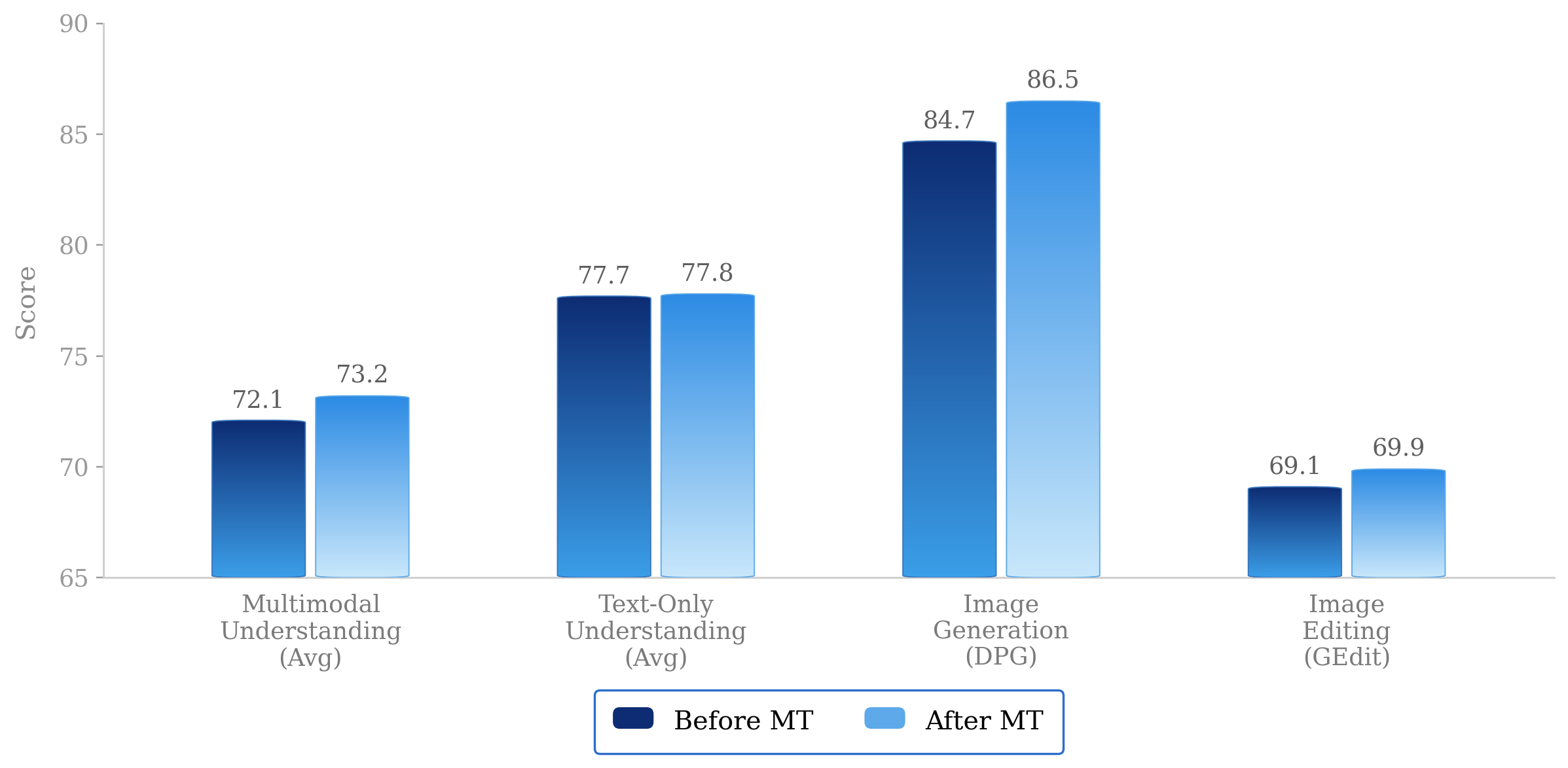}
    \caption{Understanding–generation co-training with 8B-MoT backbone. GEdit-Bench scores are normalized to 0–100 scale.}
    \label{fig:ablation-conflict}
\end{figure}

\subsubsection{Native Encoder-Free Design Preserves Both Semantic and Pixel Representations}

\textbf{Image Reconstruction.}
As reported in Table~\ref{tab:token_comp}, our previously released NEO-unify (2B)~\cite{sensenova2026neounify} attains 31.56 PSNR and 0.85 SSIM on MS-COCO 2017~\cite{Datasets:MSCOCO} after only 90K pretraining steps, approaching the 31.56 PSNR and 0.93 SSIM achieved by the FLUX.1-dev VAE. This result suggests that the native near-lossless interface is capable of retaining both high-level semantic information and fine-grained visual details without depending on pretrained vision encoders or latent autoencoders. Representative reconstruction examples are presented in Figure~\ref{fig:ablation-recon}.

\textbf{Image Editing.} 
For editing tasks, NEO-unify (2B)~\cite{sensenova2026neounify} routes all conditional contexts through the understanding branch, while the generation branch directly synthesizes the target images. Despite freezing the understanding branch throughout training, the model still exhibits strong editing capability, together with substantially improved token efficiency. Using only public text-to-image and editing datasets, it achieves an ImgEdit score of 3.32 after an initial 60K-step mixed training process. Representative editing examples on ImgEdit prompts are shown in Figure~\ref{fig:ablation-edit}.

\subsubsection{Understanding and Generation Synergize with Native MoT Backbone}
Starting from pretrained dual branches, we jointly optimize all components during mid-training. Even with low data ratios and small understanding loss weights, understanding remains stable while generation converges rapidly. As shown in Figure~\ref{fig:ablation-conflict}, the two capabilities co-evolve effectively within the MoT backbone with minimal intrinsic conflict.

\begin{figure*}[t]
    \centering

    \begin{subfigure}[t]{0.46\textwidth}
        \centering
        \includegraphics[width=\linewidth]{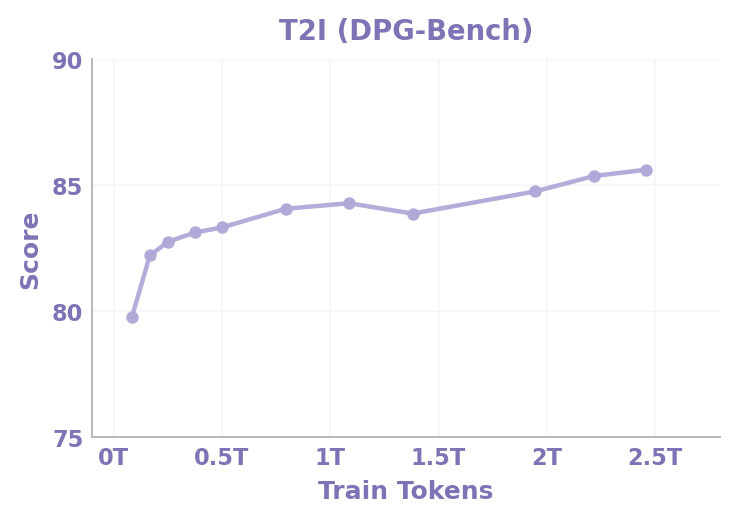}
        \caption{Scores on the image generation task (DPG-Bench).}
    \end{subfigure}
    \hfill
    \begin{subfigure}[t]{0.46\textwidth}
        \centering
        \includegraphics[width=\linewidth]{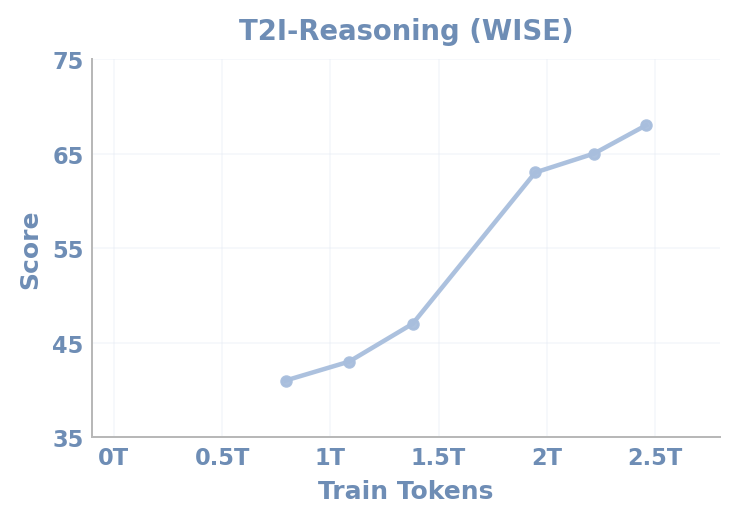}
        \caption{Scores on the reasoning image generation task (WISE).}
    \end{subfigure}

    \vspace{0.6em}

    \begin{subfigure}[t]{0.46\textwidth}
        \centering
        \includegraphics[width=\linewidth]{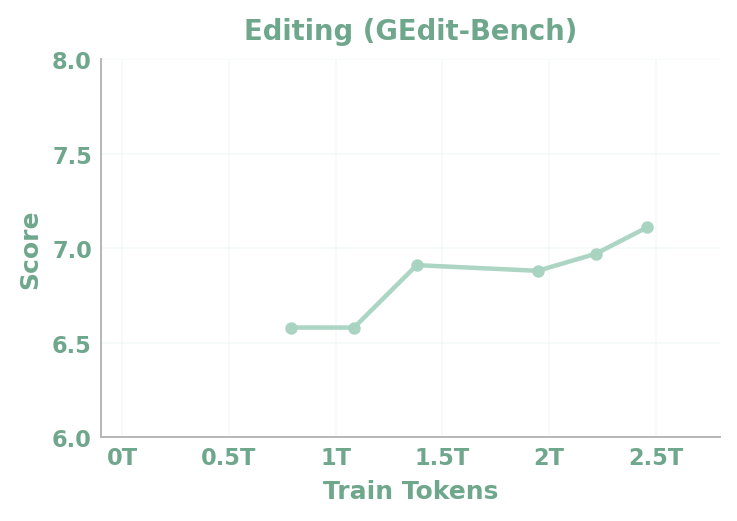}
        \caption{Scores on the image editing task (GEdit-Bench).}
    \end{subfigure}
    \hfill
    \begin{subfigure}[t]{0.46\textwidth}
        \centering
        \includegraphics[width=\linewidth]{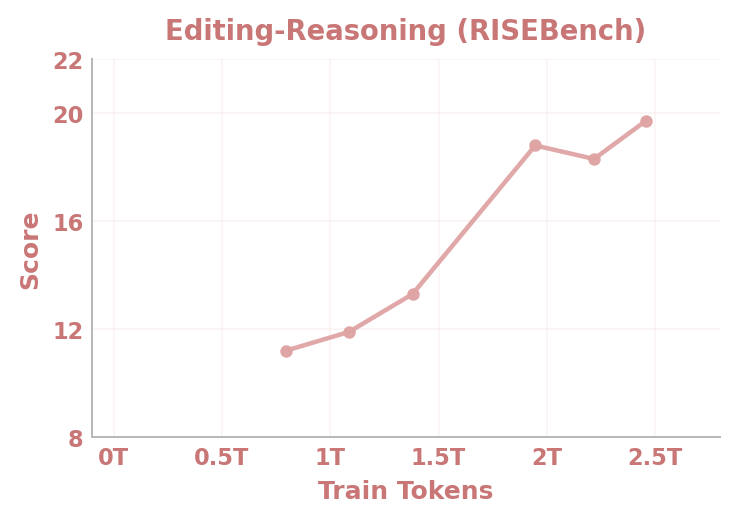}
        \caption{Scores on the reasoning image editing task (RISEBench).}
    \end{subfigure}
    \caption{
    Data-scaling curves of the 8B-MoT backbone. We set the resolution as 512 and 1,024 for DPG-Bench and others.
    }
    \label{fig:ablation-efficiency}
\end{figure*}

\subsubsection{Native Multimodal Architecture Shows High Data-Scaling Efficiency}
We begin with web-scale pretraining, followed by mid-training and supervised fine-tuning using diverse, high-quality data corpora spanning both understanding and generation tasks. As shown in Figure~\ref{fig:ablation-efficiency}, the model delivers strong data-scaling efficiency, with both generation quality and understanding–generation synergy improving steadily as training data scale.
Overall, these results further validate the advantages of the native MoT design from three complementary perspectives: preserving both semantic structure and pixel-level fidelity, reducing intrinsic conflict between understanding and generation, and enabling efficient scaling across data scale and multimodal generation tasks.

\subsection{Visualization Results}
\label{sec:visualization-results}
In addition to quantitative evaluations, we provide qualitative visualizations to illustrate the behaviors of SenseNova-U1 in complex multimodal scenarios. We focus on representative cases spanning text-to-image generation, infographic generation, image editing, interleaved image-text generation, visual understanding, and agentic multimodal interaction, covering both general and reasoning-intensive settings. Additional showcases are available at \href{https://github.com/OpenSenseNova/SenseNova-U1/blob/main/docs/showcases.md}{https://github.com/OpenSenseNova/SenseNova-U1/blob/main/docs/showcases.md}.

\begin{figure*}[t]
    \centering
    \includegraphics[width=\linewidth]{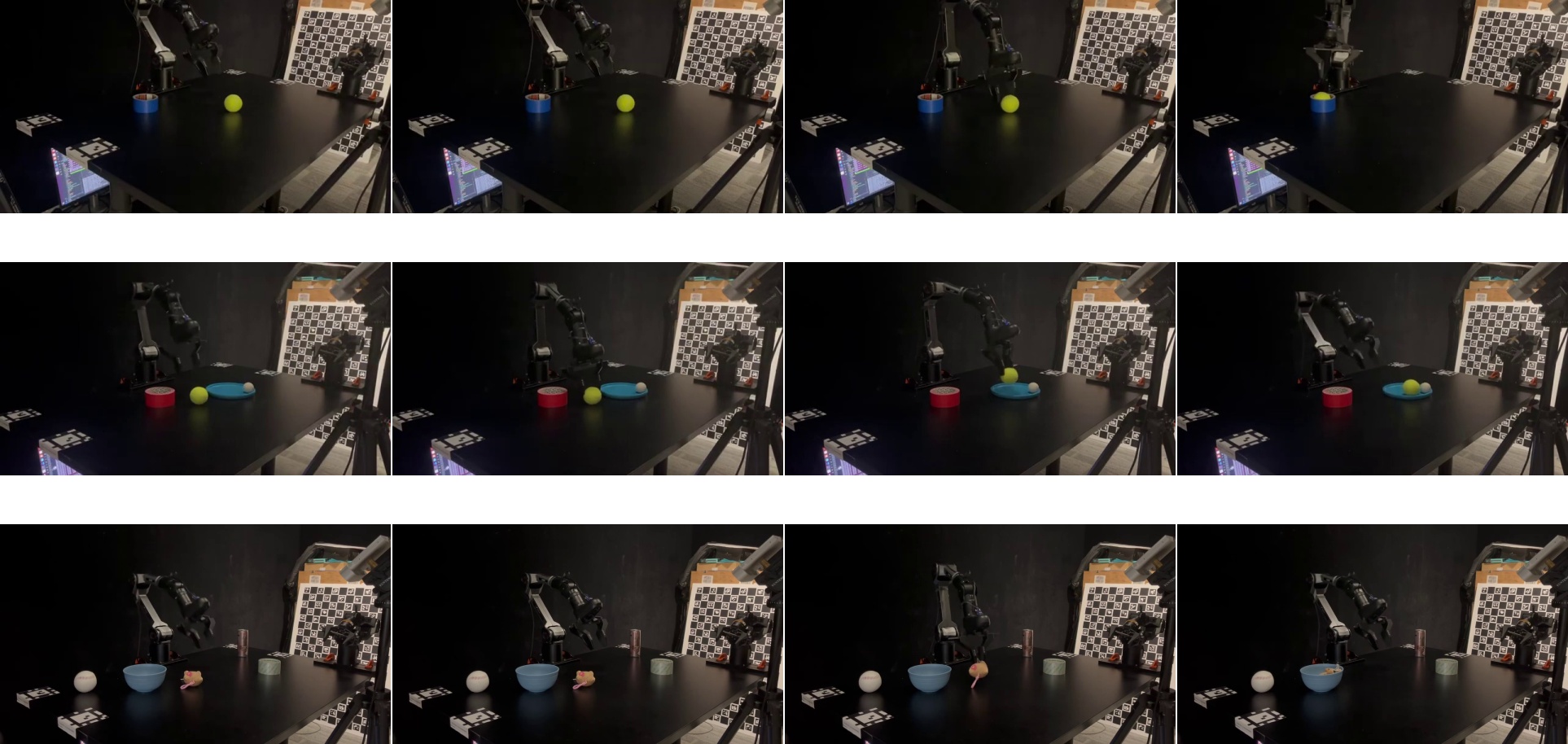}
    \caption{Visualizing vision-language-action behaviors of SenseNova-U1 on robotic manipulation videos.}
    \label{fig:visualization-vla}
\end{figure*}

\begin{figure*}[!h]
    \centering
    \includegraphics[width=\linewidth]{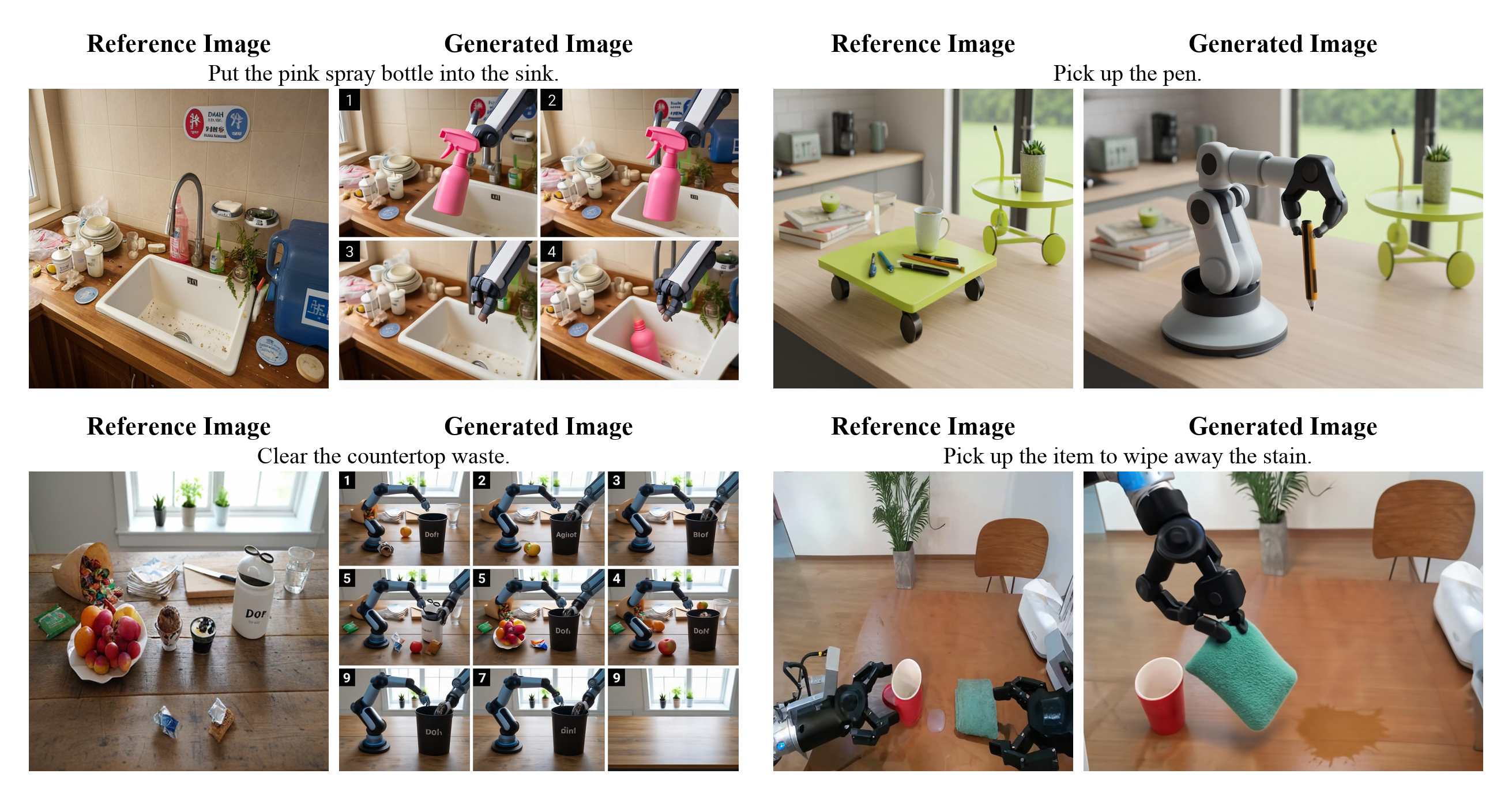}
    \caption{Visualizing world-modeling predictions from the robotic arm view with SenseNova-U1.}
    \label{fig:visualization-world-model}
\end{figure*}

\textbf{Vision-Language-Action.}
We first present representative video-based action reasoning examples in Figure~\ref{fig:visualization-vla}. For each example, four frames are uniformly sampled from the input video and arranged in a single row to illustrate the temporal progression of the manipulation process. These cases show that SenseNova-U1 can capture action-relevant visual dynamics across time, maintain coherent visual understanding under embodied settings, and reason about object states, manipulation trajectories, and task progression from sparse temporal observations.

\textbf{World Modeling.}
We further visualize the world modeling capability of SenseNova-U1 in Figure~\ref{fig:visualization-world-model}. Given an input image and an action-oriented instruction, the model is required to predict the corresponding visual outcome. For readability, simplified instructions are shown in the figure, while the original prompts are used during inference. The selected examples demonstrate that SenseNova-U1 can translate structured action instructions into plausible visual state transitions while preserving overall scene consistency and object coherence.


\section{Conclusion}

We present a unified multimodal foundation model in which understanding, generation, and reasoning emerge within a single native architecture rather than through the coordination of separate systems. Across a broad range of tasks, the model exhibits strong capabilities in vision-language perception, semantic reasoning, high-fidelity generation, and interleaved multimodal interaction, suggesting that a shared representation can simultaneously support analytical and creative intelligence.
More fundamentally, our results point toward a broader transition in multimodal AI. Rather than merely aligning isolated modalities, unified models begin to internalize a coherent abstraction of the world itself, enabling perception, imagination, and decision-making to arise within a shared latent space. Early advances in vision-language-action models and world modeling further indicate a path from passive understanding toward embodied, goal-directed intelligence.
We believe the next generation of AI will emerge not from increasingly complex collections of modular components, but from unified architectures grounded in a single underlying intelligence.

\section{Contributors}

The list is organized by contribution role, with individuals listed alphabetically by their first name within each category.\\

\textbf{Project Sponsor and Advisor}:\\
Dahua Lin

\textbf{Senior Project Lead}: \\
Lei Yang, Lewei Lu, Quan Wang, Ruihao Gong, Wenxiu Sun, Ziwei Liu

\textbf{Project Lead}: \\
Haiwen Diao

\textbf{Core Contributor}: \\
Hanming Deng, Jiahao Wang, Penghao Wu, Shihao Bai, Silei Wu, Weichen Fan, Wenjie Ye, Wenwen Tong, Xiangyu Fan, Yan Li, Yubo Wang, Zhijie Cao, Zhiqian Lin, Zhitao Yang, Zhongang Cai

\textbf{Contributor}: \\
Bo Liu, Chengguang Lv, Haojia Yu, Haozhe Xie, Hongli Wang, Jianan Fan, Jiaqi Li, Jiefan Lu, Jingcheng Ni, Junxiang Xu, Kaihuan Liang, Lianqiang Shi, Linjun Dai, Linyan Wang, Oscar Qian, Peng Gao, Pengfei Liu, Qingping Sun, Rui Shen, Ruisi Wang, Shengnan Ma, Shuang Yang, Siyi Xie, Siying Li, Tianbo Zhong, Xiangli Kong, Xuanke Shi, Yang Gao, Yongqiang Yao, Yue Zhu, Yuwei Niu, Yves Wang, Zhengqi Bai, Zhengyu Lin, Zixin Yin

\textbf{Acknowledgement}: \\
We would like to thank Bo Yang, Boxuan Li, Chen Feng, Chen Wei, Chenyang Gu, Fanyi Pu, Fanzhou Wang, Guanzhou Chen, Haoge Deng, Hongyu Liang, Houyuan Chen, Huaping Zhong, Huchuan Lu, Jiawei Hong, Jinkun Xie, Jinwei Liang, Mingxuan Li, Mutian Xu, Ruize Ma, Siqi Luo, Tiankuo Yao, Tongxi Zhou, Wangqi Yin, Xiaotong Li, Yinfei Zeng, Yong Xien Chng, Yuhao Dong, Yukang Cao, Zheng Ma, Ziming Wu, Zongpu Zhang, and Zukai Chen for their valuable support and contributions to this project, including data preparation, model evaluation, infrastructure support, architecture analysis, and helpful discussions.

\clearpage

\bibliographystyle{plainnat}
\bibliography{main}

@String(CVPR= {IEEE Conf. Comput. Vis. Pattern Recog.})

@String(CVPR  = {CVPR})

@article{hurst2024gpt,
  title={Gpt-4o system card},
  author={Hurst, Aaron and Lerer, Adam and Goucher, Adam P and Perelman, Adam and Ramesh, Aditya and Clark, Aidan and Ostrow, AJ and Welihinda, Akila and Hayes, Alan and Radford, Alec and others},
  journal={arXiv preprint arXiv:2410.21276},
  year={2024}
}

@article{openai_gpt5_systemcard,
  title={Openai gpt-5 system card},
  author={Singh, Aaditya and Fry, Adam and Perelman, Adam and Tart, Adam and Ganesh, Adi and El-Kishky, Ahmed and McLaughlin, Aidan and Low, Aiden and Ostrow, AJ and Ananthram, Akhila and others},
  journal={arXiv preprint arXiv:2601.03267},
  year={2025}
}

@article{team2023gemini,
  title={Gemini: a family of highly capable multimodal models},
  author={Team, Gemini and Anil, Rohan and Borgeaud, Sebastian and Alayrac, Jean-Baptiste and Yu, Jiahui and Soricut, Radu and Schalkwyk, Johan and Dai, Andrew M and Hauth, Anja and Millican, Katie and others},
  journal={arXiv preprint arXiv:2312.11805},
  year={2023}
}

@misc{gemini_3_pro_systemcard,
  author       = {{Google DeepMind}},
  title        = {{Gemini 3 Pro Model Card}},
  url = {https://storage.googleapis.com/deepmind-media/Model-Cards/Gemini-3-Pro-Model-Card.pdf},
  year         = {2025},
  month        = nov,
  day          = 18,
}

@article{PixelFlow,
  title={Pixelflow: Pixel-space generative models with flow},
  author={Chen, Shoufa and Ge, Chongjian and Zhang, Shilong and Sun, Peize and Luo, Ping},
  journal={arXiv preprint arXiv:2504.07963},
  year={2025}
}

@article{Qwen3-VL,
  title={Qwen3-vl technical report},
  author={Bai, Shuai and Cai, Yuxuan and Chen, Ruizhe and Chen, Keqin and Chen, Xionghui and Cheng, Zesen and Deng, Lianghao and Ding, Wei and Gao, Chang and Ge, Chunjiang and others},
  journal={arXiv preprint arXiv:2511.21631},
  year={2025}
}

@misc{qwen35blog,
  author = {{Qwen Team}},
  title = {Qwen3.5: Towards Native Multimodal Agents},
  year = {2026},
  month = {February},
  url = {https://qwen.ai/blog?id=qwen3.5},
}

@article{liu2023llava,
  title={Visual instruction tuning},
  author={Liu, Haotian and Li, Chunyuan and Wu, Qingyang and Lee, Yong Jae},
  journal={Advances in Neural Information Processing Systems},
  volume={36},
  pages={34892--34916},
  year={2023}
}

@article{wang2025internvl3,
  title={Internvl3. 5: Advancing open-source multimodal models in versatility, reasoning, and efficiency},
  author={Wang, Weiyun and Gao, Zhangwei and Gu, Lixin and Pu, Hengjun and Cui, Long and Wei, Xingguang and Liu, Zhaoyang and Jing, Linglin and Ye, Shenglong and Shao, Jie and others},
  journal={arXiv preprint arXiv:2508.18265},
  year={2025}
}

@article{deng2025bagel,
  title   = {Emerging Properties in Unified Multimodal Pretraining},
  author  = {Deng, Chaorui and Zhu, Deyao and Li, Kunchang and Gou, Chenhui and Li, Feng and Wang, Zeyu and Zhong, Shu and Yu, Weihao and Nie, Xiaonan and Song, Ziang and Shi, Guang and Fan, Haoqi},
  journal = {arXiv preprint arXiv:2505.14683},
  year    = {2025}
}

@inproceedings{vsi,
  title={Thinking in space: How multimodal large language models see, remember, and recall spaces},
  author={Yang, Jihan and Yang, Shusheng and Gupta, Anjali W and Han, Rilyn and Fei-Fei, Li and Xie, Saining},
  booktitle={Proceedings of the IEEE/CVF Conference on Computer Vision and Pattern Recognition,},
  pages={10632--10643},
  year={2025}
}

@inproceedings{mindcube,
  title={Spatial mental modeling from limited views},
  author={Yin, Baiqiao and Wang, Qineng and Zhang, Pingyue and Zhang, Jianshu and Wang, Kangrui and Wang, Zihan and Zhang, Jieyu and Chandrasegaran, Keshigeyan and Liu, Han and Krishna, Ranjay and others},
  booktitle={Proceedings of the IEEE/CVF International Conference on Computer Vision Workshop},
  year={2025}
}

@inproceedings{liu2024mmbench,
  title={Mmbench: Is your multi-modal model an all-around player?},
  author={Liu, Yuan and Duan, Haodong and Zhang, Yuanhan and Li, Bo and Zhang, Songyang and Zhao, Wangbo and Yuan, Yike and Wang, Jiaqi and He, Conghui and Liu, Ziwei and others},
  booktitle={Proceedings of the European Conference on Computer Vision},
  pages={216--233},
  year={2024},
  organization={Springer}
}

@article{li2025viewspatial,
  title={ViewSpatial-Bench: Evaluating Multi-perspective Spatial Localization in Vision-Language Models},
  author={Li, Dingming and Li, Hongxing and Wang, Zixuan and Yan, Yuchen and Zhang, Hang and Chen, Siqi and Hou, Guiyang and Jiang, Shengpei and Zhang, Wenqi and Shen, Yongliang and others},
  journal={arXiv preprint arXiv:2505.21500},
  year={2025}
}

@article{ma20243dsrbench,
  title={3dsrbench: A comprehensive 3d spatial reasoning benchmark},
  author={Ma, Wufei and Chen, Haoyu and Zhang, Guofeng and Chou, Yu-Cheng and de Melo, Celso M and Yuille, Alan},
  journal={arXiv preprint arXiv:2412.07825},
  year={2024}
}

@article{ghosh2023geneval,
  title={Geneval: An object-focused framework for evaluating text-to-image alignment},
  author={Ghosh, Dhruba and Hajishirzi, Hannaneh and Schmidt, Ludwig},
  journal={Advances in Neural Information Processing Systems},
  volume={36},
  pages={52132--52152},
  year={2023}
}

@article{hu2024ella,
  title={Ella: Equip diffusion models with llm for enhanced semantic alignment},
  author={Hu, Xiwei and Wang, Rui and Fang, Yixiao and Fu, Bin and Cheng, Pei and Yu, Gang},
  journal={arXiv preprint arXiv:2403.05135},
  year={2024}
}

@article{chang2025oneig,
  title={OneIG-Bench: Omni-dimensional Nuanced Evaluation for Image Generation},
  author={Chang, Jingjing and Fang, Yixiao and Xing, Peng and Wu, Shuhan and Cheng, Wei and Wang, Rui and Zeng, Xianfang and Yu, Gang and Chen, Hai-Bao},
  journal={arXiv preprint arXiv:2506.07977},
  year={2025}
}

@article{wei2025tiif,
  title={TIIF-Bench: How Does Your T2I Model Follow Your Instructions?},
  author={Wei, Xinyu and Zhang, Jinrui and Wang, Zeqing and Wei, Hongyang and Guo, Zhen and Zhang, Lei},
  journal={arXiv preprint arXiv:2506.02161},
  year={2025}
}

@article{geng2025x,
  title={X-omni: Reinforcement learning makes discrete autoregressive image generative models great again},
  author={Geng, Zigang and Wang, Yibing and Ma, Yeyao and Li, Chen and Rao, Yongming and Gu, Shuyang and Zhong, Zhao and Lu, Qinglin and Hu, Han and Zhang, Xiaosong and others},
  journal={arXiv preprint arXiv:2507.22058},
  year={2025}
}

@article{du2025textcrafter,
  title={Textcrafter: Accurately rendering multiple texts in complex visual scenes},
  author={Du, Nikai and Chen, Zhennan and Gao, Shan and Chen, Zhizhou and Chen, Xi and Jiang, Zhengkai and Yang, Jian and Tai, Ying},
  journal={arXiv preprint arXiv:2503.23461},
  year={2025}
}

@article{niu2025wise,
  title={Wise: A world knowledge-informed semantic evaluation for text-to-image generation},
  author={Niu, Yuwei and Ning, Munan and Zheng, Mengren and Jin, Weiyang and Lin, Bin and Jin, Peng and Liao, Jiaqi and Feng, Chaoran and Ning, Kunpeng and Zhu, Bin and others},
  journal={arXiv preprint arXiv:2503.07265},
  year={2025}
}

@article{tang2026igenbench,
  title={IGenBench: Benchmarking the Reliability of Text-to-Infographic Generation},
  author={Tang, Yinghao and Liu, Xueding and Zhang, Boyuan and Lan, Tingfeng and Xie, Yupeng and Lao, Jiale and Wang, Yiyao and Li, Haoxuan and Gao, Tingting and Pan, Bo and others},
  journal={arXiv preprint arXiv:2601.04498},
  year={2026}
}

@article{li2026bizgeneval,
  title={BizGenEval: A Systematic Benchmark for Commercial Visual Content Generation},
  author={Li, Yan and Zeng, Zezi and Zhou, Ziwei and Gao, Xin and Tian, Muzhao and Yang, Yifan and Cheng, Mingxi and Dai, Qi and Yang, Yuqing and Qiu, Lili and others},
  journal={arXiv preprint arXiv:2603.25732},
  year={2026}
}

@inproceedings{zhou2025opening,
  title={OpenING: A Comprehensive Benchmark for Judging Open-ended Interleaved Image-Text Generation},
  author={Zhou, Pengfei and Peng, Xiaopeng and Song, Jiajun and Li, Chuanhao and Xu, Zhaopan and Yang, Yue and Guo, Ziyao and Zhang, Hao and Lin, Yuqi and He, Yefei and others},
  booktitle={Proceedings of the Computer Vision and Pattern Recognition Conference},
  pages={56--66},
  year={2025}
}

@article{shi2025realunify,
  title={RealUnify: Do Unified Models Truly Benefit from Unification? A Comprehensive Benchmark},
  author={Shi, Yang and Dong, Yuhao and Ding, Yue and Wang, Yuran and Zhu, Xuanyu and Zhou, Sheng and Liu, Wenting and Tian, Haochen and Wang, Rundong and Wang, Huanqian and others},
  journal={arXiv preprint arXiv:2509.24897},
  year={2025}
}

@article{zou2025unimmmu,
  title={Uni-MMMU: A Massive Multi-discipline Multimodal Unified Benchmark},
  author={Zou, Kai and Huang, Ziqi and Dong, Yuhao and Tian, Shulin and Zheng, Dian and Liu, Hongbo and He, Jingwen and Liu, Bin and Qiao, Yu and Liu, Ziwei},
  journal={arXiv preprint arXiv:2510.13759},
  year={2025}
}

@inproceedings{yue2024mmmu,
  title={Mmmu: A massive multi-discipline multimodal understanding and reasoning benchmark for expert agi},
  author={Yue, Xiang and Ni, Yuansheng and Zhang, Kai and Zheng, Tianyu and Liu, Ruoqi and Zhang, Ge and Stevens, Samuel and Jiang, Dongfu and Ren, Weiming and Sun, Yuxuan and others},
  booktitle={Proceedings of the IEEE/CVF Conference on Computer Vision and Pattern Recognition},
  pages={9556--9567},
  year={2024}
}

@article{lu2023mathvista,
  title={Mathvista: Evaluating mathematical reasoning of foundation models in visual contexts},
  author={Lu, Pan and Bansal, Hritik and Xia, Tony and Liu, Jiacheng and Li, Chunyuan and Hajishirzi, Hannaneh and Cheng, Hao and Chang, Kai-Wei and Galley, Michel and Gao, Jianfeng},
  journal={arXiv preprint arXiv:2310.02255},
  year={2023}
}

@inproceedings{guan2024hallusionbench,
  title={Hallusionbench: an advanced diagnostic suite for entangled language hallucination and visual illusion in large vision-language models},
  author={Guan, Tianrui and Liu, Fuxiao and Wu, Xiyang and Xian, Ruiqi and Li, Zongxia and Liu, Xiaoyu and Wang, Xijun and Chen, Lichang and Huang, Furong and Yacoob, Yaser and others},
  booktitle={Proceedings of the IEEE/CVF Conference on Computer Vision and Pattern Recognition},
  pages={14375--14385},
  year={2024}
}

@article{fu2024ocrbench,
  title={Ocrbench v2: An improved benchmark for evaluating large multimodal models on visual text localization and reasoning},
  author={Fu, Ling and Kuang, Zhebin and Song, Jiajun and Huang, Mingxin and Yang, Biao and Li, Yuzhe and Zhu, Linghao and Luo, Qidi and Wang, Xinyu and Lu, Hao and others},
  journal={arXiv preprint arXiv:2501.00321},
  year={2024}
}

@article{zheng2025diffusion,
  title={Diffusion transformers with representation autoencoders},
  author={Zheng, Boyang and Ma, Nanye and Tong, Shengbang and Xie, Saining},
  journal={arXiv preprint arXiv:2510.11690},
  year={2025}
}

@article{fan2025prism,
  title={The Prism Hypothesis: Harmonizing Semantic and Pixel Representations via Unified Autoencoding},
  author={Fan, Weichen and Diao, Haiwen and Wang, Quan and Lin, Dahua and Liu, Ziwei},
  journal={arXiv preprint arXiv:2512.19693},
  year={2025}
}

@article{yang2025qwen3,
  title={Qwen3 technical report},
  author={Yang, An and Li, Anfeng and Yang, Baosong and Zhang, Beichen and Hui, Binyuan and Zheng, Bo and Yu, Bowen and Gao, Chang and Huang, Chengen and Lv, Chenxu and others},
  journal={arXiv preprint arXiv:2505.09388},
  year={2025}
}

@inproceedings{qu2025tokenflow,
  title={Tokenflow: Unified image tokenizer for multimodal understanding and generation},
  author={Qu, Liao and Zhang, Huichao and Liu, Yiheng and Wang, Xu and Jiang, Yi and Gao, Yiming and Ye, Hu and Du, Daniel K and Yuan, Zehuan and Wu, Xinglong},
  booktitle={Proceedings of the Computer Vision and Pattern Recognition Conference},
  pages={2545--2555},
  year={2025}
}

@article{ma2025unitok,
  title={Unitok: A unified tokenizer for visual generation and understanding},
  author={Ma, Chuofan and Jiang, Yi and Wu, Junfeng and Yang, Jihan and Yu, Xin and Yuan, Zehuan and Peng, Bingyue and Qi, Xiaojuan},
  journal={arXiv preprint arXiv:2502.20321},
  year={2025}
}

@article{Dualtoken,
  title={Dualtoken: Towards unifying visual understanding and generation with dual visual vocabularies},
  author={Song, Wei and Wang, Yuran and Song, Zijia and Li, Yadong and Sun, Haoze and Chen, Weipeng and Zhou, Zenan and Xu, Jianhua and Wang, Jiaqi and Yu, Kaicheng},
  journal={arXiv preprint arXiv:2503.14324},
  year={2025}
}

@article{yue2025uniflow,
  title={UniFlow: A Unified Pixel Flow Tokenizer for Visual Understanding and Generation},
  author={Yue, Zhengrong and Zhang, Haiyu and Zeng, Xiangyu and Chen, Boyu and Wang, Chenting and Zhuang, Shaobin and Dong, Lu and Du, KunPeng and Wang, Yi and Wang, Limin and others},
  journal={arXiv preprint arXiv:2510.10575},
  year={2025}
}

@article{MOMA,
  title={Moma: Efficient early-fusion pre-training with mixture of modality-aware experts},
  author={Lin, Xi Victoria and Shrivastava, Akshat and Luo, Liang and Iyer, Srinivasan and Lewis, Mike and Ghosh, Gargi and Zettlemoyer, Luke and Aghajanyan, Armen},
  journal={arXiv preprint arXiv:2407.21770},
  year={2024}
}

@article{MoT,
  title={Mixture-of-transformers: A sparse and scalable architecture for multi-modal foundation models},
  author={Liang, Weixin and Yu, Lili and Luo, Liang and Iyer, Srinivasan and Dong, Ning and Zhou, Chunting and Ghosh, Gargi and Lewis, Mike and Yih, Wen-tau and Zettlemoyer, Luke and others},
  journal={arXiv preprint arXiv:2411.04996},
  year={2024}
}

@article{zhou2024transfusion,
  title={Transfusion: Predict the next token and diffuse images with one multi-modal model},
  author={Zhou, Chunting and Yu, Lili and Babu, Arun and Tirumala, Kushal and Yasunaga, Michihiro and Shamis, Leonid and Kahn, Jacob and Ma, Xuezhe and Zettlemoyer, Luke and Levy, Omer},
  journal={arXiv preprint arXiv:2408.11039},
  year={2024}
}

@article{gu2025thinkmorph,
  title={Thinkmorph: Emergent properties in multimodal interleaved chain-of-thought reasoning},
  author={Gu, Jiawei and Hao, Yunzhuo and Wang, Huichen Will and Li, Linjie and Shieh, Michael Qizhe and Choi, Yejin and Krishna, Ranjay and Cheng, Yu},
  journal={arXiv preprint arXiv:2510.27492},
  year={2025}
}

@article{Chameleon,
  author = {Chameleon Team},
  doi = {10.48550/arXiv.2405.09818},
  journal = {arXiv preprint arXiv:2405.09818},
  title = {Chameleon: Mixed-Modal Early-Fusion Foundation Models},
  url = {https://github.com/facebookresearch/chameleon},
  year = {2024}
}

@article{wang2024emu3,
  title={Emu3: Next-token prediction is all you need},
  author={Wang, Xinlong and Zhang, Xiaosong and Luo, Zhengxiong and Sun, Quan and Cui, Yufeng and Wang, Jinsheng and Zhang, Fan and Wang, Yueze and Li, Zhen and Yu, Qiying and others},
  journal={arXiv preprint arXiv:2409.18869},
  year={2024}
}

@article{xie2024show,
  title={Show-o: One single transformer to unify multimodal understanding and generation},
  author={Xie, Jinheng and Mao, Weijia and Bai, Zechen and Zhang, David Junhao and Wang, Weihao and Lin, Kevin Qinghong and Gu, Yuchao and Chen, Zhijie and Yang, Zhenheng and Shou, Mike Zheng},
  journal={arXiv preprint arXiv:2408.12528},
  year={2024}
}

@article{li2025back,
  title={Back to basics: Let denoising generative models denoise},
  author={Li, Tianhong and He, Kaiming},
  journal={arXiv preprint arXiv:2511.13720},
  year={2025}
}

@article{yu2025pixeldit,
  title={Pixeldit: Pixel diffusion transformers for image generation},
  author={Yu, Yongsheng and Xiong, Wei and Nie, Weili and Sheng, Yichen and Liu, Shiqiu and Luo, Jiebo},
  journal={arXiv preprint arXiv:2511.20645},
  year={2025}
}

@article{Dip,
  title={Dip: Taming diffusion models in pixel space},
  author={Chen, Zhennan and Zhu, Junwei and Chen, Xu and Zhang, Jiangning and Hu, Xiaobin and Zhao, Hanzhen and Wang, Chengjie and Yang, Jian and Tai, Ying},
  journal={arXiv preprint arXiv:2511.18822},
  year={2025}
}

@article{chen2025janus,
  title={Janus-Pro: Unified Multimodal Understanding and Generation with Data and Model Scaling},
  author={Chen, Xiaokang and Wu, Zhiyu and Liu, Xingchao and Pan, Zizheng and Liu, Wen and Xie, Zhenda and Yu, Xingkai and Ruan, Chong},
  journal={arXiv preprint arXiv:2501.17811},
  year={2025}
}

@article{wu2024janus,
  title={Janus: Decoupling visual encoding for unified multimodal understanding and generation},
  author={Wu, Chengyue and Chen, Xiaokang and Wu, Zhiyu and Ma, Yiyang and Liu, Xingchao and Pan, Zizheng and Liu, Wen and Xie, Zhenda and Yu, Xingkai and Ruan, Chong and others},
  journal={arXiv preprint arXiv:2410.13848},
  year={2024}
}

@inproceedings{ma2024janusflow,
  title={Janusflow: Harmonizing autoregression and rectified flow for unified multimodal understanding and generation},
  author={Ma, Yiyang and Liu, Xingchao and Chen, Xiaokang and Liu, Wen and Wu, Chengyue and Wu, Zhiyu and Pan, Zizheng and Xie, Zhenda and Zhang, Haowei and Yu, Xingkai and others},
  booktitle={Proceedings of the IEEE/CVF Conference on Computer Vision and Pattern Recognition},
  pages={7739--7751},
  year={2025}
}

@article{Diao2025NEO,
  title        = {From Pixels to Words--Towards Native Vision-Language Primitives at Scale},
  author       = {Diao, Haiwen and Li, Mingxuan and Wu, Silei and Dai, Linjun and Wang, Xiaohua and Deng, Hanming and Lu, Lewei and Lin, Dahua and Liu, Ziwei},
  journal      = {arXiv preprint arXiv:2510.14979},
  year         = {2025}
}

@article{easi2025,
  title={Holistic Evaluation of Multimodal LLMs on Spatial Intelligence},
  author={Cai, Zhongang and Wang, Yubo and Sun, Qingping and Wang, Ruisi and Gu, Chenyang and Yin, Wanqi and Lin, Zhiqian and Yang, Zhitao and Wei, Chen and Shi, Xuanke and Deng, Kewang and Han, Xiaoyang and Chen, Zukai and Li, Jiaqi and Fan, Xiangyu and Deng, Hanming and Lu, Lewei and Li, Bo and Liu, Ziwei and Wang, Quan and Lin, Dahua and Yang, Lei},
  journal={arXiv preprint arXiv:2508.13142},
  year={2025}
}

@InProceedings{sensenova-si,
  title = {Scaling Spatial Intelligence with Multimodal Foundation Models},
  author = {Cai, Zhongang and Wang, Ruisi and Gu, Chenyang and Pu, Fanyi and Xu, Junxiang and Wang, Yubo and Yin, Wanqi and Yang, Zhitao and Wei, Chen and Sun, Qingping and Zhou, Tongxi and Li, Jiaqi and Pang, Hui En and Qian, Oscar and Wei, Yukun and Lin, Zhiqian and Shi, Xuanke and Deng, Kewang and Han, Xiaoyang and Chen, Zukai and Fan, Xiangyu and Deng, Hanming and Lu, Lewei and Pan, Liang and Li, Bo and Liu, Ziwei and Wang, Quan and Lin, Dahua and Yang, Lei},
  booktitle = {Proceedings of the IEEE/CVF Conference on Computer Vision and Pattern Recognition (CVPR)},
  year = {2026}
}

@inproceedings{gong2025pastfuture,
  author = {Gong, Ruihao and Bai, Shihao and Wu, Siyu and Fan, Yunqian and Wang, Zaijun and Li, Xiuhong and Yang, Hailong and Liu, Xianglong},
  title = {Past-Future Scheduler for LLM Serving under SLA Guarantees},
  year = {2025},
  isbn = {9798400710797},
  booktitle = {Proceedings of the 30th ACM International Conference on Architectural Support for Programming Languages and Operating Systems, Volume 2},
  pages = {798--813},
  numpages = {16}
}

@misc{lightllm,
  title = {{LightLLM}: A Python-based LLM Inference and Serving Framework},
  author = {{ModelTC}},
  url = {https://github.com/ModelTC/LightLLM},
  year = {2025}
}

@misc{lightx2v,
  title = {{LightX2V}: A Lightweight Video and Image Generation Inference Framework},
  author = {{ModelTC}},
  url = {https://github.com/ModelTC/LightX2V},
  year = {2025}
}

@article{zhao2025envisioning,
  title={Envisioning beyond the pixels: Benchmarking reasoning-informed visual editing},
  author={Zhao, Xiangyu and Zhang, Peiyuan and Tang, Kexian and Zhu, Xiaorong and Li, Hao and Chai, Wenhao and Zhang, Zicheng and Xia, Renqiu and Zhai, Guangtao and Yan, Junchi and others},
  journal={arXiv preprint arXiv:2504.02826},
  year={2025}
}

@article{liu2025step1x,
  title={Step1x-edit: A practical framework for general image editing},
  author={Liu, Shiyu and Han, Yucheng and Xing, Peng and Yin, Fukun and Wang, Rui and Cheng, Wei and Liao, Jiaqi and Wang, Yingming and Fu, Honghao and Han, Chunrui and others},
  journal={arXiv preprint arXiv:2504.17761},
  year={2025}
}

@article{ye2025imgedit,
  title={Imgedit: A unified image editing dataset and benchmark},
  author={Ye, Yang and He, Xianyi and Li, Zongjian and Lin, Bin and Yuan, Shenghai and Yan, Zhiyuan and Hou, Bohan and Yuan, Li},
  journal={arXiv preprint arXiv:2505.20275},
  year={2025}
}

@article{liu2025flow,
  title={Flow-grpo: Training flow matching models via online rl},
  author={Liu, Jie and Liu, Gongye and Liang, Jiajun and Li, Yangguang and Liu, Jiaheng and Wang, Xintao and Wan, Pengfei and Zhang, Di and Ouyang, Wanli},
  journal={arXiv preprint arXiv:2505.05470},
  year={2025}
}

@inproceedings{ma2025hpsv3widespectrumhumanpreference,
  title={Hpsv3: Towards wide-spectrum human preference score},
  author={Ma, Yuhang and Wu, Xiaoshi and Sun, Keqiang and Li, Hongsheng},
  booktitle={Proceedings of the IEEE/CVF International Conference on Computer Vision},
  pages={15086--15095},
  year={2025}
}

@article{cui2025paddleocr30technicalreport,
  title={Paddleocr 3.0 technical report},
  author={Cui, Cheng and Sun, Ting and Lin, Manhui and Gao, Tingquan and Zhang, Yubo and Liu, Jiaxuan and Wang, Xueqing and Zhang, Zelun and Zhou, Changda and Liu, Hongen and others},
  journal={arXiv preprint arXiv:2507.05595},
  year={2025}
}

@article{tuna2,
  title={TUNA-2: Pixel Embeddings Beat Vision Encoders
         for Unified Understanding and Generation},
  author={Liu, Zhiheng and Ren, Weiming and Huang, Xiaoke
          and Chen, Shoufa and Li, Tianhong and Chen, Mengzhao
          and Ji, Yatai and He, Sen and Schult, Jonas
          and Xiang, Tao and Chen, Wenhu and Luo, Ping
          and Zettlemoyer, Luke and Cong, Yuren},
  journal={arXiv preprint arXiv:2604.24763},
  year={2026}
}

@article{AlignTok,
  title={Aligning visual foundation encoders to tokenizers for diffusion models},
  author={Chen, Bowei and Bi, Sai and Tan, Hao and Zhang, He and Zhang, Tianyuan and Li, Zhengqi and Xiong, Yuanjun and Zhang, Jianming and Zhang, Kai},
  journal={arXiv preprint arXiv:2509.25162},
  year={2025}
}

@article{liu2025tuna,
  title={Tuna: Taming unified visual representations for native unified multimodal models},
  author={Liu, Zhiheng and Ren, Weiming and Liu, Haozhe and Zhou, Zijian and Chen, Shoufa and Qiu, Haonan and Huang, Xiaoke and An, Zhaochong and Yang, Fanny and Patel, Aditya and others},
  journal={arXiv preprint arXiv:2512.02014},
  year={2025}
}

@article{wu2025qwenimagetechnicalreport,
  title={Qwen-image technical report},
  author={Wu, Chenfei and Li, Jiahao and Zhou, Jingren and Lin, Junyang and Gao, Kaiyuan and Yan, Kun and Yin, Sheng-ming and Bai, Shuai and Xu, Xiao and Chen, Yilei and others},
  journal={arXiv preprint arXiv:2508.02324},
  year={2025}
}

@misc{flux2024,
  author={Black Forest Labs},
  title={FLUX},
  year={2024},
  url={https://github.com/black-forest-labs/flux}
}

@misc{gemini-2.0-flash,
  author={Google DeepMind},
  title={Gemini 2.0 Flash},
  year={2025},
  url={https://developers.googleblog.com/en/experiment-with-gemini-20-flash-native-image-generation}
}

@misc{flux-2-2025,
  author={Black Forest Labs},
  title={{FLUX.2: Frontier Visual Intelligence}},
  year={2025},
  url={https://bfl.ai/blog/flux-2}
}

@misc{minimax_image01,
  author = {{Minimax}},
  title  = {Image-01},
  url = {https://www.minimax.io/news/image-01},
  year   = {2025}
}

@misc{pimage,
  author = {{Pruna AI}},
  title  = {Pruna AI},
  url = {https://www.pruna.ai},
  year   = {2025}
}

@article{labs2025flux1kontextflowmatching,
  title={FLUX. 1 Kontext: Flow Matching for In-Context Image Generation and Editing in Latent Space},
  author={Labs, Black Forest and Batifol, Stephen and Blattmann, Andreas and Boesel, Frederic and Consul, Saksham and Diagne, Cyril and Dockhorn, Tim and English, Jack and English, Zion and Esser, Patrick and others},
  journal={arXiv preprint arXiv:2506.15742},
  year={2025}
}

@article{cai2025z,
  title={Z-Image: An Efficient Image Generation Foundation Model with Single-Stream Diffusion Transformer},
  author={Cai, Huanqia and Cao, Sihan and Du, Ruoyi and Gao, Peng and Hoi, Steven and Huang, Shijie and Hou, Zhaohui and Jiang, Dengyang and Jin, Xin and Li, Liangchen and others},
  journal={arXiv preprint arXiv:2511.22699},
  year={2025}
}

@misc{seedream45,
  title={Seedream 4.5},
  author={ByteDance},
  url={https://seed.bytedance.com/en/seedream4_5},
  year={2025}
}

@misc{Seedream5,
  title={Seedream 5.0 Lite},
  author={Bytedance Seed},
  year={2026},
  howpublished={\url{https://seed.bytedance.com/en/seedream5_0_lite}}
}

@article{seedream2025seedream,
  title={Seedream 4.0: Toward next-generation multimodal image generation},
  author={Seedream, Team and Chen, Yunpeng and Gao, Yu and Gong, Lixue and Guo, Meng and Guo, Qiushan and Guo, Zhiyao and Hou, Xiaoxia and Huang, Weilin and Huang, Yixuan and others},
  journal={arXiv preprint arXiv:2509.20427},
  year={2025}
}

@article{gao2025seedream,
  title={Seedream 3.0 technical report},
  author={Gao, Peiyuan and Zhuo, Le and Lin, Ziyi and Liu, Chris and Chen, Junsong and Du, Ruoyi and Luo, Enze and Zhang, Long and Chen, Guo and Zhang, Shengyu and others},
  journal={arXiv preprint arXiv:2504.11346},
  year={2025}
}

@misc{deepmind_gemini3proimage_2025,
  author={Google DeepMind},
  title={Gemini 3 Pro Image Model Card},
  year={2025},
  month={Nov},
  url={https://storage.googleapis.com/deepmind-media/Model-Cards/Gemini-3-Pro-Image-Model-Card.pdf}
}

@misc{nanobanana2,
  title={Nano Banana 2: Combining Pro capabilities with lightning-fast speed},
  author={Google},
  year={2026},
  url={https://blog.google/innovation-and-ai/technology/ai/nano-banana-2/}
}

@misc{google2025gemini25flashmodelcard,
  author={Google DeepMind},
  title={Gemini 2.5 Flash \& Gemini 2.5 Flash Image Model Card},
  url={https://storage.googleapis.com/deepmind-media/Model-Cards/Gemini-2-5-Flash-Model-Card.pdf},
  year={2025}
}

@misc{GPT-Image-1,
  author={OpenAI},
  title={GPT-Image-1},
  url={https://platform.openai.com/docs/guides/image-generation},
  year={2025}
}

@misc{GPT-Image-1.5,
  title={GPT-Image-1.5},
  author={OpenAI},
  url={https://openai.com/index/new-chatgpt-images-is-here/},
  year={2025}
}

@article{cui2025emu35nativemultimodalmodels,
  title={Emu3. 5: Native multimodal models are world learners},
  author={Cui, Yufeng and Chen, Honghao and Deng, Haoge and Huang, Xu and Li, Xinghang and Liu, Jirong and Liu, Yang and Luo, Zhuoyan and Wang, Jinsheng and Wang, Wenxuan and others},
  journal={arXiv preprint arXiv:2510.26583},
  year={2025}
}

@inproceedings{emu2,
  title={Generative multimodal models are in-context learners},
  author={Sun, Quan and Cui, Yufeng and Zhang, Xiaosong and Zhang, Fan and Yu, Qiying and Wang, Yueze and Rao, Yongming and Liu, Jingjing and Huang, Tiejun and Wang, Xinlong},
  booktitle={Proceedings of the IEEE/CVF Conference on Computer Vision and Pattern Recognition},
  pages={14398--14409},
  year={2024}
}

@misc{glm_image,
  author       = {Z.ai},
  title        = {GLM-Image: Auto-regressive for Dense-knowledge and High-fidelity Image Generation},
  url = {https://z.ai/blog/glm-image/},
  year         = {2026}
  }

@article{wan2025,
  title={Wan: Open and advanced large-scale video generative models},
  author={Wan, Team and Wang, Ang and Ai, Baole and Wen, Bin and Mao, Chaojie and Xie, Chen-Wei and Chen, Di and Yu, Feiwu and Zhao, Haiming and Yang, Jianxiao and others},
  journal={arXiv preprint arXiv:2503.20314},
  year={2025}
}

@article{team2025longcat,
  title={LongCat-Image Technical Report},
  author={Team, Meituan LongCat and Ma, Hanghang and Tan, Haoxian and Huang, Jiale and Wu, Junqiang and He, Jun-Yan and Gao, Lishuai and Xiao, Songlin and Wei, Xiaoming and Ma, Xiaoqi and others},
  journal={arXiv preprint arXiv:2512.07584},
  year={2025}
}

@article{tian2026internvludemocratizingunifiedmultimodal,
  title={Internvl-u: Democratizing unified multimodal models for understanding, reasoning, generation and editing},
  author={Tian, Changyao and Yang, Danni and Chen, Guanzhou and Cui, Erfei and Wang, Zhaokai and Duan, Yuchen and Yin, Penghao and Chen, Sitao and Yang, Ganlin and Liu, Mingxin and others},
  journal={arXiv preprint arXiv:2603.09877},
  year={2026}
}

@article{xie2025show,
  title={Show-o2: Improved Native Unified Multimodal Models},
  author={Xie, Jinheng and Mao, Weijia and Bai, Zechen and Zhang, David Junhao and Wang, Weihao and Lin, Kevin Qinghong and Gu, Yuchao and Chen, Zhijie and Yang, Zhenheng and Shou, Mike Zheng and others},
  journal={arXiv preprint arXiv:2510.23218},
  year={2025}
}

@article{lin2025uniworld,
  title={Uniworld: High-resolution semantic encoders for unified visual understanding and generation},
  author={Lin, Bin and Li, Zongjian and Cheng, Xinhua and Niu, Yuwei and Ye, Yang and He, Xianyi and Yuan, Shenghai and Yu, Wangbo and Wang, Shaodong and Ge, Yunyang and others},
  journal={arXiv preprint arXiv:2506.03147},
  year={2025}
}

@article{li2025uniworld,
  title={Uniworld-v2: Reinforce image editing with diffusion negative-aware finetuning and mllm implicit feedback},
  author={Li, Zongjian and Liu, Zheyuan and Zhang, Qihui and Lin, Bin and Wu, Feize and Yuan, Shenghai and Yan, Zhiyuan and Ye, Yang and Yu, Wangbo and Niu, Yuwei and others},
  journal={arXiv preprint arXiv:2510.16888},
  year={2025}
}

@article{wu2025omnigen2,
  title={OmniGen2: Exploration to Advanced Multimodal Generation},
  author={Wu, Chenyuan and Zheng, Pengfei and Yan, Ruiran and Xiao, Shitao and Luo, Xin and Wang, Yueze and Li, Wanli and Jiang, Xiyan and Liu, Yexin and Zhou, Junjie and others},
  journal={arXiv preprint arXiv:2506.18871},
  year={2025}
}

@article{vqvae,
  title={Neural discrete representation learning},
  author={Van Den Oord, Aaron and Vinyals, Oriol and others},
  journal={Advances in Neural Information Processing Systems},
  volume={30},
  year={2017}
}

@inproceedings{REPA-E,
  title={Repa-e: Unlocking vae for end-to-end tuning of latent diffusion transformers},
  author={Leng, Xingjian and Singh, Jaskirat and Hou, Yunzhong and Xing, Zhenchang and Xie, Saining and Zheng, Liang},
  booktitle={Proceedings of the IEEE/CVF International Conference on Computer Vision},
  pages={18262--18272},
  year={2025}
}

@article{xiao2024omnigen,
  title={Omnigen: Unified image generation},
  author={Xiao, Shitao and Wang, Yueze and Zhou, Junjie and Yuan, Huaying and Xing, Xingrun and Yan, Ruiran and Wang, Shuting and Huang, Tiejun and Liu, Zheng},
  journal={arXiv preprint arXiv:2409.11340},
  year={2024}
}

@article{xin2025lumina,
  title={Lumina-dimoo: An omni diffusion large language model for multi-modal generation and understanding},
  author={Xin, Yi and Qin, Qi and Luo, Siqi and Zhu, Kaiwen and Yan, Juncheng and Tai, Yan and Lei, Jiayi and Cao, Yuewen and Wang, Keqi and Wang, Yibin and others},
  journal={arXiv preprint arXiv:2510.06308},
  year={2025}
}

@article{wang2025ovis,
  title={Ovis-U1 Technical Report},
  author={Wang, Guo-Hua and Zhao, Shanshan and Zhang, Xinjie and Cao, Liangfu and Zhan, Pengxin and Duan, Lunhao and Lu, Shiyin and Fu, Minghao and Chen, Xiaohao and Zhao, Jianshan and others},
  journal={arXiv preprint arXiv:2506.23044},
  year={2025}
}

@article{cao2025hunyuanimage,
  title={HunyuanImage 3.0 Technical Report},
  author={Cao, Siyu and Chen, Hangting and Chen, Peng and Cheng, Yiji and Cui, Yutao and Deng, Xinchi and Dong, Ying and Gong, Kipper and Gu, Tianpeng and Gu, Xiusen and others},
  journal={arXiv preprint arXiv:2509.23951},
  year={2025}
}

@article{li2025onecat,
  title={Onecat: Decoder-only auto-regressive model for unified understanding and generation},
  author={Li, Han and Peng, Xinyu and Wang, Yaoming and Peng, Zelin and Chen, Xin and Weng, Rongxiang and Wang, Jingang and Cai, Xunliang and Dai, Wenrui and Xiong, Hongkai},
  journal={arXiv preprint arXiv:2509.03498},
  year={2025}
}

@article{liao2025mogao,
  title={Mogao: An omni foundation model for interleaved multi-modal generation},
  author={Liao, Chao and Liu, Liyang and Wang, Xun and Luo, Zhengxiong and Zhang, Xinyu and Zhao, Wenliang and Wu, Jie and Li, Liang and Tian, Zhi and Huang, Weilin},
  journal={arXiv preprint arXiv:2505.05472},
  year={2025}
}

@inproceedings{esser2024scaling,
  title={Scaling rectified flow transformers for high-resolution image synthesis},
  author={Esser, Patrick and Kulal, Sumith and Blattmann, Andreas and Entezari, Rahim and M{\"u}ller, Jonas and Saini, Harry and Levi, Yam and Lorenz, Dominik and Sauer, Axel and Boesel, Frederic and others},
  booktitle={International Conference on Machine Learning},
  year={2024}
}

@article{ge2024seed,
  title={SEED-X: Multimodal Models with Unified Multi-granularity Comprehension and Generation},
  author={Ge, Yuying and Zhao, Sijie and Zhu, Jinguo and Ge, Yixiao and Yi, Kun and Song, Lin and Li, Chen and Ding, Xiaohan and Shan, Ying},
  journal={arXiv preprint arXiv:2404.14396},
  year={2024}
}

@article{ge2023making,
  title={Making llama see and draw with seed tokenizer},
  author={Ge, Yuying and Zhao, Sijie and Zeng, Ziyun and Ge, Yixiao and Li, Chen and Wang, Xintao and Shan, Ying},
  journal={arXiv preprint arXiv:2310.01218},
  year={2023}
}

@article{chern2024anole,
  title={ANOLE: An Open, Autoregressive, Native Large Multimodal Models for Interleaved Image-Text Generation},
  author={Chern, Ethan and Su, Jiadi and Ma, Yan and Liu, Pengfei},
  journal={arXiv preprint arXiv:2407.06135},
  year={2024}
}

@article{zheng2023minigpt,
  title={{MiniGPT-5}: Interleaved vision-and-language generation via generative vokens},
  author={Zheng, Kaizhi and He, Xuehai and Wang, Xin Eric},
  journal={arXiv preprint arXiv:2310.02239},
  year={2023}
}

@inproceedings{wunext,
  title={{NExT-GPT}: {Any-to-Any Multimodal LLM}},
  author={Wu, Shengqiong and Fei, Hao and Qu, Leigang and Ji, Wei and Chua, Tat-Seng},
  booktitle={Proceedings of the International Conference on Machine Learning},
  year={2024}
}

@article{QLIP,
  title={Qlip: Text-aligned visual tokenization unifies auto-regressive multimodal understanding and generation},
  author={Zhao, Yue and Xue, Fuzhao and Reed, Scott and Fan, Linxi and Zhu, Yuke and Kautz, Jan and Yu, Zhiding and Kr{\"a}henb{\"u}hl, Philipp and Huang, De-An},
  journal={arXiv preprint arXiv:2502.05178},
  year={2025}
}

@article{TokLIP,
  title={Toklip: Marry visual tokens to clip for multimodal comprehension and generation},
  author={Lin, Haokun and Wang, Teng and Ge, Yixiao and Ge, Yuying and Lu, Zhichao and Wei, Ying and Zhang, Qingfu and Sun, Zhenan and Shan, Ying},
  journal={arXiv preprint arXiv:2505.05422},
  year={2025}
}

@article{wu2024vila,
  title={VILA-U: a Unified Foundation Model Integrating Visual Understanding and Generation},
  author={Wu, Yecheng and Zhang, Zhuoyang and Chen, Junyu and Tang, Haotian and Li, Dacheng and Fang, Yunhao and Zhu, Ligeng and Xie, Enze and Yin, Hongxu and Yi, Li and others},
  journal={arXiv preprint arXiv:2409.04429},
  year={2024}
}

@misc{betker2023dalle3,
  title={Improving Image Generation with Better Captions},
  author={Betker, James and Goh, Gabriel and Jing, Li and Brooks, Tim and Wang, Jianfeng and Li, Linjie and Ouyang, Long and Zhuang, Juntang and Lee, Joyce and Guo, Yufei and Manassra, Wesam and Dhariwal, Prafulla and Chu, Casey and Jiao, Yunxin and Ramesh, Aditya},
  url={https://cdn.openai.com/papers/dall-e-3.pdf},
  year={2023}
}

@article{cai2025hidream,
  title={HiDream-I1: An Open-Source High-Efficient Image Generative Foundation Model},
  author={Cai, Qi and Chen, Jingwen and Chen, Yang and Li, Yehao and Long, Fuchen and Pan, Yingwei and Qiu, Zhaofan and Zhang, Yiheng and Gao, Fengbin and Xu, Peihan and others},
  journal={arXiv preprint arXiv:2505.22705},
  year={2025}
}

@misc{google2025imagen4,
  author={Google},
  title={Imagen 4 Model Card},
  year={2025},
  url={https://storage.googleapis.com/deepmind-media/Model-Cards/Imagen-4-Model-Card.pdf}
}

@article{Imagen3,
  title={Imagen 3},
  author={Baldridge, Jason and Bauer, Jakob and Bhutani, Mukul and Brichtova, Nicole and Bunner, Andrew and Castrejon, Lluis and Chan, Kelvin and Chen, Yichang and Dieleman, Sander and Du, Yuqing and others},
  journal={arXiv preprint arXiv:2408.07009},
  year={2024}
}

@misc{recraftv3,
  author={Recraft},
  title={Recraft V3},
  year={2024},
  url={https://www.recraft.ai/docs/recraft-models/recraft-V3}
}

@misc{kuaishou2025kolors,
  author={Kolors Team, Kuaishou},
  title={Kolors 2.0},
  url={https://kolors.kuaishou.com/},
  year={2025}
}

@misc{midjourneyV7,
  author={Midjourney},
  title={Midjourney V7},
  year={2025},
  url={https://www.midjourney.com/home}
}

@inproceedings{han2025infinity,
  title={Infinity: Scaling bitwise autoregressive modeling for high-resolution image synthesis},
  author={Han, Jian and Liu, Jinlai and Jiang, Yi and Yan, Bin and Zhang, Yuqi and Yuan, Zehuan and Peng, Bingyue and Liu, Xiaobing},
  booktitle={Proceedings of the Computer Vision and Pattern Recognition Conference},
  pages={15733--15744},
  year={2025}
}

@article{huang2025illume+,
  title={Illume+: Illuminating unified mllm with dual visual tokenization and diffusion refinement},
  author={Huang, Runhui and Wang, Chunwei and Yang, Junwei and Lu, Guansong and Yuan, Yunlong and Han, Jianhua and Hou, Lu and Zhang, Wei and Hong, Lanqing and Zhao, Hengshuang and others},
  journal={arXiv preprint arXiv:2504.01934},
  year={2025}
}

@article{vteam2025glm45vglm41vthinkingversatilemultimodal,
  title={Glm-4.5 v and glm-4.1 v-thinking: Towards versatile multimodal reasoning with scalable reinforcement learning},
  author={Hong, Wenyi and Yu, Wenmeng and Gu, Xiaotao and Wang, Guo and Gan, Guobing and Tang, Haomiao and Cheng, Jiale and Qi, Ji and Ji, Junhui and Pan, Lihang and others},
  journal={arXiv preprint arXiv:2507.01006},
  year={2025}
}

@article{team2026longcat,
  title={LongCat-Next: Lexicalizing Modalities as Discrete Tokens},
  author={Team, Meituan LongCat and Xiao, Bin and Wang, Chao and Li, Chengjiang and Zhang, Chi and Peng, Chong and Yu, Hang and Yang, Hao and Yan, Haonan and Sun, Haoze and others},
  journal={arXiv preprint arXiv:2603.27538},
  year={2026}
}

@article{wei2025skywork,
  title={Skywork unipic 2.0: Building kontext model with online rl for unified multimodal model},
  author={Wei, Hongyang and Xu, Baixin and Liu, Hongbo and Wu, Size and Liu, Jie and Peng, Yi and Wang, Peiyu and Liu, Zexiang and He, Jingwen and Xietian, Yidan and others},
  journal={arXiv preprint arXiv:2509.04548},
  year={2025}
}

@misc{gpt_image_2,
  author    = {OpenAI},
  title     = {Introducing ChatGPT Images 2.0},
  year      = {2026},
  journal   = {OpenAI blog},
  url       = {https://openai.com/index/introducing-chatgpt-images-2-0/}
}

@misc{joyai_image,
  author    = {JD Open Source},
  title     = {JoyAI-Image: Awakening Spatial Intelligence in Unified Multimodal Understanding and Generation},
  year      = {2026},
  journal   = {GitHub repository},
  url       = {https://github.com/jd-opensource/JoyAI-Image}
}

@article{dmd2,
  title={Improved distribution matching distillation for fast image synthesis},
  author={Yin, Tianwei and Gharbi, Micha{\"e}l and Park, Taesung and Zhang, Richard and Shechtman, Eli and Durand, Fredo and Freeman, William T},
  journal={Advances in Neural Information Processing Systems},
  volume={37},
  pages={47455--47487},
  year={2024}
}

@article{phased_dmd,
  title={Phased dmd: Few-step distribution matching distillation via score matching within subintervals},
  author={Fan, Xiangyu and Qiu, Zesong and Wu, Zhuguanyu and Wang, Fanzhou and Lin, Zhiqian and Ren, Tianxiang and Lin, Dahua and Gong, Ruihao and Yang, Lei},
  journal={arXiv preprint arXiv:2510.27684},
  year={2025}
}

@misc{sensenova2026neounify,
  title        = {NEO-unify: Building Native Multimodal Unified Models End to End},
  author       = {SenseNova},
  year         = {2026},
  journal   = {Hugging Face blog},
  url       = {https://huggingface.co/blog/sensenova/neo-unify}
}

@article{xing2026wan,
  title={Wan-Weaver: Interleaved Multi-modal Generation via Decoupled Training},
  author={Xing, Jinbo and Jiang, Zeyinzi and Tuo, Yuxiang and Mao, Chaojie and Gai, Xiaotang and Chen, Xi and Zhang, Jingfeng and Pan, Yulin and Han, Zhen and Xiao, Jie and others},
  journal={arXiv preprint arXiv:2603.25706},
  year={2026}
}

@article{wang2026very,
  title={A very big video reasoning suite},
  author={Wang, Maijunxian and Wang, Ruisi and  Lin, Juyi and Ji, Ran and
             Wiedemer, Thadd{\"a}us and Gao, Qingying and Luo, Dezhi and
             Qian, Yaoyao and Huang, Lianyu and Hong, Zelong and Ge, Jiahui and
             Ma, Qianli and He, Hang and Zhou, Yifan and Guo, Lingzi and
             Mei, Lantao and Li, Jiachen and Xing, Hanwen and Zhao, Tianqi and
             Yu, Fengyuan and Xiao, Weihang and Jiao, Yizheng and
             Hou, Jianheng and Zhang, Danyang and Xu, Pengcheng and
             Zhong, Boyang and Zhao, Zehong and Fang, Gaoyun and Kitaoka, John and
             Xu, Yile and Xu, Hua bureau and Blacutt, Kenton and Nguyen, Tin and
             Song, Siyuan and Sun, Haoran and Wen, Shaoyue and He, Linyang and
             Wang, Runming and Wang, Yanzhi and Yang, Mengyue and Ma, Ziqiao and
             Milli{\`e}re, Rapha{\"e}l and Shi, Freda and Vasconcelos, Nuno and
             Khashabi, Daniel and Yuille, Alan and Du, Yilun and Liu, Ziming and
             Lin, Dahua and Liu, Ziwei and Kumar, Vikash and Li, Yijiang and
             Yang, Lei and Cai, Zhongang and Deng, Hokin},
  journal={arXiv preprint arXiv:2602.20159},
  year={2026}
}

@article{sun2023eva,
  title={Eva-clip: Improved training techniques for clip at scale},
  author={Sun, Quan and Fang, Yuxin and Wu, Ledell and Wang, Xinlong and Cao, Yue},
  journal={arXiv preprint arXiv:2303.15389},
  year={2023}
}

@inproceedings{VLP:CLIP,
  title={Learning transferable visual models from natural language supervision},
  author={Radford, Alec and Kim, Jong Wook and Hallacy, Chris and Ramesh, Aditya and Goh, Gabriel and Agarwal, Sandhini and Sastry, Girish and Askell, Amanda and Mishkin, Pamela and Clark, Jack and others},
  booktitle={International conference on machine learning},
  pages={8748--8763},
  year={2021},
}

@inproceedings{VLP:SigLIP,
  title={Sigmoid loss for language image pre-training},
  author={Zhai, Xiaohua and Mustafa, Basil and Kolesnikov, Alexander and Beyer, Lucas},
  booktitle={Proceedings of the IEEE/CVF international conference on computer vision},
  pages={11975--11986},
  year={2023}
}

@article{vae,
  title={Auto-encoding variational bayes},
  author={Kingma, Diederik P and Welling, Max},
  journal={arXiv preprint arXiv:1312.6114},
  year={2013}
}

@inproceedings{vavae,
  title={Reconstruction vs. generation: Taming optimization dilemma in latent diffusion models},
  author={Yao, Jingfeng and Yang, Bin and Wang, Xinggang},
  booktitle={Proceedings of the Computer Vision and Pattern Recognition Conference},
  pages={15703--15712},
  year={2025}
}

@inproceedings{blip2,
  title={Blip-2: Bootstrapping language-image pre-training with frozen image encoders and large language models},
  author={Li, Junnan and Li, Dongxu and Savarese, Silvio and Hoi, Steven},
  booktitle={International conference on machine learning},
  pages={19730--19742},
  year={2023},
  organization={PMLR}
}

@article{flamingo,
  title={Flamingo: a visual language model for few-shot learning},
  author={Alayrac, Jean-Baptiste and Donahue, Jeff and Luc, Pauline and Miech, Antoine and Barr, Iain and Hasson, Yana and Lenc, Karel and Mensch, Arthur and Millican, Katherine and Reynolds, Malcolm and others},
  journal={Advances in Neural Information Processing Systems},
  volume={35},
  pages={23716--23736},
  year={2022}
}

@article{tong2026beyond,
  title={Beyond language modeling: An exploration of multimodal pretraining},
  author={Tong, Shengbang and Fan, David and Nguyen, John and Brown, Ellis and Zhou, Gaoyue and Qian, Shengyi and Zheng, Boyang and Vallaeys, Th{\'e}ophane and Han, Junlin and Fergus, Rob and others},
  journal={arXiv preprint arXiv:2603.03276},
  year={2026}
}

@article{claweval,
  title={Claw-Eval: Toward Trustworthy Evaluation of Autonomous Agents},
  author={Ye, Bowen and Li, Rang and Yang, Qibin and Liu, Yuanxin and Yao, Linli and Lv, Hanglong and Xie, Zhihui and An, Chenxin and Li, Lei and Kong, Lingpeng and Liu, Qi and Sui, Zhifang and Yang, Tong},
  journal={arXiv preprint arXiv:2604.06132},
  year={2026}
}

@article{tau2bench,
  title={$\tau$-bench: A Benchmark for Tool-Agent-User Interaction in Real-World Domains},
  author={Yao, Shunyu and Narasimhan, Karthik and Cao, Jian},
  journal={arXiv preprint arXiv:2406.12045},
  year={2024}
}

@inproceedings{mathew2022infographicvqa,
  title={Infographicvqa},
  author={Mathew, Minesh and Bagal, Viraj and Tito, Rub{\`e}n and Karatzas, Dimosthenis and Valveny, Ernest and Jawahar, CV},
  booktitle={Proceedings of the IEEE/CVF Winter Conference on Applications of Computer Vision},
  pages={1697--1706},
  year={2022}
}

@article{kimik25,
  title   = {{Kimi K2}: Open Agentic Intelligence},
  author  = {{Kimi Team}},
  journal = {arXiv preprint arXiv:2507.20534},
  year    = {2025}
}

@article{chen2024we,
  title={Are we on the right way for evaluating large vision-language models?},
  author={Chen, Lin and Li, Jinsong and Dong, Xiaoyi and Zhang, Pan and Zang, Yuhang and Chen, Zehui and Duan, Haodong and Wang, Jiaqi and Qiao, Yu and Lin, Dahua and others},
  journal={Advances in Neural Information Processing Systems},
  volume={37},
  pages={27056--27087},
  year={2024}
}

@article{liu2024ocrbench,
  title={Ocrbench: on the hidden mystery of ocr in large multimodal models},
  author={Liu, Yuliang and Li, Zhang and Huang, Mingxin and Yang, Biao and Yu, Wenwen and Li, Chunyuan and Yin, Xu-Cheng and Liu, Cheng-Lin and Jin, Lianwen and Bai, Xiang},
  journal={Science China Information Sciences},
  volume={67},
  number={12},
  pages={220102},
  year={2024},
  publisher={Springer}
}

@article{hiippala2021ai2d,
  title={AI2D-RST},
  author={Hiippala, Tuomo and Alikhani, Malihe and Haverinen, Jonas and Kalliokoski, Timo and Logacheva, Evanfiya and Orekhova, Serafina and Tuomainen, Aino and Stone, Matthew and Bateman, John A},
  journal={Language Resources and Evaluation},
  volume={55},
  number={3},
  pages={661--688},
  year={2021},
  publisher={JSTOR}
}

@article{chen2026babyvision,
  title={BabyVision: Visual Reasoning Beyond Language},
  author={Chen, Liang and Xie, Weichu and Liang, Yiyan and He, Hongfeng and Zhao, Hans and Yang, Zhibo and Huang, Zhiqi and Wu, Haoning and Lu, Haoyu and Bao, Yiping and others},
  journal={arXiv preprint arXiv:2601.06521},
  year={2026}
}

@article{li2025tir,
  title={TIR-Bench: A Comprehensive Benchmark for Agentic Thinking-with-Images Reasoning},
  author={Li, Ming and Zhong, Jike and Zhao, Shitian and Zhang, Haoquan and Lin, Shaoheng and Lai, Yuxiang and Wei, Chen and Psounis, Konstantinos and Zhang, Kaipeng},
  journal={arXiv preprint arXiv:2511.01833},
  year={2025}
}

@inproceedings{yue2025mmmu,
  title={Mmmu-pro: A more robust multi-discipline multimodal understanding benchmark},
  author={Yue, Xiang and Zheng, Tianyu and Ni, Yuansheng and Wang, Yubo and Zhang, Kai and Tong, Shengbang and Sun, Yuxuan and Yu, Botao and Zhang, Ge and Sun, Huan and others},
  booktitle={Proceedings of the 63rd Annual Meeting of the Association for Computational Linguistics (Volume 1: Long Papers)},
  pages={15134--15186},
  year={2025}
}

@article{wang2024measuring,
  title={Measuring multimodal mathematical reasoning with math-vision dataset},
  author={Wang, Ke and Pan, Junting and Shi, Weikang and Lu, Zimu and Ren, Houxing and Zhou, Aojun and Zhan, Mingjie and Li, Hongsheng},
  journal={Advances in Neural Information Processing Systems},
  volume={37},
  pages={95095--95169},
  year={2024}
}

@article{wang2024mmlu,
  title={Mmlu-pro: A more robust and challenging multi-task language understanding benchmark},
  author={Wang, Yubo and Ma, Xueguang and Zhang, Ge and Ni, Yuansheng and Chandra, Abhranil and Guo, Shiguang and Ren, Weiming and Arulraj, Aaran and He, Xuan and Jiang, Ziyan and others},
  journal={Advances in Neural Information Processing Systems},
  volume={37},
  pages={95266--95290},
  year={2024}
}

@inproceedings{gema2025we,
  title={Are we done with mmlu?},
  author={Gema, Aryo Pradipta and Leang, Joshua Ong Jun and Hong, Giwon and Devoto, Alessio and Mancino, Alberto Carlo Maria and Saxena, Rohit and He, Xuanli and Zhao, Yu and Du, Xiaotang and Madani, Mohammad Reza Ghasemi and others},
  booktitle={Proceedings of the 2025 Conference of the Nations of the Americas Chapter of the Association for Computational Linguistics: Human Language Technologies (Volume 1: Long Papers)},
  pages={5069--5096},
  year={2025}
}

@article{huang2023c,
  title={C-eval: A multi-level multi-discipline chinese evaluation suite for foundation models},
  author={Huang, Yuzhen and Bai, Yuzhuo and Zhu, Zhihao and Zhang, Junlei and Zhang, Jinghan and Su, Tangjun and Liu, Junteng and Lv, Chuancheng and Zhang, Yikai and Fu, Yao and others},
  journal={Advances in Neural Information Processing Systems},
  volume={36},
  pages={62991--63010},
  year={2023}
}

@article{du2025supergpqa,
  title={Supergpqa: Scaling llm evaluation across 285 graduate disciplines},
  author={Du, Xinrun and Yao, Yifan and Ma, Kaijing and Wang, Bingli and Zheng, Tianyu and Zhu, King and Liu, Minghao and Liang, Yiming and Jin, Xiaolong and Wei, Zhenlin and others},
  journal={arXiv preprint arXiv:2502.14739},
  year={2025}
}

@article{zhou2023instruction,
  title={Instruction-following evaluation for large language models},
  author={Zhou, Jeffrey and Lu, Tianjian and Mishra, Swaroop and Brahma, Siddhartha and Basu, Sujoy and Luan, Yi and Zhou, Denny and Hou, Le},
  journal={arXiv preprint arXiv:2311.07911},
  year={2023}
}

@article{zhang2025if,
  title={IF-Bench: Benchmarking and Enhancing MLLMs for Infrared Images with Generative Visual Prompting},
  author={Zhang, Tao and Hong, Yuyang and Xia, Yang and Ding, Kun and Zhang, Zeyu and Wang, Ying and Xiang, Shiming and Pan, Chunhong},
  journal={arXiv preprint arXiv:2512.09663},
  year={2025}
}

@misc{VLM:Fuyu-8b,
  author = {Bavishi, Rohan and Elsen, Erich and Hawthorne, Curtis and Nye, Maxwell and Odena, Augustus and Somani, Arushi and  Ta\c{s}\i{}rlar, Sa\u{g}nak},
  title = {Introducing our multimodal models},
  url = {https://www.adept.ai/blog/fuyu-8b},
  year = {2023}
}

@article{VLM:EVE,
  title={Unveiling encoder-free vision-language models},
  author={Diao, Haiwen and Cui, Yufeng and Li, Xiaotong and Wang, Yueze and Lu, Huchuan and Wang, Xinlong},
  journal={Advances in Neural Information Processing Systems},
  volume={37},
  pages={52545--52567},
  year={2024}
}

@article{VLM:SOLO,
  title={A single transformer for scalable vision-language modeling},
  author={Chen, Yangyi and Wang, Xingyao and Peng, Hao and Ji, Heng},
  journal={Transactions on Machine Learning Research},
  year={2024}
}

@article{VLM:HaploVL,
  title={Haplovl: A single-transformer baseline for multi-modal understanding},
  author={Yang, Rui and Song, Lin and Xiao, Yicheng and Huang, Runhui and Ge, Yixiao and Shan, Ying and Zhao, Hengshuang},
  journal={arXiv preprint arXiv:2503.14694},
  year={2025}
}

@inproceedings{VLM:SAIL,
  title={The scalability of simplicity: Empirical analysis of vision-language learning with a single transformer},
  author={Lei, Weixian and Wang, Jiacong and Wang, Haochen and Li, Xiangtai and Liew, Jun Hao and Feng, Jiashi and Huang, Zilong},
  booktitle={Proceedings of the IEEE/CVF International Conference on Computer Vision},
  pages={20758--20769},
  year={2025}
}

@article{VLM:VoRA,
  title={Vision as lora},
  author={Wang, Han and Ye, Yongjie and Li, Bingru and Nie, Yuxiang and Lu, Jinghui and Tang, Jingqun and Wang, Yanjie and Huang, Can},
  journal={arXiv preprint arXiv:2503.20680},
  year={2025}
}

@inproceedings{VLM:BREEN,
  title={BREEN: bridge data-efficient encoder-free multimodal learning with learnable queries},
  author={Li, Tianle and Rao, Yongming and Hu, Winston and Cheng, Yu},
  booktitle={Proceedings of the IEEE/CVF Winter Conference on Applications of Computer Vision},
  pages={5384--5395},
  year={2026}
}

@inproceedings{VLM:EVEv2,
  title={Evev2: Improved baselines for encoder-free vision-language models},
  author={Diao, Haiwen and Li, Xiaotong and Cui, Yufeng and Wang, Yueze and Deng, Haoge and Pan, Ting and Wang, Wenxuan and Lu, Huchuan and Wang, Xinlong},
  booktitle={Proceedings of the IEEE/CVF International Conference on Computer Vision},
  pages={21014--21025},
  year={2025}
}

@inproceedings{VLM:HoVLE,
  title={Hovle: Unleashing the power of monolithic vision-language models with holistic vision-language embedding},
  author={Tao, Chenxin and Su, Shiqian and Zhu, Xizhou and Zhang, Chenyu and Chen, Zhe and Liu, Jiawen and Wang, Wenhai and Lu, Lewei and Huang, Gao and Qiao, Yu and others},
  booktitle={Proceedings of the Computer Vision and Pattern Recognition Conference},
  pages={14559--14569},
  year={2025}
}

@article{VLM:Mono-InternVL-1.5,
  title={Mono-internvl-1.5: Towards cheaper and faster monolithic multimodal large language models},
  author={Luo, Gen and Dou, Wenhan and Li, Wenhao and Wang, Zhaokai and Yang, Xue and Tian, Changyao and Li, Hao and Wang, Weiyun and Wang, Wenhai and Zhu, Xizhou and others},
  journal={arXiv preprint arXiv:2507.12566},
  year={2025}
}

@inproceedings{VLM:Mono-InternVL,
  title={Mono-internvl: Pushing the boundaries of monolithic multimodal large language models with endogenous visual pre-training},
  author={Luo, Gen and Yang, Xue and Dou, Wenhan and Wang, Zhaokai and Liu, Jiawen and Dai, Jifeng and Qiao, Yu and Zhu, Xizhou},
  booktitle={Proceedings of the IEEE/CVF Conference on Computer Vision and Pattern Recognition},
  pages={24960--24971},
  year={2025}
}

@article{shi2025latent,
  title={Latent diffusion model without variational autoencoder},
  author={Shi, Minglei and Wang, Haolin and Zheng, Wenzhao and Yuan, Ziyang and Wu, Xiaoshi and Wang, Xintao and Wan, Pengfei and Zhou, Jie and Lu, Jiwen},
  journal={arXiv preprint arXiv:2510.15301},
  year={2025}
}

@inproceedings{rombach2021highresolution,
  title={High-resolution image synthesis with latent diffusion models},
  author={Rombach, Robin and Blattmann, Andreas and Lorenz, Dominik and Esser, Patrick and Ommer, Bj{\"o}rn},
  booktitle={Proceedings of the IEEE/CVF Conference on Computer Vision and Pattern Recognition},
  pages={10684--10695},
  year={2022}
}

@article{podell2023sdxl,
  title={Sdxl: Improving latent diffusion models for high-resolution image synthesis},
  author={Podell, Dustin and English, Zion and Lacey, Kyle and Blattmann, Andreas and Dockhorn, Tim and M{\"u}ller, Jonas and Penna, Joe and Rombach, Robin},
  journal={arXiv preprint arXiv:2307.01952},
  year={2023}
}

@misc{lopez2025sd3,
  author       = {Joshua Lopez},
  title        = {Stable Diffusion 3: Research Paper},
  url = {https://stability.ai/news/stable-diffusion-3-research-paper},
  note         = {Stability AI},
  year         = {2025},
}

@misc{gemma42026,
  title        = {Gemma 4: Byte for byte, the most capable open models},
  author       = {Google DeepMind},
  year         = {2026},
  url = {https://blog.google/innovation-and-ai/technology/developers-tools/gemma-4/},
}

@article{zhang2025context,
  title={In-context edit: Enabling instructional image editing with in-context generation in large scale diffusion transformer},
  author={Zhang, Zechuan and Xie, Ji and Lu, Yu and Yang, Zongxin and Yang, Yi},
  journal={arXiv preprint arXiv:2504.20690},
  year={2025}
}

@inproceedings{Datasets:MSCOCO,
  title={Microsoft coco: Common objects in context},
  author={Lin, Tsung-Yi and Maire, Michael and Belongie, Serge and Hays, James and Perona, Pietro and Ramanan, Deva and Doll{\'a}r, Piotr and Zitnick, C Lawrence},
  booktitle={European conference on computer vision},
  pages={740--755},
  year={2014},
}

@article{chen2025blip3o,
  title={BLIP3-o: A Family of Fully Open Unified Multimodal Models-Architecture, Training and Dataset}, 
  author={Jiuhai Chen and Zhiyang Xu and Xichen Pan and Yushi Hu and Can Qin and Tom Goldstein and Lifu Huang and Tianyi Zhou and Saining Xie and Silvio Savarese and Le Xue and Caiming Xiong and Ran Xu},
  journal={arXiv preprint arXiv:2505.09568},
  year={2025},
}

\clearpage

\end{document}